\documentclass[onecolumn]{article}         
\usepackage[top=3.7cm, bottom=3.5cm, left=2.5cm, right=2.5cm]{geometry}

\usepackage{color}
\usepackage{graphicx}
\usepackage{cite}
\usepackage[cmex10]{amsmath}
\usepackage{amssymb}
\usepackage{subfigure}
\usepackage{rotating}
\usepackage{multirow}
\usepackage{makecell}
\usepackage[colorlinks=true,linkcolor=red,citecolor=blue]{hyperref}
\usepackage{algorithmic}

\usepackage{algorithm}
\begin{document}

\title{Image Stitching by Line-guided Local Warping\\with Global Similarity Constraint}

\author{Tianzhu~Xiang$^1$, Gui-Song~Xia$^1$, Xiang~Bai$^2$, Liangpei~Zhang$^1$\\
\\
$^1${\em State Key Lab. LIESMARS, Wuhan University, Wuhan, China.}\\
$^2${\em Electronic Information School, Huazhong University of Science and Technology, China.}
}

\maketitle
\begin{abstract}
Low-textured image stitching remains a challenging problem. It is difficult to achieve good alignment and it is easy to break image structures due to insufficient and unreliable point correspondences. Moreover, because of the viewpoint variations between multiple images, the stitched images suffer from projective distortions. To solve these problems, this paper presents a line-guided local warping method with a global similarity constraint for image stitching. Line features which serve well for geometric descriptions and scene constraints, are employed to guide image stitching accurately. On one hand, the line features are integrated into a local warping model through a designed weight function. On the other hand, line features are adopted to impose strong geometric constraints, including line correspondence and line colinearity, to improve the stitching performance through mesh optimization. To mitigate projective distortions, we adopt a global similarity constraint, which is integrated with the projective warps via a designed weight strategy. This constraint causes the final warp to slowly change from a projective to a similarity transformation across the image. Finally, the images undergo a two-stage alignment scheme that provides accurate alignment and reduces projective distortion. We evaluate our method on a series of images and compare it with several other methods. The experimental results demonstrate that the proposed method provides a convincing stitching performance and that it outperforms other state-of-the-art methods.
\end{abstract}
%
%
%

\section{Introduction}
\label{sec:intro}

Because images are limited by a camera's narrow field of view (FOV), image stitching combines a group of images with overlapping regions to generate a single, but larger, mosaic with a wider FOV. Image stitching has been widely used in many tasks in photogrammetry~\cite{pang2016sgm-based}, remote sensing~\cite{li2015a} and computer vision~\cite{Zhang2014, Brown2003}. 

In the literature~\cite{Zaragoza2014}, there are typically two main approaches that have been attempted to produce image stitching with satisfactory visual results: (1) developing better alignment models and (2) employing image composition algorithms, such as seam cutting~\cite{Gao2013} and blending~\cite{Wang2011}. Image alignment is the first and  most crucial step in image stitching. Although advanced image composition methods can reduce stitching artifacts and improve the stitching performance, they cannot address obvious misalignments. When a seam or blending area coincides with misaligned areas, the current image composition schemes will fail to provide a satisfactory stitched image~\cite{Hu2015}. 

Most previous image stitching methods estimate global geometric transformations (\emph{e.g.}, similarity, affine or projective transformation) to bring the overlapping images into alignment. However, these methods require the camera rotation to have a fixed projective center or the scenes to have limited depth variance~\cite{Szeliski2006}, which are restrictive assumptions that are often violated in practice, resulting in artifacts in the stitched images, \textit{e.g.,} misalignments or ghosting.

\begin{figure*}[!ht]
	\centering
	\subfigure[The original images and the detected features]{
		\includegraphics[width=0.9\textwidth]{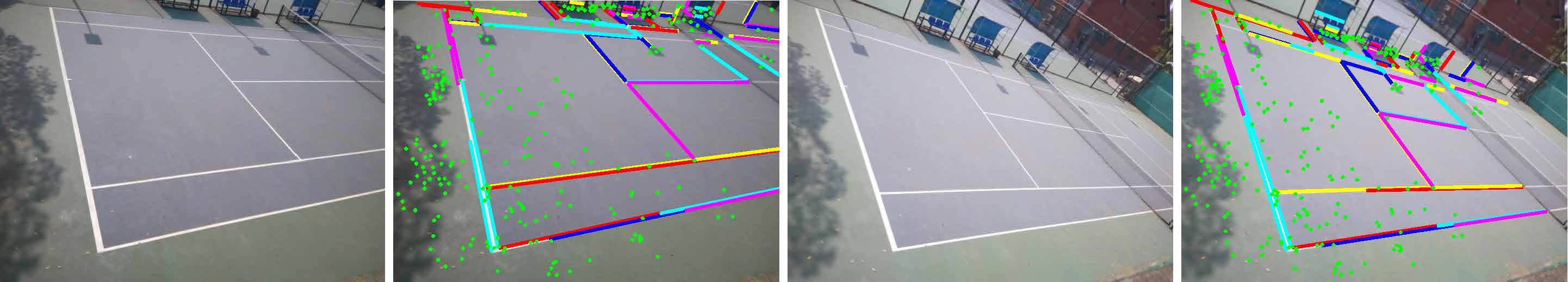}
	} \\[0.1mm]
	\subfigure[Stitching results of global homography, CPW, APAP and our method.]{
		\begin{minipage}[b]{0.9\textwidth}
			\includegraphics[width=1\textwidth]{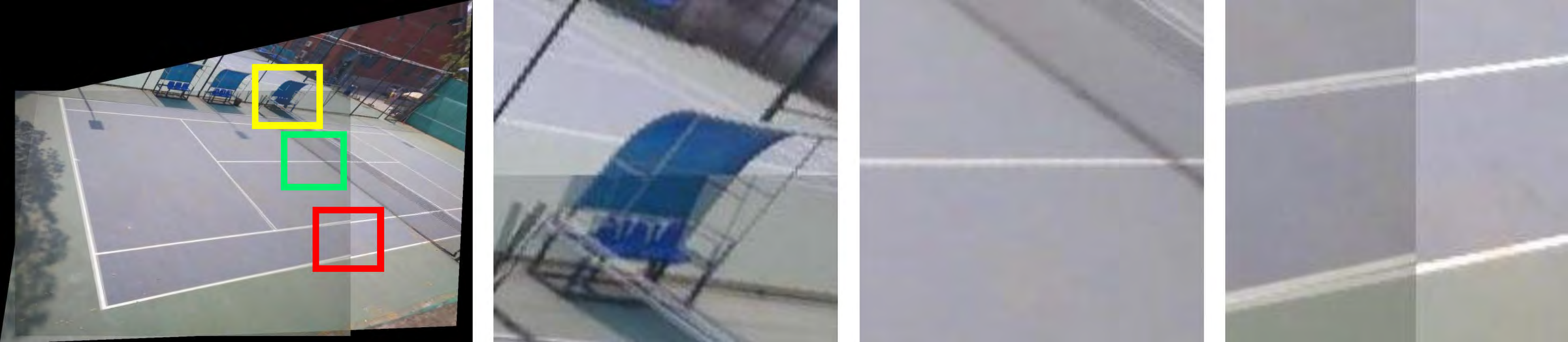} \\[1mm]
			\includegraphics[width=1\textwidth]{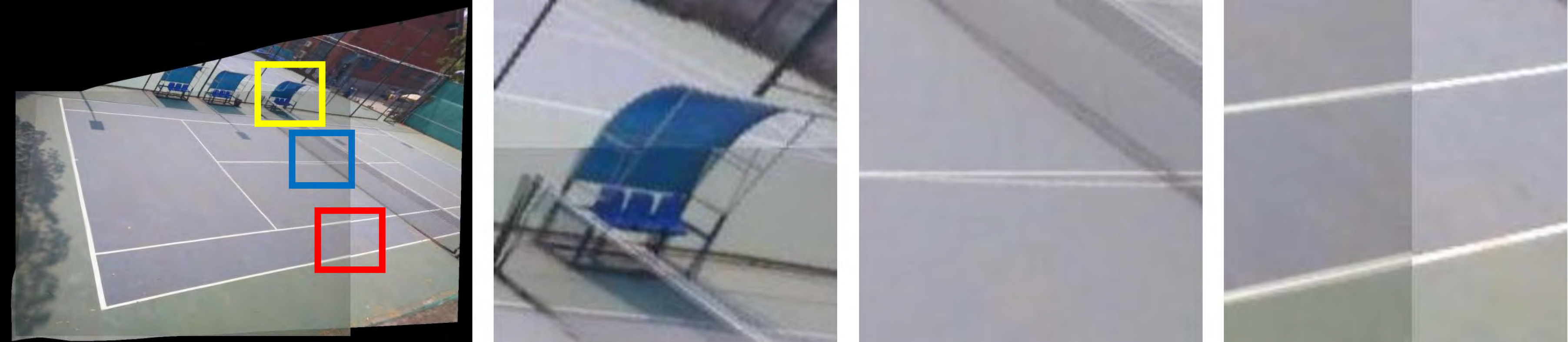} \\[1mm]
			\includegraphics[width=1\textwidth]{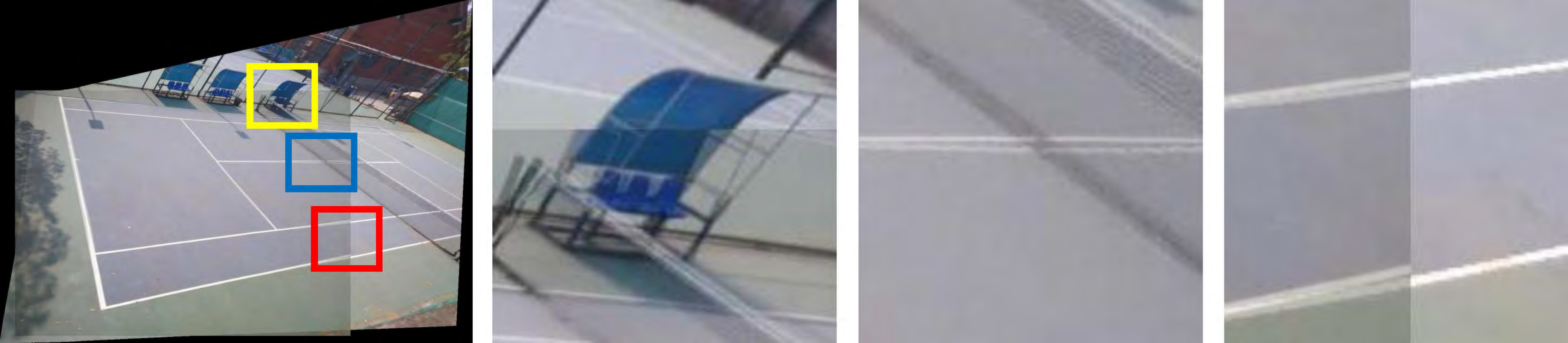} \\[1mm]
			\includegraphics[width=1\textwidth]{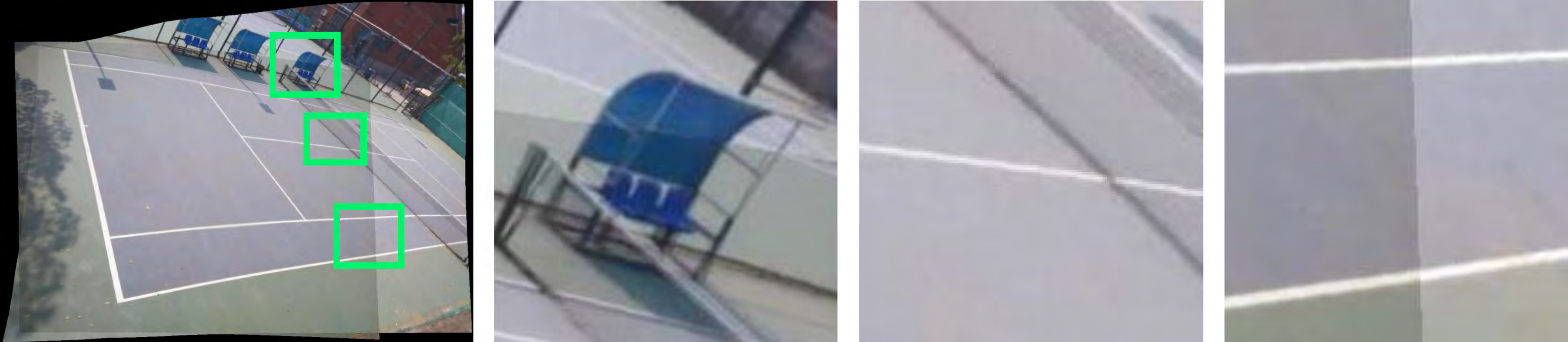}
		\end{minipage}
	}
	\caption{The challenges in image stitching. (a) The original images and the detected features (points $\&$ lines). (b) The stitching results. From top to bottom, the images are the results of global homography~\cite{Brown2007}, CPW~\cite{Liu2009}, APAP~\cite{Zaragoza2014}, and our proposed method. The details are highlighted to simplify the comparison. The red boxes denote alignment errors; the yellow boxes show distortions; the blue boxes denote structural deformations; and the green boxes denote satisfactorily stitched areas with no misalignments or distortions.}
	\label{fig_prob}
\end{figure*}

To compensate for these geometric assumptions, some spatially-varying warping methods for image stitching have been proposed in recent years that can be roughly categorized into two groups: multiple homographies and mesh-based warping. The former estimates multiple homographies that are compatible with local geometries to align the input images, \textit{e.g.,} \emph{as-projective-as-possible} (APAP) warping~\cite{Zaragoza2014}. Mesh-based warping first pre-warps the image using global homography; then, it adopts some energy functions to optimize the alignment, treating it as a mesh warping problem, \textit{e.g.,} \emph{content-preserving warping} (CPW)~\cite{Liu2009}. The high degrees of freedom (DoFs) involved in these methods can better handle parallax than can global transformations; thus, they can provide satisfactory stitching results. However, some challenges remain to be addressed:

\begin{enumerate}
	\item[-] The current methods often fail to achieve satisfactory alignment in low-texture images. Due to the high DoFs, these methods inevitably depend heavily on point correspondences~\cite{XuMGN2014}. However, keypoints are difficult to detect in some low-texture images because the homogeneous regions, such as indoor walls, sky, artificial structures, are not distinctive enough to provide rich and reliable correspondences. Hence, these methods often erroneously estimate the warping model, which causes misalignments. 
	\item[-] The influence of projective distortions has not been fully considered. Because many methods are based on projective transformations, \textit{e.g.,} CPW~\cite{Liu2009}, APAP~\cite{Zaragoza2014}, the stitched results of images taken under various photographing viewpoints may suffer from projective distortions~\cite{Chang2014} in the non-overlapping regions, including both shape and perspective distortions. For instance, some regions in the stitched image may be stretched or non-uniformly enlarged, and it is difficult to preserve the perspective of each image (Fig.~\ref{fig_prob}(b), Fig.~\ref{fig7}(a)).
	\item[-] The image structure distortion has not been fully considered. Some local warping models, \textit{e.g.,} CPW~\cite{Liu2009}, APAP~\cite{Zaragoza2014}, may bend line structures, especially when stitching low-texture images. For instance, insufficient or unreliable keypoints cause APAP to erroneously estimate some local transformations, which results in misalignment of the local regions and distorts the line structures that span multiple local regions, while CPW employs  only feature correspondences and content smoothness to optimize the global transformation and does not consider structural constraints.
\end{enumerate}

The challenges to image stitching can be clearly seen in Fig. 1. Fig. 1 (a) shows the original images and the detected features (points and lines). In some homogeneous regions, only a few points are detected and matched, making it difficult to estimate an accurate transformation. Fig. 1 (b) shows the stitching results from global homography~\cite{Brown2007}, CPW~\cite{Liu2009}, APAP~\cite{Zaragoza2014} and the proposed method. When the restrictive imaging conditions are violated, the global homography model does not fit the data correctly; thus, it results in obvious misalignments (the red boxes). In low-textured areas with insufficient correspondence (red boxes), CPW lacks sufficient data to align the pre-warping result, and APAP cannot estimate accurate local homographies, causing obvious misalignments. The lack of point correspondences also leads to structural deformations in CPW and APAP (blue boxes), where straight lines are deformed into curves. Due to the projective transformation used in these three models and the fact that no measures are taken to eliminate distortions, the stitched image results of these methods suffer from severe projective distortions (the yellow boxes), where the chairs are enlarged non-uniformly. 

The above problems provide strong motivation for improving the performance of image stitching. To our knowledge, only a few studies have been conducted to address either of the aforementioned problems; consequently, additional efforts are needed. Recent studies (\cite{Joo2015} and~\cite{Li2015}) have reported that line features can be used to improve the alignment performance, and \cite{Chang2014} and~\cite{Lin2015} recently showed that similarity transformations are advantageous in reducing distortions. Inspired by these studies, our work is based on the following two assumptions: 

\begin{enumerate}
	\item[-] In most man-made environments, line features are relatively abundant, thus they can be regarded as effective supplements that can provide rich correspondences for accurate warping model estimation~\cite{XiangXXZ17}. Furthermore, line features depict the geometrical and structural information of scenes~\cite{Xia2014, XXZ2016, XueXBZS2018}; thus, they can also be used to preserve the image structures.
	\item[-] Similarity transformation~\cite{Chang2014} does not introduce shape distortion because it consists only of translation, rotation and uniform scaling. A similarity transformation can be regarded as a combination of panning, zooming and in-plane camera rotation; therefore, it preserves the viewing direction. 
\end{enumerate}

It is thus of great interest to investigate how to integrate line features and global similarity transformation to improve the image stitching performance. To this end, this paper presents a line-guided local warping model for image stitching with a global similarity constraint. More precisely, this method adopts a two-stage scheme to achieve good alignment. First, pre-warping is jointly estimated using both point and line features. Then, extended mesh-based warping is used to further align the pre-warping result. Line features are integrated into mesh-based warping framework and act as structural constraints to preserve image structures. Finally, to prevent  undesirable distortions, the global similarity transformation is adopted as a similarity constraint and used to adjust the estimated warping model. The contributions of our work are as follows:

\begin{itemize}
	\item[-] We introduce line features to guide image stitching, especially in low-texture cases. Line features play a significant role mainly in two aspects: 1) they are integrated into the local warping model using a weight function to achieve accurate alignment; 2) they are employed to impose strong geometric constraints (i.e. line correspondence and line collinearity) to refine the stitching performance.
	\item[-] We present a weight integration strategy to combine the global similarity constraint with models of global homography or multiple homographies. Using this strategy, the resultant warp achieves a smooth transition from a projective to a similarity transformation across the image, which significantly mitigates the projective distortions in non-overlapping regions.
	\item[-] We propose a robust and effective two-stage stitching framework that combines the local multiple homographies model and the mesh-based warping model with line and global similarity constraints. The proposed method addresses local variation well to ensure image alignment by local stitching and flexible refinement. The method also preserves image structures and multi-perspective through strong geometrical and structural constraints.	The proposed method achieves a state-of-the-art performance.
\end{itemize}

The remainder of this paper is organized as follows. Section~\ref{sec:work} gives a brief review of the related works. Section~\ref{sec:method} describes the proposed method in detail. The experimental results and analyses are reported in Section~\ref{sec:experiment}. Finally, we draw some conclusions and provide remarks in Section~\ref{sec:conclusion}.


\section{Related works}
\label{sec:review}

Numerous studies have been devoted to image stitching; a comprehensive survey can be found in~\cite{Szeliski2006}. The global homography model~\cite{Brown2007} works well for planar scenes or for scenes acquired with parallax-free camera motion, but violation of these assumptions may lead to ghosting artifacts.

Recently, spatially-varying warping methods have been proposed that flexibly address parallax. Liu \emph{et al.}~\cite{Liu2009} proposed the \emph{content-preserving warping} (CPW) method, which was first used in video stabilization. CPW adopts registration error and content smoothness to refine the pre-warping result obtained by global homography. A simple extension of global homography method was presented in~\cite{Gao2011}, called \emph{dual-homography warping} (DHW), which divides the entire scene into two planes: a distant plane and a ground plane. The final warping is obtained by a linear combination of these two homographies estimated by the point correspondences of each plane. However, this method has difficulties on complex scenes. Lin \emph{et al.}~\cite{Lin2011} proposed the \emph{smoothly varying affine} (SVA) warping method for image stitching. SVA can handle local deformations while preserving global affinity. However, because there are insufficient DoFs in the affine model, SVA cannot achieve projective warping. Zaragoza \emph{et al.}~\cite{Zaragoza2014} extended the previous method and proposed an \emph{as-projective-as-possible} (APAP) warping method for image stitching. APAP achieves a smoothly varying projective stitching field estimated by a moving direct linear transformation (DLT)~\cite{Hartley2003}. It maintains a global projection while allowing local non-projective deviations. Zhang \emph{et al.}~\cite{Zhang2014} proposed a parallax-tolerant image stitching method that seeks the optimal homography evaluated by the seam cost and uses CPW to refine the alignment. However, except for SVA, these methods are based on projective transformations, thus the stitched images often suffer from projective distortions. In addition, the resulting images may suffer from structural deformations because of the nonlinear loceal transformations in the model.

In recent years, similarity transformation, which is composed of translation, rotation and scaling, was introduced. Similarity transformation constructs a combined warping with projective transformations to constrain the projective distortions. Chang \emph{et al.}~\cite{Chang2014} proposed a \emph{shape-preserving half-projective} (SPHP) warping for image stitching that adopts projective, transition and similarity transformation to achieve a gradual change from a projective to a similarity transformation across the image. SPHP can significantly reduce the distortions and preserve the image shape; however, it may introduce structural deformations, \textit{e.g.,} line distortions, when the scene is dominated by line structures. Lin \emph{et al.}~\cite{Lin2015} proposed an \emph{adaptive as-natural-as-possible} (AANAP) warping that linearizes the homography in the non-overlapping regions and combines these homographies with a global similarity transformation using a direct and simple distance-based weight strategy to mitigate perspective distortions. However, some distortions still exist locally when stitching images (Fig.~\ref{fig_templedis}(b)).

It is worth noting that spatially-varying warping-based image stitching is highly dependent on point correspondences. When there are insufficient reliable keypoints (such as in low-texture images), the effects of the estimated models will degrade. More recently, Joo \emph{et al.}~\cite{Joo2015} introduced line correspondences into the local warping model, but this approach requires a user to annotate the straight lines, and setting the parameters for this method is complex. Li \emph{et al.}~\cite{Li2015} proposed a dual-feature warping method for motion model estimation that combines line segments and points to estimate the global homography. However, this method still suffers from projective distortions.

\section{The proposed approach}
\label{sec:method}

\begin{figure*}[ht!]
	\centering
	\includegraphics[width=0.98\linewidth]{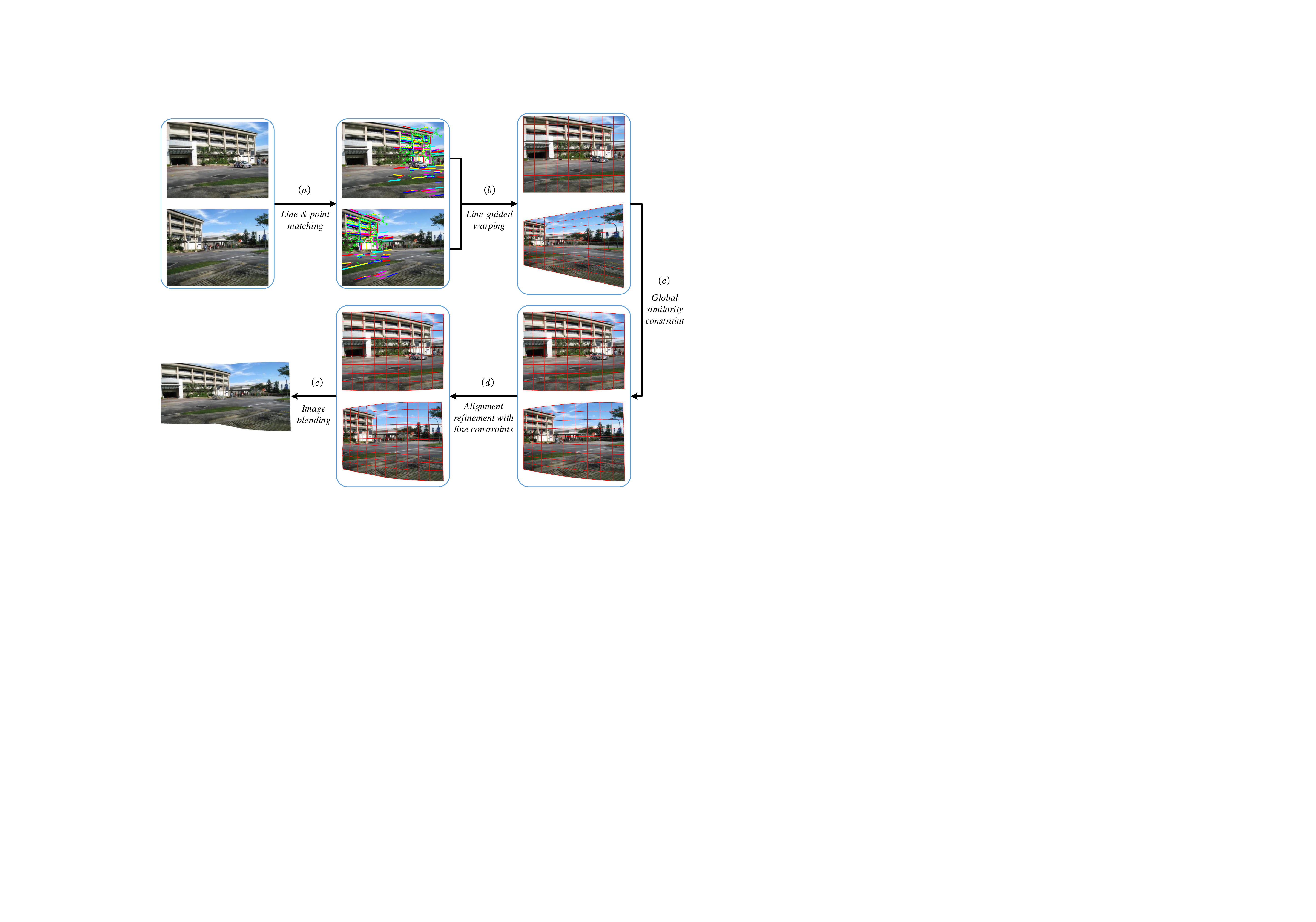}
	\caption{Flowchart of the proposed approach: (a) feature matching, (b) line-guided warping, (c) distortion reduction by global similarity constraint, (d) alignment refinement with line constraints, (e) image blending.}
	\label{fig_flow}
\end{figure*}

This section introduces the proposed method for image stitching in detail. The main idea is to integrate line constraints and a global similarity constraint into a two-stage alignment framework. The outline of our method is illustrated in Fig.~\ref{fig_flow}. The first-stage alignment (presented in Section~\ref{sec:line}) involves estimating an accurate warping model using line guidance. Linear features are adopted as alignment constraints to jointly estimate both global and local homography with point correspondences, which provide rich and reliable correspondences even in low-texture images. To further improve the stitching performance, we adopt mesh optimization based on the extended content-preserving warping framework presented in Section~\ref{sec:mesh}. Then, the linear feature constraints (\emph{i.e.}, line correspondence and line collinearity) are combined to further refine the alignment and preserve the image structures. Finally, to mitigate the projective distortions, a global similarity transformation, estimated by a set of selected points in the approximate image projection plane, is employed to constrain the distortions caused by projective warping via a weighted integration strategy (Section~\ref{sec:sim}). Based on the proposed warping model, we are able to achieve accurate and distortion-free image stitching.

\subsection{Line-guided warping model}
\label{sec:line}

Point features are often adopted for image alignment. Given the target and reference images $I, I^{'}$, $\mathbb{R} \times \mathbb{R} \mapsto \mathbb{R}$, and a pair of matching points: $\mathbf{p}=[x,y,1]$ and $\mathbf{p^{'}}=[x^{'},y^{'},1]$ where $x,y\in \mathbb{R}$, the global homography, $\mathbf{H}\in\mathbb{R}^{3 \times 3}$: $\mathbf{p^{'}} = \mathbf{Hp}$, can be estimated by minimizing the algebraic distance $\sum\nolimits_i {{\left\| {\mathbf{p}_i^{'} \times \mathbf{H} {\mathbf{p}_i}} \right\|}^2} $ between a set of matching points, where $i$ is the index of matching points.

However, as stated previously, keypoints extracted from images are rare in some low-texture scenarios, thus it is difficult to estimate an accurate global homography for image stitching. Hence, line features, which are salient in artificial scenarios, are adopted as the alignment constraint to guide the global homography estimation.

Let $\mathbf{l}=[ {a,b,c} ]^T$, $\mathbf{l^{'}} = [ {a^{'},b^{'},c^{'}} ]^T$, with $a,b,c\in \mathbb{R}$ be a pair of matching lines in the target and reference images respectively. Here, $\mathbf{p}^{0,1}=[x^{0,1},y^{0,1},1]$ denotes the two endpoints of line $\mathbf{l}$. They satisfy ${{\mathbf{l^{'}}}^T} \mathbf{H} {\mathbf{p}^{0,1}}=0$, which means that the endpoints transformed by $\mathbf{H}$ from $\mathbf{l}$ should lie on the corresponding line $\mathbf{l}^{'}$. Therefore, $\mathbf{H}$ can be estimated by minimizing the algebraic distance $\sum\nolimits_j {{{\left\| { {\mathbf{l}_j^{'}}^T \times \mathbf{H} \mathbf{p}_j^{0,1} } \right\|}^2}} $ using a set of matching lines, where $j$ is the index of the matching lines.	

The homography is then estimated jointly by point and line correspondences:
\begin{equation}\label{Eq1}
	\begin{split}
		\mathbf{\hat h}
		& = \mathop {\arg \min }\limits_\mathbf{h} \left( {\sum\nolimits_i {{{\left\| {\mathbf{p}_i^{'} \times \mathbf{H} \mathbf{p}_i } \right\|}^2}}  + \sum\nolimits_j {{{\left\| {\mathbf{l}{{_j^{'}}^T} \times \mathbf{H} \mathbf{p}_j^{0,1}} \right\|}^2}} } \right)  \\
		& = \mathop {\arg \min }\limits_\mathbf{h} \left( {\sum\nolimits_i {{{\left\| {{\mathbf{A}_i} \mathbf{h}} \right\|}^2}}  + \sum\nolimits_j {{{\left\| {{\mathbf{B}_j} \mathbf{h}} \right\|}^2}} } \right), \quad s.t. \left\| \mathbf{h} \right\| = 1,
	\end{split}
\end{equation}
where $\mathbf{h}=[{h_1},{h_2},{h_3},{h_4},{h_5},{h_6},{h_7},{h_8},{h_9}]$ is the column vector representation of $\mathbf{H}$, and $\mathbf{A}_i$, $\mathbf{B}_j \in \mathbb{R}^{2 \times 9}$ are the coefficient matrixes computed by the $i\textendash$th matching point and $j\textendash$th matching line, respectively. 

Stacking all the coefficient matrices of points ($\mathbf{A}_i$) and lines ($\mathbf{B}_j$) vertically into a unified matrix, $\mathbf{C} = [\mathbf{A};\mathbf{B}]$, and Eq.~\eqref{Eq1} can be rewritten as follows:
\begin{equation}\label{}
	\mathbf{\hat h}
	= \mathop {\arg \min }\limits_\mathbf{h} {{\left\| {\mathbf{C} \mathbf{h}} \right\|}^2}, \quad s.t. \left\| \mathbf{h} \right\| = 1,
\end{equation}
The global homography $\mathbf{H}$ is the smallest significant right singular vector of $\mathbf{C}$. Note that before estimation, all the entries of the stacked matrices $\left[ {{A_i};{B_j}} \right]$ should be normalized for numerical stability. In this study, we adopt the point-centric normalization approach proposed in~\cite{Dubrofsky2008}.

Local homography can handle parallax better than global homography due to the higher DoFs~\cite{Zaragoza2014}. Therefore, we extend the line-guided global homography to local homographies. The input images are first divided into uniform grid meshes. The local homography $\mathbf{h}_k$ of the $k\textendash$th mesh located at $\mathbf{p}_* = [x_*,y_*]$ is estimated by
\begin{equation}\label{Eq4}
	{\mathbf{h}_k} =
	\underset{\mathbf{h}}{ \mathop{\arg \min }} {{\left\| \mathbf{W}_k \mathbf{C} \mathbf{h} \right\|}^{2}}, s.t \left\| h \right\| = 1,
\end{equation}
where $\mathbf{W}_k=diag\left( \left[ {\mathbf{w}^{p}},{\mathbf{w}^{l}} \right] \right)$, $\mathbf{w}^{p}\in\mathbb{R}^{2N}$, and $\mathbf{w}^{l}\in\mathbb{R}^{2M}$ denote the weight factors for the point and line correspondences, respectively. Specifically, $\mathbf{w}^p=[w^{p_1} w^{p_1} ... w^{p_N} w^{p_N}]$, and $\mathbf{w}^l=[w^{l_1} w^{l_1} ... w^{l_M} w^{l_M}]$. Therefore, the solution is the smallest significant right singular vector of $\mathbf{W}\mathbf{C}$.

\begin{figure}[ht!]
	\centering
	\includegraphics[height=0.3\textwidth]{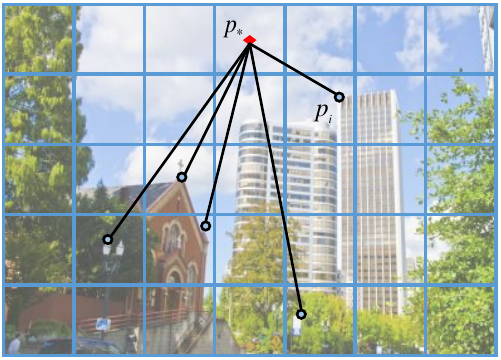}
	\hspace{1mm}
	\includegraphics[height=0.3\textwidth]{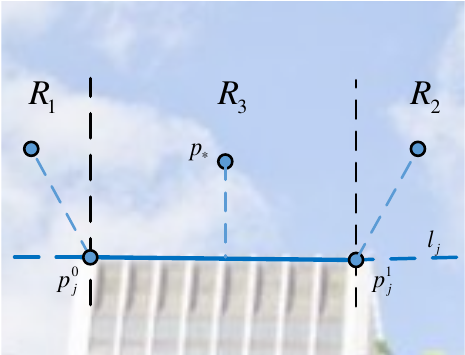}
	\caption{Weight computation of points and lines. Left: (a) The distance between mesh center $\mathbf{p}_{*}$ and keypoint $\mathbf{p}_i$. Right: (b) The distance between mesh center $\mathbf{p}_{*}$ and line segment $\mathbf{l}_j$.}
	\label{fig_dis}
\end{figure}

The point weight factor $\mathbf{w}^p$ is calculated by the Gaussian weighted Euclidean distance:
\begin{equation}\label{Eq5}
	{\mathbf{w}^{{{p}_{i}}}}=
	\max \left( \exp \left( -{{\left\| {\mathbf{p}_{*}}-{\mathbf{p}_{i}} \right\|}^{2}}/{{\sigma }^{2}} \right),\eta  \right),
\end{equation}
where $\mathbf{p}_i$ is the $i\textendash$th keypoint, $\sigma$ is the scale parameter, and $\eta \in [0,1]$ is used to avoid the numerical issues caused by the small weights when the mesh center $\mathbf{p}_*$ is far away from keypoint $\mathbf{p}_i$, as shown in Fig.~\ref{fig_dis}(a).

The line weight factor $\mathbf{w}^l$ is calculated as follows:
\begin{equation}\label{Eq6}
	{\mathbf{w}^{{{l}_{j}}}}=
	\max \left( \exp \left( -{{d}_{l}}{{\left( {\mathbf{p}_{*}},{\mathbf{l}_{j}} \right)}^{2}}/{{\sigma }^{2}} \right),\eta  \right),
\end{equation}
where ${d_l}{(\mathbf{p}_{*},\mathbf{l}_j)}$ is the shortest distance between the mesh center $\mathbf{p}_*$ and line $\mathbf{l}_j$, calculated as follows:

\begin{equation}\label{Eq7}
	{{d}_{l}}({\mathbf{p}_{*}},{\mathbf{l}_{j}})=
	\left\{ \begin{aligned}
		& \min (\left\| {\mathbf{p}_{*}}-{\mathbf{p}_{j}^0} \right\|,\left\| {\mathbf{p}_{*}}-{\mathbf{p}_j^1} \right\|) & \left( a \right) \\
		& \left| {{a}_{j}}{{x}_{*}}+{{b}_{j}}{{y}_{*}}+c_j \right|/\sqrt{a_{j}^{2}+b_{j}^{2}} & \left( b \right) \\
	\end{aligned} ,\right.
\end{equation}
where $\mathbf{p}_{j}^0$, $\mathbf{p}_{j}^1$ are the endpoints of line $\mathbf{l}_j: \mathbf{l}_j=[a_j,b_j,c_j]$. As shown in Fig.~\ref{fig_dis}(b), when $\mathbf{p}_*$ is in the $R_1$ or $R_2$ region, the $d_l$ is calculated by (a), and when $\mathbf{p}_*$ is in the $R_3$ region, $d_l$ is calculated by (b). From Eq.\eqref{Eq5} and \eqref{Eq6}, the weight is greater when the keypoint or line is closer to the mesh center $\mathbf{p}_*$, which causes the local homography to be a better fit for the local structure around $\mathbf{p}_*$.

\subsection{Alignment refinement with line constraints}
\label{sec:mesh}

This section describes the adoption of mesh optimization as the second step of the two-stage alignment scheme to further improve the performance of image stitching. Content-preserving warping is a mesh-based warping method that was first used for video stabilization in \cite{Liu2009} and, later, successfully applied to image stitching \cite{ZhangLiu2014, ChangChen2014, ChenYS2016}. It is well-suited for small local adjustments. In our work, the line feature constraints (\emph{e.g.,} the line correspondence constraint and line colinearity constraint) are integrated into the content-preserving warping framework to both maintain the image structures and refine the alignment satisfactorily.

The target image $I$ is first divided into a regular grid mesh. In our case, the grid mesh is used to guide the image warping. Supposing $\overline{\mathbf{V}}$ denotes the vertices of the grid mesh in the pre-warping image transformed by the line-guided warping model. Alignment refinement is performed to find a group of deformed vertices $\mathbf{V}$ using energy optimization.
An arbitrary point $\mathbf{p}$ in the pre-warping image can be represented by a linear combination of four mesh vertices ${\mathbf{V}} = {\left[ {\mathbf{V}_1,\mathbf{V}_2,\mathbf{V}_3,\mathbf{V}_4} \right]^T}$ in its locating quad: ${\mathbf{p}} = \mathbf{w}^T{\mathbf{V}}$, and $\mathbf{w} = {\left[ {w_1,w_2,w_3,w_4} \right]^T}$ are calculated by inverse bilinear interpolation \cite{Heckbert1989} and sum to 1. Therefore, the image warping problem can be formulated as a mesh warping problem. In fact, it is an optimization problem in which the objective is to accurately align the pre-warping image to the reference image while avoiding obvious distortions. The energy terms used in this paper are detailed below.

\subsubsection{Content-preserving warping}

Content-preserving warping~\cite{ZhangLiu2014} includes three energy terms: a point alignment term, a global alignment term and a smoothness term. 

The point alignment term $E_p$ is used to align the feature points in the target image or pre-warping image to the corresponding points in the reference image as much as possible. It is defined as follows:
\begin{equation}\label{}
	E_p = {\sum\nolimits_{i} {\left\| \mathbf{w}_i^T{ \mathbf{V}_i } - \mathbf{p}_i^{'} \right\|} ^2},
\end{equation}
where $\mathbf{p}_i^{'}$ is the matching point in the reference image. This term ensures the alignment of the overlapping region.

The global alignment term $E_g$ is used to constrain the image regions without feature correspondences to be as consistent as possible with the pre-warping result:  
\begin{equation}\label{}
	E_g = {\sum\nolimits_{i} {\left\| \mathbf{V}_i - \overline{\mathbf{V}}_i \right\|} ^2},
\end{equation}
where $\overline{\mathbf{V}}_i$ is the corresponding vertex in the pre-warping result.

The smoothness term $E_s$ encourages each grid in the pre-warping result to preserve similarity during warping to avoid shape distortions as much as possible. Precisely, given a triangle $\vartriangle \overline{\mathbf{V}}_0 \overline{\mathbf{V}}_1 \overline{\mathbf{V}}_2$ in the pre-warping result, the vertex $\overline{\mathbf{V}}_0$ can be represented by $\overline{\mathbf{V}}_1$ and $\overline{\mathbf{V}}_2$ as shown below:
\begin{equation}\label{}
	\begin{array}{l}
		\overline{\mathbf{V}}_1 = \overline{\mathbf{V}}_2 + \mu (\overline{\mathbf{V}}_3-\overline{\mathbf{V}}_2) + \nu \mathbf{R} (\overline{\mathbf{V}}_3-\overline{\mathbf{V}}_2), 
	\end{array} 
	\mathbf{R} = \left[
	\begin{matrix}
		& 0 & 1\\
		& -1 & 0
	\end{matrix} \right],
\end{equation}
where $\mu, \nu$ are the coordinate values of $\overline{\mathbf{V}}_0$ in the coordinated system defined by the other two vertices. During warping, the triangle uses a similarity transformation to preserve the relative relationship of the three vertices and avoid local distortions. The smoothness term is
\begin{equation}\label{Es}
	E_s(\mathbf{V}_1) = \varphi {\left\| \mathbf{V}_1 - ( \mathbf{V}_2 + \mu (\mathbf{V}_3-\mathbf{V}_2) + \nu \mathbf{R} (\mathbf{V}_3-\mathbf{V}_2))  \right\|} ^2,
\end{equation}
where $\varphi$ is a weight used to measure the salience of the triangle as in \cite{Liu2009}. The weight more strongly preserves the shapes of high-salience regions than those of low-salience regions. The full smoothness energy term is formed by summing Eq.~\eqref{Es} over all the vertices.

\subsubsection{Line correspondence term}

However, content-preserving warping terms only ensure the point alignment in the overlapping regions; thus, the line correspondences are taken into consideration to further improve the alignment.

A line correspondence term is utilized to ensure that the line correspondences are well aligned. Let $\mathbf{l}_j$, $\mathbf{l}_{j}^{{'}}$ be a pair of corresponding lines in the target and reference images, respectively. Line $\mathbf{l}_j$ is cut into several short line segments by the edges of mesh if the line $\mathbf{l}_j$ traverses this mesh. The endpoints of the short line segments from $\mathbf{l}_j$ are denoted by ${\mathbf{p}_{j,k}} $, where $k$ is the index of the endpoints, and ${\mathbf{p}_{j,k}^{'}}$ denotes the endpoints in the pre-warping image transformed from ${\mathbf{p}_{j,k}} $ by the preceding warping process, ${\mathbf{p}_{j,k}^{'}} = \mathbf{w}_{j,k}^T{\mathbf{V}_{j,k}}$. The line correspondence term can be expressed by the idea that the distance from ${\mathbf{p}_{j,k}^{'}} $ to the corresponding line $\mathbf{l}_{j}^{'}$ should be the minimum distance:
\begin{equation}\label{Eq16}
	E_l = {\sum\nolimits_{j,k} {\left\| {(\mathbf{l}{{_j^{'}}^T} \cdot \mathbf{w}_{j,k}^T{\mathbf{V}_{j,k}})/\sqrt {a{{_j^{'}}^2} + b{{_j^{'}}^2}} } \right\|} ^2}.
\end{equation}

The line correspondence term not only enhances the image alignment but also, together with line collinearity term below, preserves the straightness of line structures.

\subsubsection{Line collinearity term}

However, the above terms may not reduce the distortions (\textit{e.g.,} line structure distortions) in the non-overlapping regions where there are few point or line correspondences. To capitalize on the line features and preserve the line structure, we adopt the line collinearity constraint.

The line collinearity term is used to preserve the straightness of linear structures in the target image as much as possible. Let $\mathbf{p}_{i,k}$ denote the endpoints and intersecting points of line $\mathbf{l}_i$ in the non-overlapping regions with the grid. Assume that $\mathbf{p}_{i,k}^{'}$ denotes the corresponding points of $\mathbf{p}_{i,k}$ in the pre-warping result. The line should maintain its straightness after warping, that is, the transformed points $\mathbf{p}_{i,k}^{'}$ should lie on the same line. This can be represented by the distance from the endpoints $\mathbf{p}_{i,k}^{'}$ to the line ${\hat {\mathbf{l}}}_i$ which should be the minimum distance. Line ${\hat {\mathbf{l}}}_i$ is calculated by the head and tail endpoints of $\mathbf{p}_{i,k}^{'}$. The term is defined as follows:
\begin{equation} \label{Eq17}
	E_c =
	{\sum\nolimits_{i,k} {\left\| {({{\hat {\mathbf{l}}}_i}^T \cdot \mathbf{w}_{i,k}^T{\mathbf{V}_{i,k}}) / \sqrt {{{\hat a}_i}^2 + {{\hat b}_i}^2} } \right\|} ^2}.
\end{equation}

Together, the line collinearity term and the line correspondence term maintain the line structures well.

\subsubsection{Objective function}
The above five energy terms are then combined as an energy optimization problem in which the objective function is
\begin{equation}\label{Eq18}
	E = \alpha {E_p} + \beta {E_g} + \gamma {E_s} + \delta {E_l} + \rho {E_c},
\end{equation}
where $\alpha,\ \beta,\ \gamma,\ \delta,\ \rho$ are the weight factors for each energy term. In our implementation, $\alpha = 1, \beta = 0.001, \gamma =0.01, \delta = 1, and ~\rho = 0.001$. The above function is quadratic; consequently, it can be solved by a sparse linear solver. The final result is obtained through texture mapping.

\subsection{Distortion reduction by global similarity constraint}
\label{sec:sim}

To reduce the projective distortions in the non-overlapping regions, the global similarity transformation is adopted to adjust the local warping model.

\begin{figure}[ht!]
	\centering
	\includegraphics[width=0.45\linewidth]{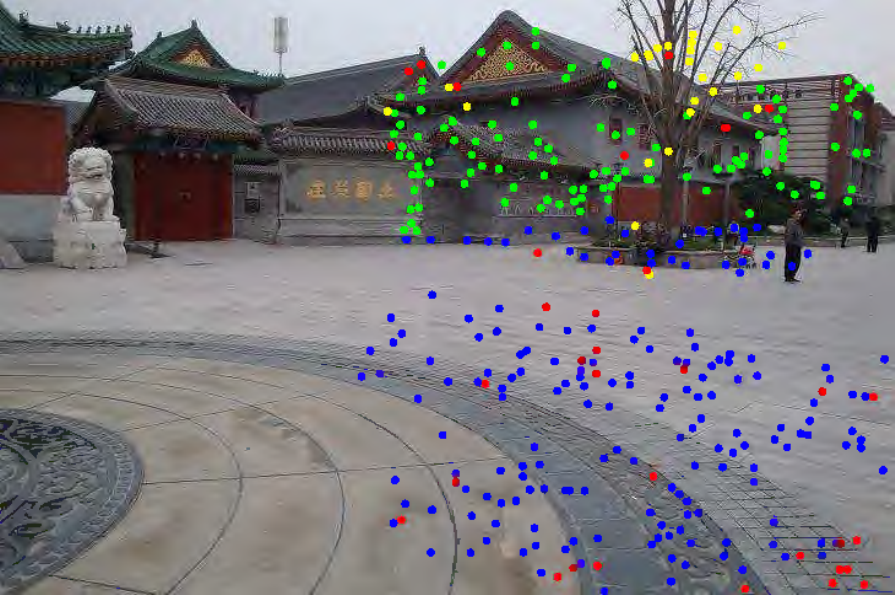}
	\hspace{1mm}
	\includegraphics[width=0.45\linewidth]{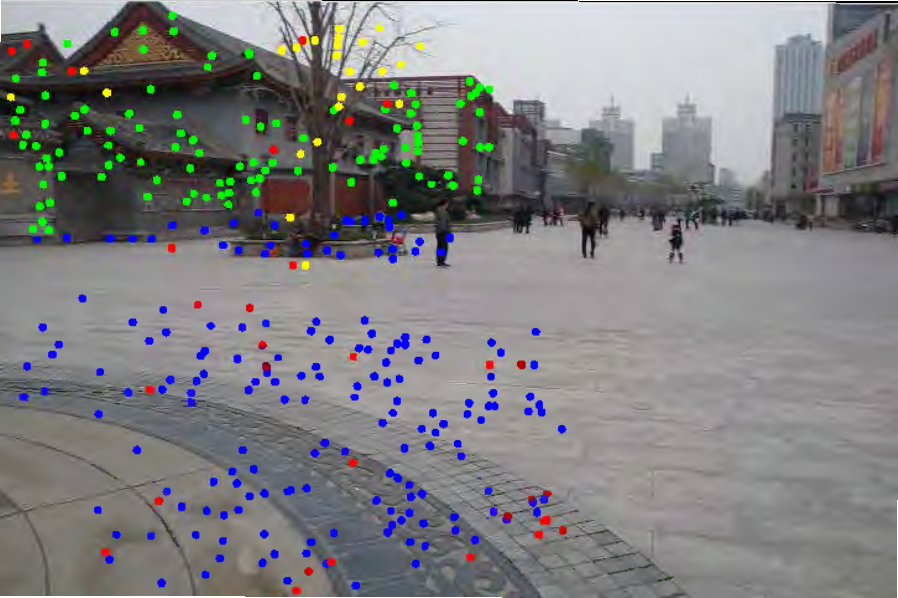}
	\caption{The optimal point correspondences for global similarity transformation estimation.}
	\label{fig_sim}
\end{figure}

Chang \emph{et al.}~\cite{Chang2014} has shown that similarity transformation is effective in mitigating distortions. If we can find a similarity transformation that approximately represents the camera motion of the image projection plane, that transformation can be applied to offset the camera motion~\cite{Lin2015}. RANSAC \cite{Chin2012} is used to iteratively segment the matching points. Each group of point correspondences can be used to estimate a similarity transformation. The estimation with the smallest rotation angle is selected as the optimal candidate~\cite{Xiang2016}. As shown in Fig.~\ref{fig_sim}, the group of points in green is chosen to estimate the global similarity transformation. The plane composed of green points approximates the image projection plane because the camera is nearly perpendicular to the ground when shooting.

\subsubsection{Similarity constraint}
An image patch can be transformed by a projective transformation (e.g. homography), which provides good alignment but may cause distortions, such as stretching. An image patch can also be warped by the similarity transformation, which, although it introduces no distortions, may result in poor alignment due to the limited DoFs. Integrating two types of transformations using weights, can therefore both ensure good alignment and reduce distortions. The similarity constraint procedure is described in \textbf{Algorithm}~\ref{alg}. The global similarity transformation is combined with global or local homographies using weight factors. To create a smooth transition, the whole image should be considered. The weight integration is calculated as follows:
\begin{equation}\label{Eq8}
	\mathbf{H}_i^{'} = \tau {\mathbf{H}_i} + \xi \mathbf{S},
\end{equation}
where $\mathbf{H}_i$ is the homography in the $i\textendash$th grid mesh, and $\mathbf{H}_i^{'}$ is the final homography in the $i\textendash$th grid mesh. Here, $\mathbf{S}$ is the similarity transformation, and $\tau$ and $\xi$ are weight coefficients with $\tau + \xi = 1$. The calculation of these two weights will be described later. In a global homography model, the homography of every grid mesh is the same.

The corresponding warping procedure should also be applied to the reference image because the similarity transformation also adjusts the overlapping regions. The warping procedure for the reference image can be formulated as follows:
\begin{equation}\label{Eq9}
	\mathbf{T}_i^{'} = {\mathbf{H}_i^{'}}{\mathbf{H}_i^{-1}},
\end{equation}
where $\mathbf{T}_i^{'}$ is the warping procedure for the reference image in the $i\textendash$th grid mesh.

\begin{algorithm}[htb]
	\caption{Global similarity constraint}
	\label{alg}  
	\begin{algorithmic}[1]
		\REQUIRE The local homography $\{H_i\}^{n}_{i=1}$ of each pair of images
		\ENSURE The improved homograph $\{H_i^{'}\}^{n}_{i=1}$ constrained by the global similarity transformation		
		\STATE Compute the rotation angle $\theta$ (\ref{Eq13})
		\STATE Construct a $(u,v)$ coordinate system with origin $o$ by rotation
		\FOR {each local homography $H_i$} 
		\STATE Compute the projection distance of the current grid on the $ou$ axis
		\STATE Compute the weight coefficients $\tau$ and $\xi$ (\ref{Eq15})
		\STATE Integrate $H_i$ with the global similarity transformation $S$ (\ref{Eq8})
		\ENDFOR 
		\FOR {each grid in the reference image}
		\STATE Adjust the warp $\mathbf{T}_i^{'} = {\mathbf{H}_i^{'}}{\mathbf{H}_i^{-1}}$
		\ENDFOR
	\end{algorithmic}	
\end{algorithm}

\begin{figure}[ht!]
	\centering
	\includegraphics[width=0.48\linewidth]{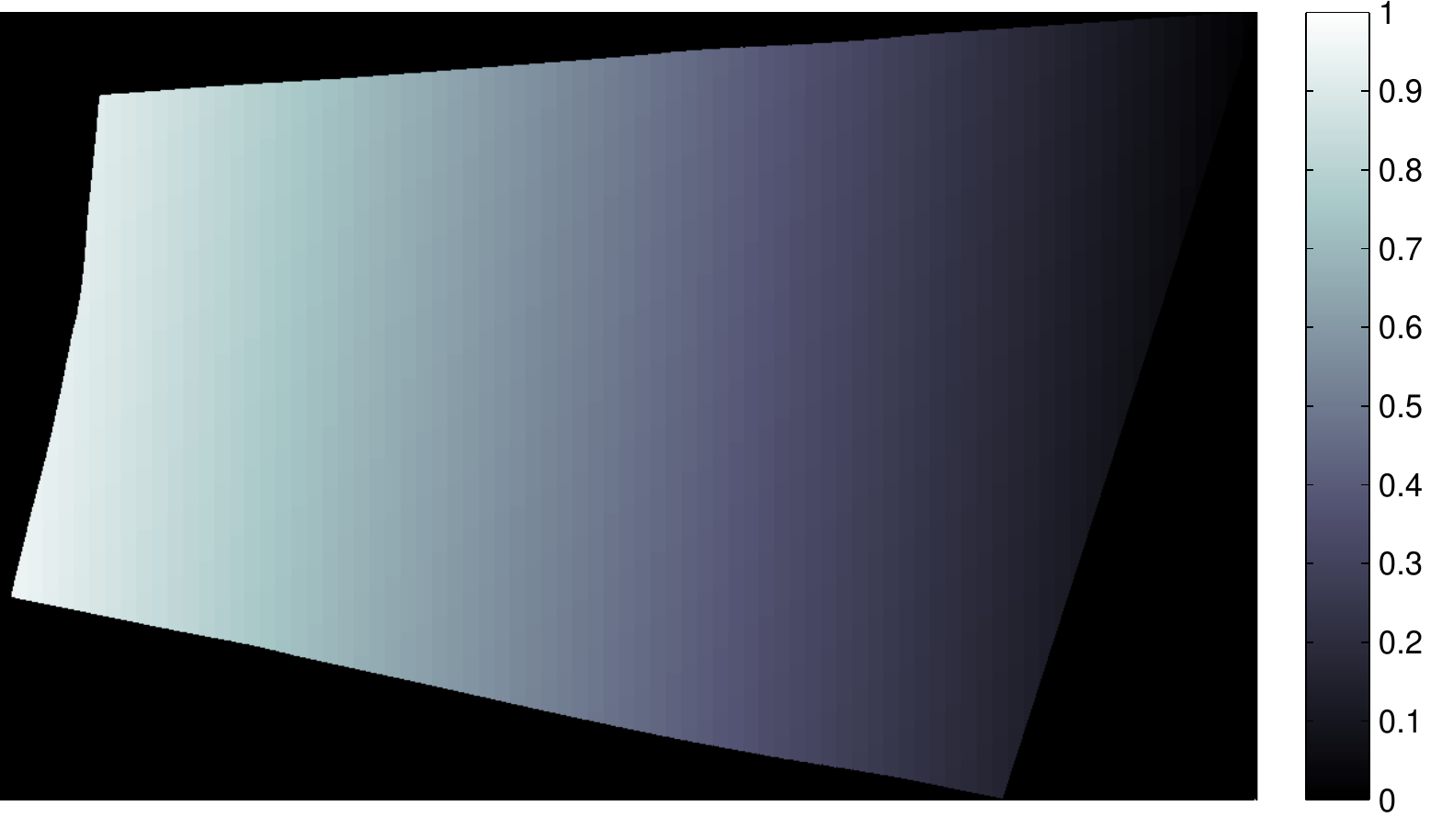}
	\hspace{1mm}
	\includegraphics[width=0.48\linewidth]{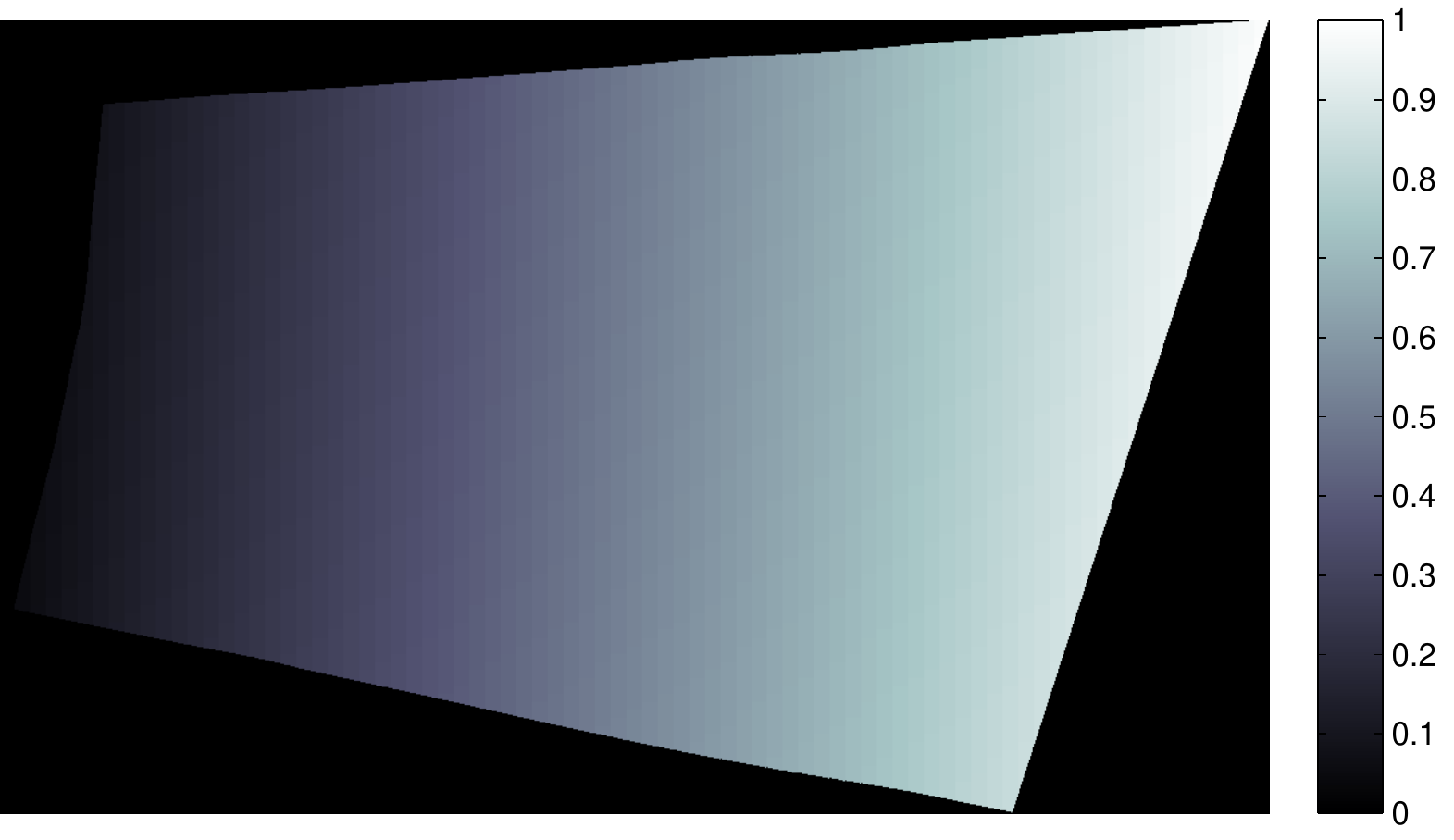}
	\caption{Weight map of the target image in Fig.\ref{fig7}. Left: weight map of homography. Right: weight map of global similarity transformation. The color denotes the weight value.}
	\label{fig_weightmap}
\end{figure}

As shown in Fig.~\ref{fig_weightmap}, when a point is far from the overlapping regions (especially the distorted non-overlapping regions) the procedure assigns a high weight for the similarity transformation to mitigate the distortions as much as possible. In contrast, for points near the overlapping regions, it assigns a high weight for the homography to ensure accurate alignment. Using this weight combination, the final warp smoothly changes from a projective to a similarity transformation across the image, which preserves the image shape and maintains the multi-perspective.

\subsubsection{Weighting strategy}
The weight coefficient calculation stems from the analysis of projective transformation. According to \cite{Chum2005}, let $\mathbf{R}$ be a rotation transformation that transforms the image coordinate $(x,y)$ to a new coordinate $(u,v)$. Based on $\mathbf{p^{'}} = \mathbf{H}\mathbf{p}$, a new projective transformation $\mathbf{Q}$ that transforms $(u,v)$ to $(x^{'},y^{'})$ meets $\mathbf{p^{'}} = \mathbf{Q} [u, v, 1]^T = \mathbf{H}\mathbf{R} [u, v, 1]^T $, where $\mathbf{H} = [h_1, h_2, h_3; h_4, h_5, h_6; h_7, h_8, 1]$, and $\mathbf{Q} = [q_1, q_2, q_3; q_4, q_5, q_6; q_7, q_8, 1]$.

Supposing that the rotation angle is $\theta  = \arctan \left( {{h_8}/{h_7}} \right)$, we can obtain ${q_8} =  - {h_7}\sin\theta + {h_8}\cos\theta = 0$. Then, $\mathbf{Q}$ can be decomposed as follows:
\begin{equation}\label{Eq13}
	\left[ \begin{matrix}
		& {q_1} & {q_2} & {q_3}\\
		& {q_4} & {q_5} & {q_6}\\
		& -c & 0 & 1
	\end{matrix} \right] =
	\underbrace {\left[ \begin{matrix}
			& {q_1} + c{q_3} & {q_2} & {q_3}\\
			& {q_4} + c{q_6} & {q_5} & {q_6}\\
			& 0 & 0 & 1
		\end{matrix} \right]}_{{Q_a}}
	\underbrace {\left[ \begin{matrix}
			& 1 & 0 & 0\\
			& 0 & 1 & 0\\
			& -c & 0 & 1
		\end{matrix} \right]}_{{Q_p}},
\end{equation}
where $c = \sqrt {h_7^2 + h_8^2}$. Here, $\mathbf{Q}_a$ is the affine transformation, and $\mathbf{Q}_p$ is the projective transformation. Defining the local scale change~\cite{Chum2010} at point $(u,v)$ under the projective transformation as the determinant of the Jacobian of $\mathbf{Q}$ at point $(u,v)$, the local scale change is calculated as follows:
\begin{equation}\label{Eq14}
	\det \mathbf{J}\left( {u,v} \right) =
	\det {\mathbf{J}_a}(u,v) \cdot \det {\mathbf{J}_p}\left( {u,v} \right) = {{\lambda}_a} \cdot \frac{1}{{\left( {1 - cu} \right)}^3},
\end{equation}
where $det$ denotes the determinant, and ${\lambda}_a$ is independent of $u$ and $v$. It can be seen that the local area change derived from $\mathbf{Q}$ relies only on the $u$ direction. In other words, the distortions of projective transformation occur only along the $u\textendash$axis. Therefore, the distortions can be effectively eliminated if the weight coefficients are calculated along the $u$ direction in the $(u,v)$ coordinate system.

\begin{figure}[htp!]
	\centering
	\includegraphics[width=0.6\linewidth]{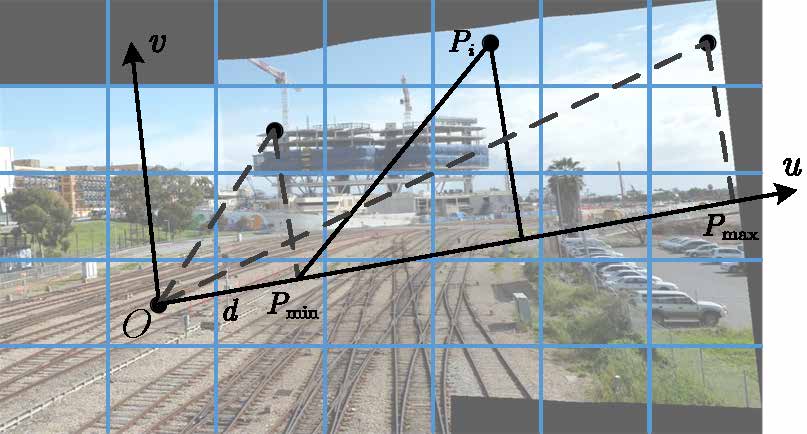}
	\caption{Weight strategy for global similarity transformation and homography.}
	\label{fig_weight}
\end{figure}

The weight coefficients are designed based on the distance of grid points in the $u$ direction; the goal is to provide a gradual change from a projective to a similarity transformation across the image to preserve the image content in non-overlapping regions. As shown in Fig.~\ref{fig_weight}, the center of the reference image is used as the origin of coordination $o$, and the unit vector on the $u\textendash$axis denotes $\overrightarrow{ou} =( {1,0} )$. For the arbitrary mesh center $\mathbf{p}$, $d$ is the projected length of vector $\overrightarrow{o{\mathbf{p}}}$ on the vector $\overrightarrow{ou}$. The projected point ${\mathbf{p}_{max}}$ with a maximum length of $d$ and the projected point $\mathbf{p}_{min}$ with a minimum length of $d$ can be calculated. For the $i\textendash$th grid, the weight coefficients are calculated as follows:
\begin{equation}\label{Eq15}    
	\xi  =
	< \overrightarrow {\mathbf{p}_{min} \mathbf{p}_i} \cdot \overrightarrow {\mathbf{p}_{min} \mathbf{p}_{max}} >  /\left| {\overrightarrow {{\mathbf{p}_{min}}{\mathbf{p}_{max}}} } \right|,
\end{equation}
where $ < \overrightarrow {\mathbf{p}_{min} \mathbf{p}_i} \cdot \overrightarrow {{\mathbf{p}_{max}}{\mathbf{p}_{min}}} > $ denotes the projection length of $\overrightarrow {{\mathbf{p}_{min}}{\mathbf{p}_i}} $ on $\overrightarrow {{\mathbf{p}_{max}}{\mathbf{p}_{min}}} $, and $\tau  = 1 - \xi $.

\begin{figure}[htp!]
	\centering
	\subfigure[APAP Warp]{
		\includegraphics[width=0.46\linewidth]{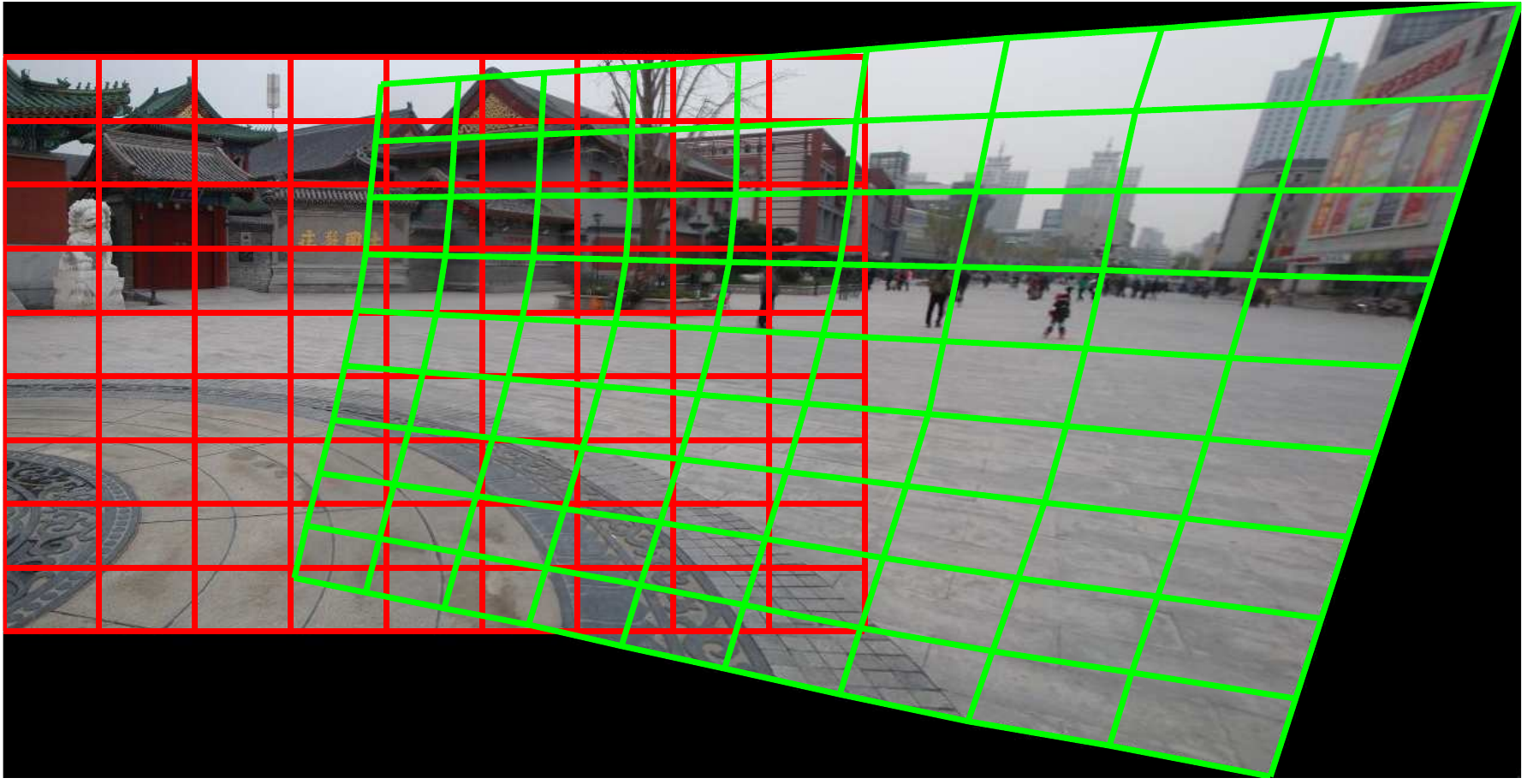}
		\label{fig_sub1}}
	\subfigure[Our Warp]{
		\includegraphics[width=0.46\linewidth]{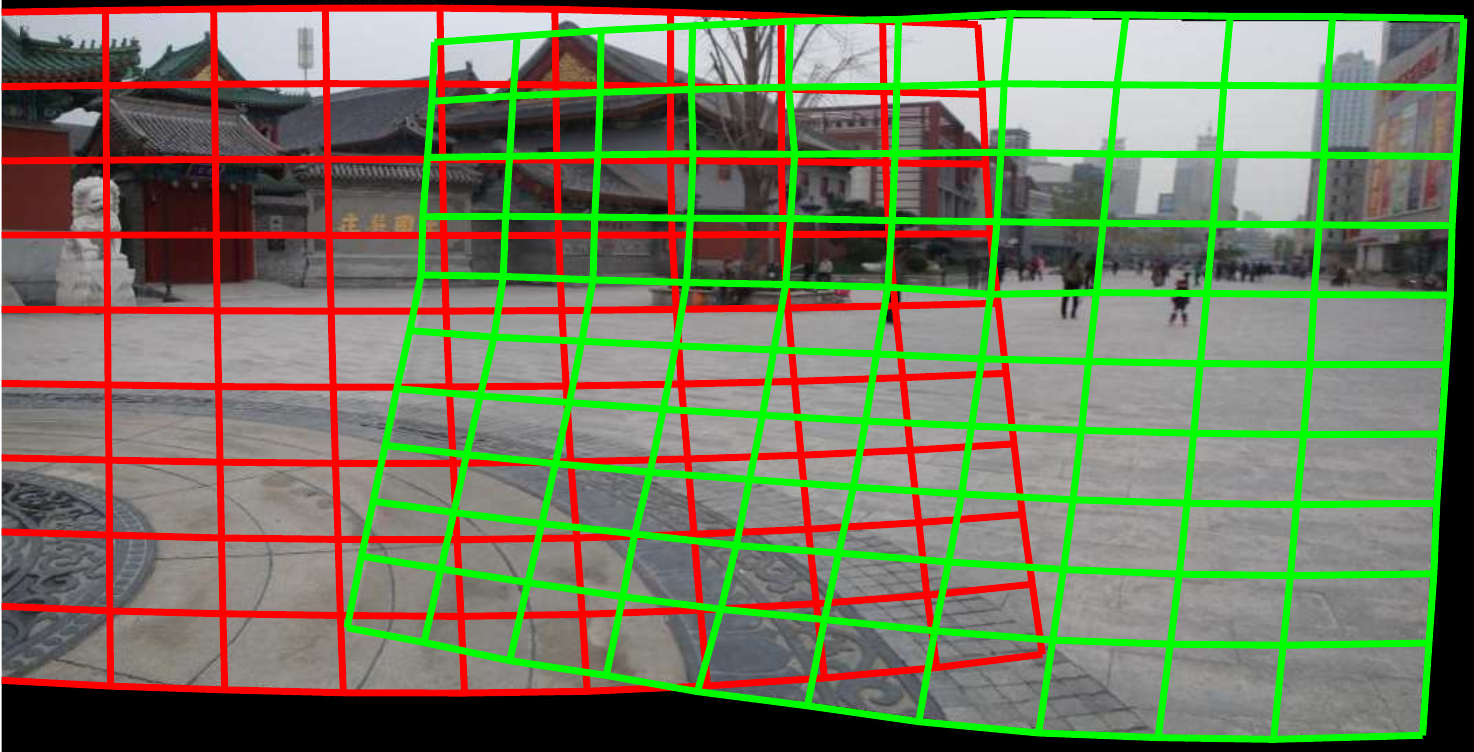}
		\label{fig_sub2}}
	\caption{Performance of similarity constraint to reduce projective distortions.}
	\label{fig7}
\end{figure}

As shown in Fig.~\ref{fig7}, APAP adopts the local homographies for alignment, which aims to be both globally projective while allowing local deviations. However, the stitched image suffers from projective distortions; for instance, the buildings are undesirably stretched and not parallel to the temples, in addition, the perspective distortions in the non-overlapping regions are obvious. In contrast, using a global similarity constraint, the proposed warping model preserves the shapes of objects and maintains the perspective of each image.

\section{Experimental results and analysis}
\label{sec:experiment}

This section describes several experiments conducted to assess the performance of the proposed method on a series of challenging images. In our experiments, the testing images were acquired casually, using different shooting positions and angles.

Given a pair of input images, the keypoints are detected and matched by SIFT~\cite{Lowe2004} in the VLFeat library~\cite{Vedaldi2010vlfeat}. The line features are detected by a line segment detector (LSD)~\cite{LSD2010} and matched by line-point invariants~\cite{Fan2012} or line-junction-line~\cite{Yao2016}. Then, RANSAC is used to remove the mismatches, and the remaining inliers are input to the stitching algorithms. We compared our approach with several other methods. The parameters of the other methods were set as suggested in the respective papers and we used the source code provided by the authors of the papers to obtain the compared results. For our method, $\sigma$ is 8.5, and $\eta$ is 0.01. The experiments were conducted on a PC with an Intel i3-2120 3.3 Ghz CPU and 8 GB of RAM. Not considering feature detection and matching, the proposed method takes 20--30 s to stitch together two images with a resolution of 800$\times$600.

To better compare the methods and reduce interference, we avoided post-processing methods such as blending or seam cutting as detailed in~\cite{Szeliski2006}. Instead, the aligned images are simply blended by intensity average so that any misalignments remain obvious.

To assess the accuracy of the image stitching alignment quantitatively, the metrics of correlation (\emph{Cor})~\cite{Lin2015} and mean geometric error (\emph{$Err_{mg}$})~\cite{Joo2015} are adopted. \emph{Cor} is defined as one minus the \emph{normalized cross correlation} (NCC) over the neighborhood of a $3 \times 3$ window, that is
\begin{equation}\label{Eq19}
	Cor\left( {I,{I^{'}}} \right) =
	\sqrt {\frac{1}{N}\sum\nolimits_\pi { \left( { 1 - NCC\left( {\mathbf{p}},{\mathbf{p}^{'}} \right) } \right)}^2 } ,
\end{equation}
where $N$ is the number of pixels in the overlapping region $\pi$, and $\mathbf{p}$ and $\mathbf{p}^{'}$ are the pixels in image $I$ and $I^{'}$, respectively. $Cor$ reflects the similarity of two images in the overlapping regions. The smaller the $Cor$ value is, the better the stitching result is.

$Err_{mg}$ is defined as the mean geometric error on points and lines, that is
\begin{equation}
	\label{Eq20}
	\begin{array}{c}
		{Err}_{mg}^{(p)}={\frac{1}{M}\sum\nolimits_{i=1}^{M}{{{\left\| f(\mathbf{p}_{i}) - \mathbf{p}_{i}^{'} \right\|}}}} \\[2mm]
		{Err}_{mg}^{(l)}=
		\frac{1}{2K}\sum\nolimits_{j=1}^{K} { \sum\nolimits_{i=0}^{1}{{d}_{l}(f({\mathbf{p}_{l_j}^i}),{\mathbf{l}_{j}^{'}})} }  \\[2mm]
		Err_{mg}=({Err}_{mg}^{(p)}*M+{Err}_{mg}^{(l)}*2K)/(M+2K) 
	\end{array} ,
\end{equation}
where $f: \mathbb{R}^2 \mapsto \mathbb{R}^2$ is the estimated warping, $M$ is the number of point correspondences, $\mathbf{p}_i$ and $\mathbf{p}_i^{'}$ are a pair of point correspondences, $K$ is the number of line correspondences, and ${d}_{l}$ denotes the projected distance of the endpoints of $\mathbf{l}_j$ to its correspondence line $\mathbf{l}_j^{'}$. A smaller $Err_{mg}$ value indicates a better stitching result.

In the following subsections, we first verify the performance of the proposed method on image alignment and distortion reduction. Then, we report the experimental comparison results including the comparison with the global-based methods and the local-based methods.

\subsection{Image alignment}  

\begin{figure*}[htp!]
	\centering
	\includegraphics[width=0.98\textwidth]{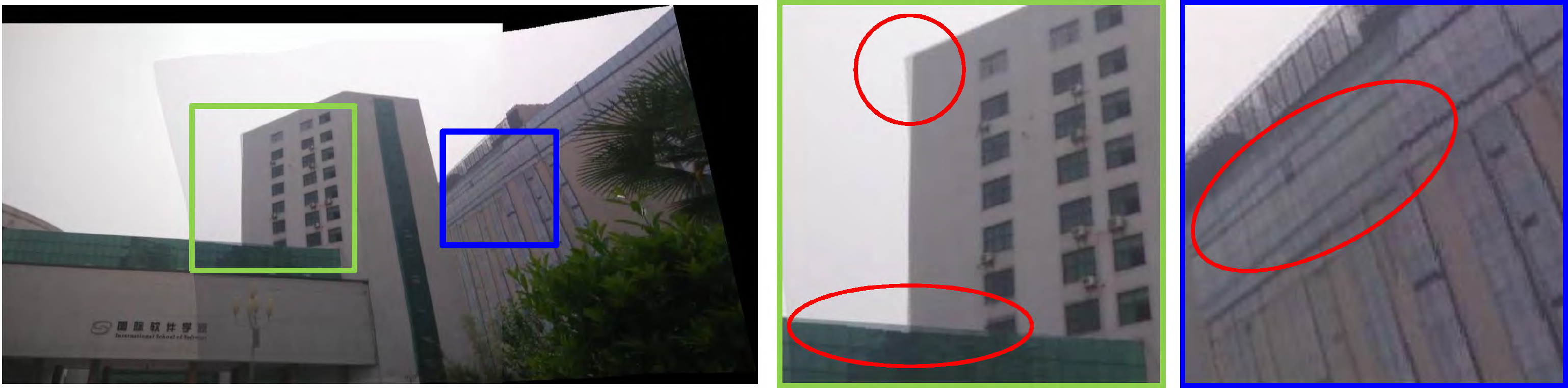}\\[1mm]
	\includegraphics[width=0.98\textwidth]{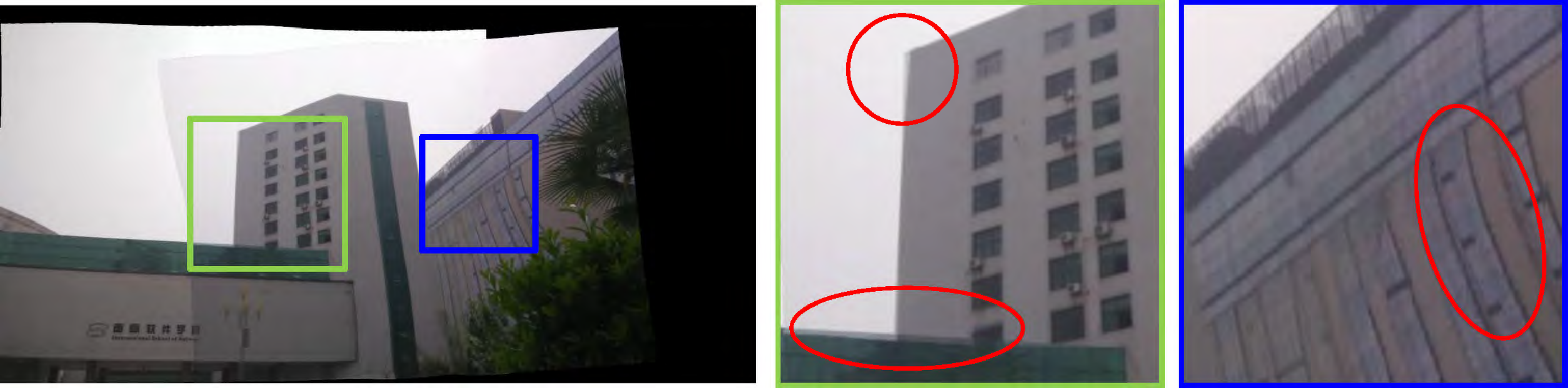}\\[1mm]
	\includegraphics[width=0.98\textwidth]{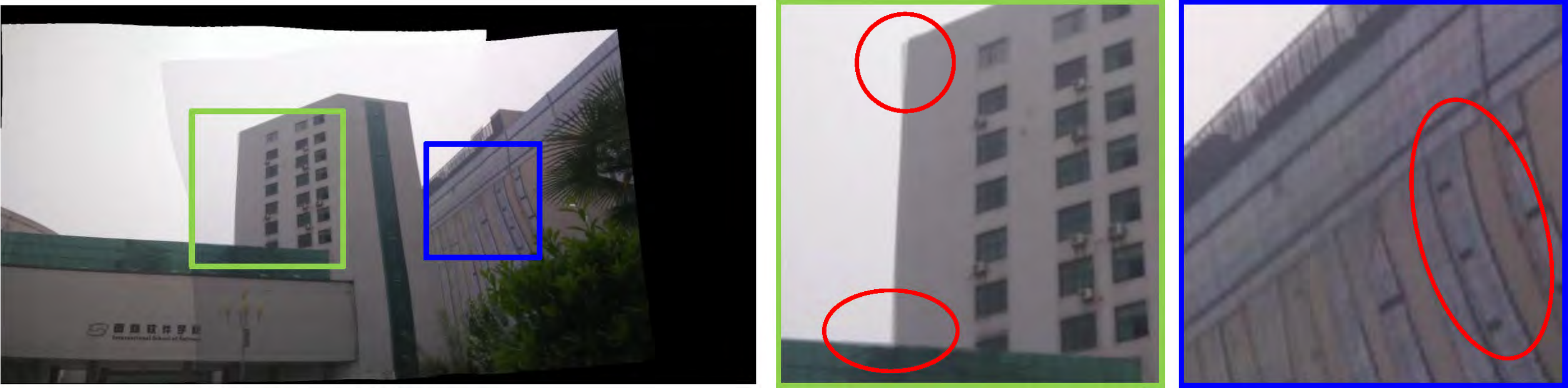}\\[1mm]
	\includegraphics[width=0.98\textwidth]{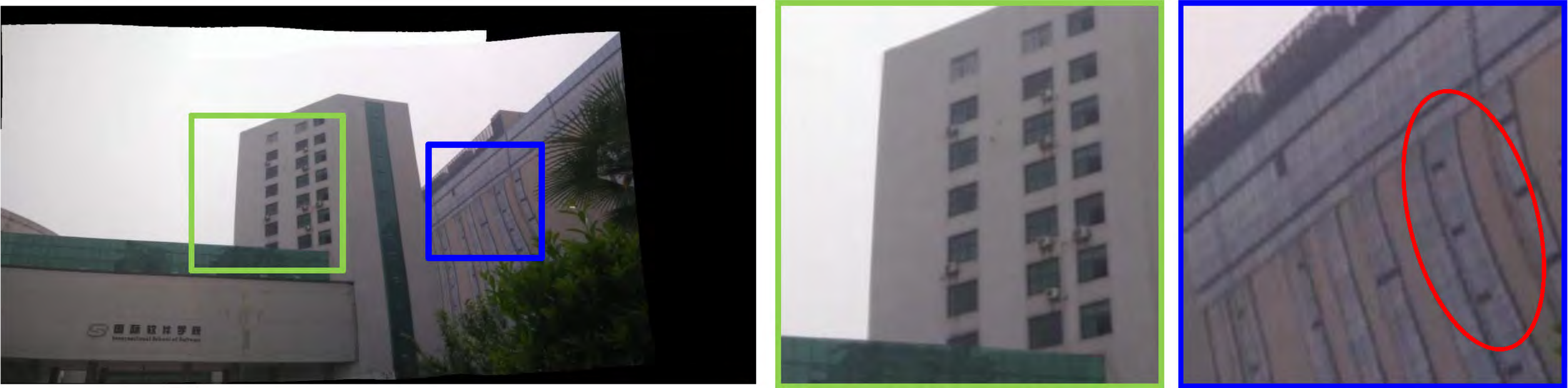}\\[1mm]
	\includegraphics[width=0.98\textwidth]{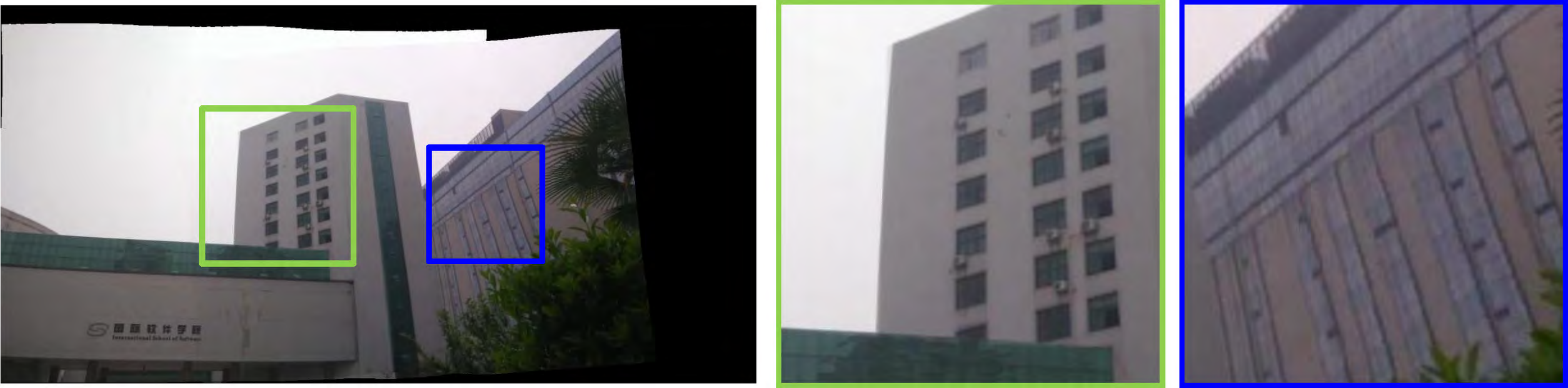}\\
	\caption{Performance of each constraint in proposed method on \emph{School}. From top to bottom: (a) APAP, (b) Line-guided APAP (LAPAP), (c) LAPAP+CPW, (d) LAPAP+CPW+LineCorr, (e) Our method. Red circles show errors or distortions.}
	\label{shool}
\end{figure*}

Fig.~\ref{shool} illustrates the performance of each constraint in the proposed method, including the line-guided local warping estimation, the line correspondence constraint, and the line colinearity constraint. Fig.~\ref{shool}(b) shows the result of line-guided warping combined with APAP (LAPAP), which largely improves the alignment compared to APAP, as can be clearly seen in the closeup. However, LAPAP introduces structural distortions, \textit{e.g.,} the bent lines on the buildings, shown by red circle in the blue closeup. With CPW optimization, LAPAP+CPW refines the alignment performance (shown in Fig.~\ref{shool}(c)), but some slight misalignments still exist. Combined with line correspondence (LineCorr) constraint, LAPAP+CPW+LineCorr provides good alignment (Fig.~\ref{shool}(d)). However, structural distortions, \textit{e.g.}, line deformations, are not handled well as can be clearly seen in the blue closeup. By adding the line collinearity constraint to restrain the structural deformation, the proposed method provides a good stitching result with less distortion in this example (Fig.~\ref{shool}(e)). Quantitative evaluations of  \emph{Cor} and $Err_{mg}$ are shown in Table~\ref{school_eva}, which demonstrates conclusions consistent with the visual effect.

\begin{table}[htp!]
	\centering
	\caption{Comparison of constraints on \emph{School} }
	\label{school_eva}
	\begin{tabular}{ccccc}
		\hline
		Methods         & APAP    & LAPAP   & LAPAP+CPW   & Proposed   \\ \hline
		{\textit{Cor}}  & 1.024   & 0.730   & 0.666       & \textbf{0.664}  \\
		{$Err_{mg}$}    & 2.766   & 2.099   & 1.785       & \textbf{1.708}  \\ \hline
	\end{tabular}
\end{table}

\begin{figure*}[!t]
	\centering
	\includegraphics[width=0.9\textwidth]{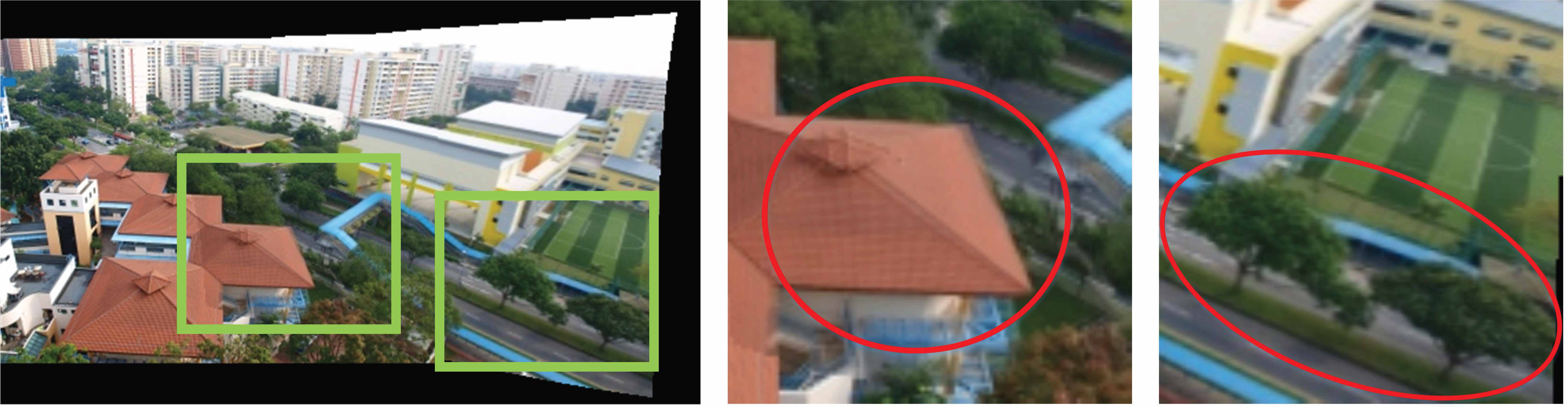}\\[1mm]
	\includegraphics[width=0.9\textwidth]{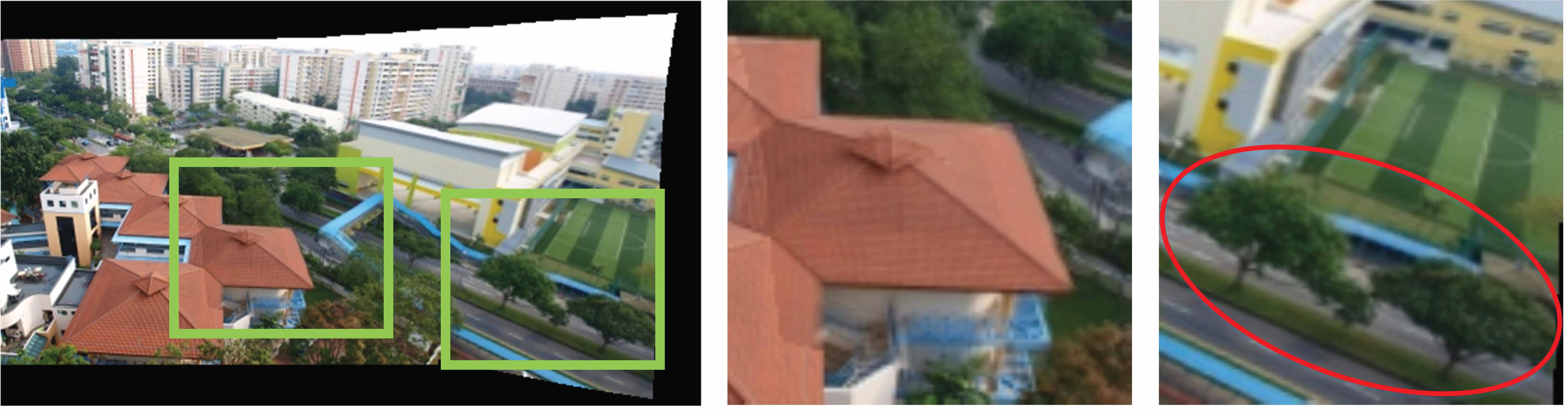}\\[1mm]
	\includegraphics[width=0.9\textwidth]{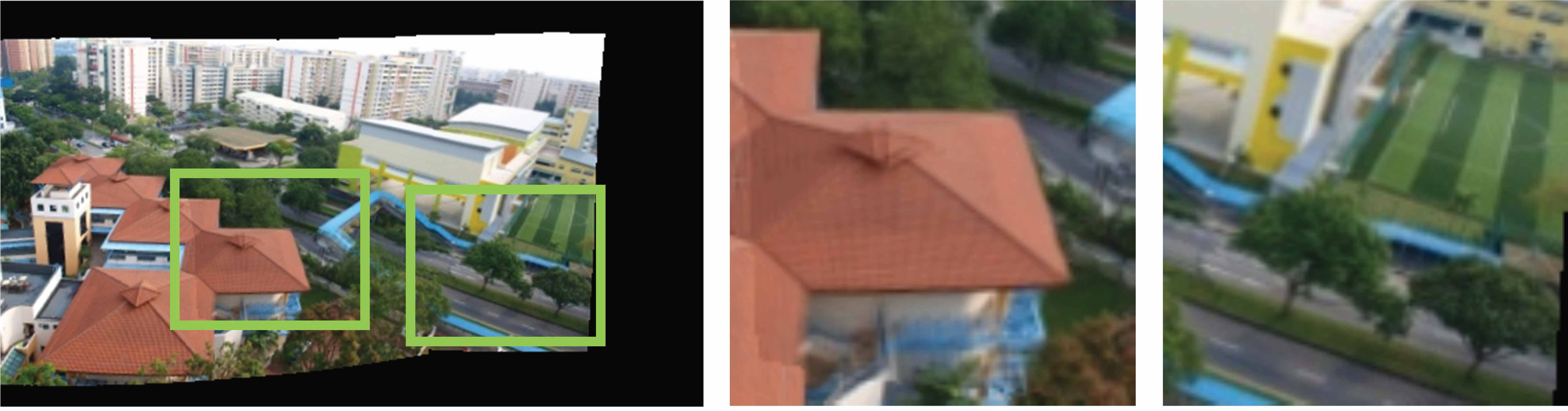}\\[1mm]
	\caption{Comparison with original CPW model~\cite{ZhangLiu2014} on \emph{Rooftops}. From top to bottom, the images show the results of (a) the original CPW, (b) improved CPW, and (c) our method (similarity constraint + improved CPW). Some details are highlighted in the closeup. The red circles indicate errors or distortions.}
	\label{mCPW}
\end{figure*}

Fig.~\ref{mCPW} shows a comparison of the original point-based CPW model~\cite{ZhangLiu2014} and the proposed CPW model on the \emph{Rooftops}\footnote{The \emph{Rooftops} images were acquired from the open dataset of~\cite{Lin2011}.} images. Some errors or distortions are highlighted by the red circles. The stitching process is based on the proposed two-stage alignment. Fig.~\ref{mCPW}(a) shows the results from the original CPW model, in which misalignments are obvious, especially on the rooftops (red circle). Additionally, the roadside trees are stretched. Under the constraints of line features, Fig.~\ref{mCPW}(b) improves the alignment performance and produces more accurate results. As can be seen, line features provide a better geometric description than do point features alone, and the line features function as strong constraints for image stitching. Fig.~\ref{mCPW}(c) shows the final stitching results. Due to the global similarity constraint, the distortions around the roadside trees are largely mitigated, and the proposed method achieves a satisfactory stitching result. Table~\ref{mCPWtab} shows the quantitative comparison. The improved CPW model largely reduces the alignment errors (mainly line errors and total error). By using the similarity constraint, the proposed method obtains a lower \emph{Cor}.

\begin{table}[!t]
	\centering
	\caption{Comparison with original CPW model on \emph{Rooftops}}
	\label{mCPWtab}
	\begin{tabular}{ccccc}
		\hline
		Methods       &{\textit{Cor}}   &{${Err}_{mg}^{(p)}$} & {${Err}_{mg}^{(l)}$} & {$Err_{mg}$} \\ \hline
		Original CPW  & 6.831           & 0.825               & 1.187               & 1.043   \\
		Improved CPW  & 6.390           & 0.973               & 0.491               & 0.682   \\
		Proposed      & 4.903           & 0.967               & 0.492               & 0.681   \\ \hline
	\end{tabular}
\end{table}

Next, we compared the proposed method with other flexible warping methods to evaluate the alignment performance, namely, global homography (baseline)~\cite{Brown2007}, CPW (using global warping for the initial alignment)~\cite{Liu2009}, and APAP~\cite{Zaragoza2014}. For completeness, the proposed method is also compared with the Image Composite Editor (ICE)~\cite{ICE2015} (a common commercial tool for image stitching) by inputting two images at once. For ICE, we used the final post-processed results for the comparison because the original alignment results are not obtainable in the standard version of ICE. In addition, no quantitative comparison of ICE is provided.

\begin{figure*}[htp!]
	\centering
	\includegraphics[width=0.24\textwidth]{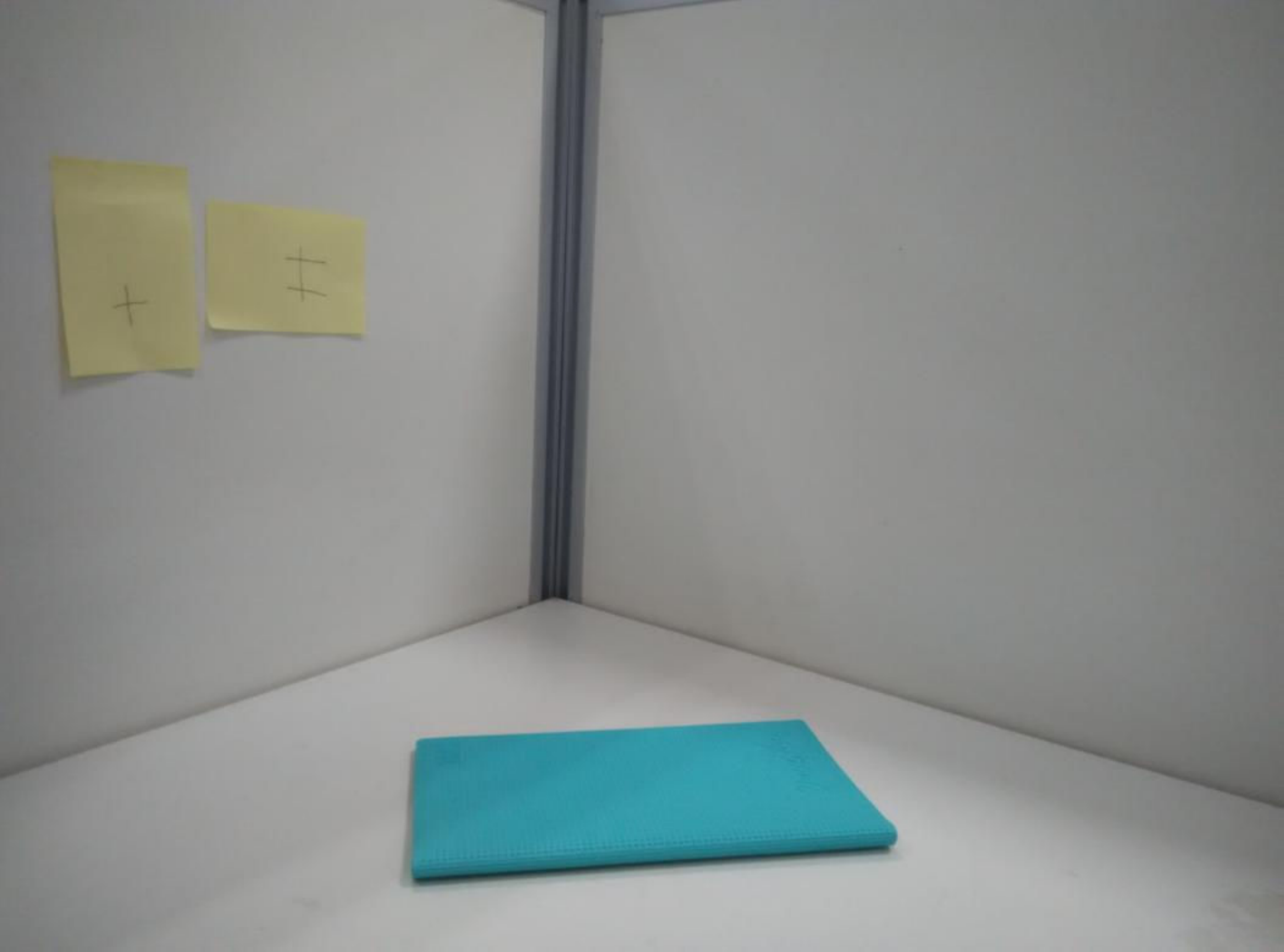}\
	\includegraphics[width=0.24\textwidth]{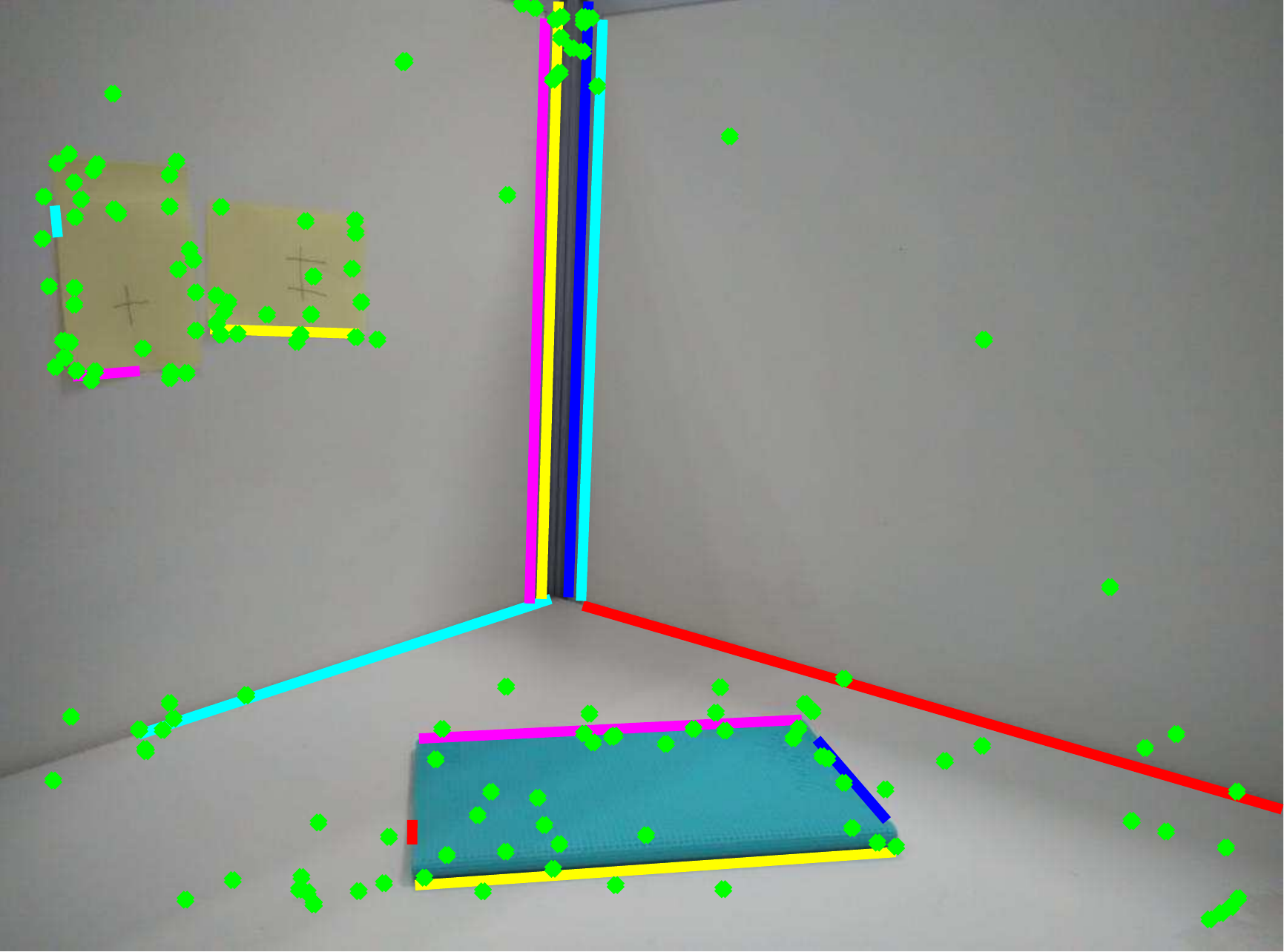}
	\includegraphics[width=0.24\textwidth]{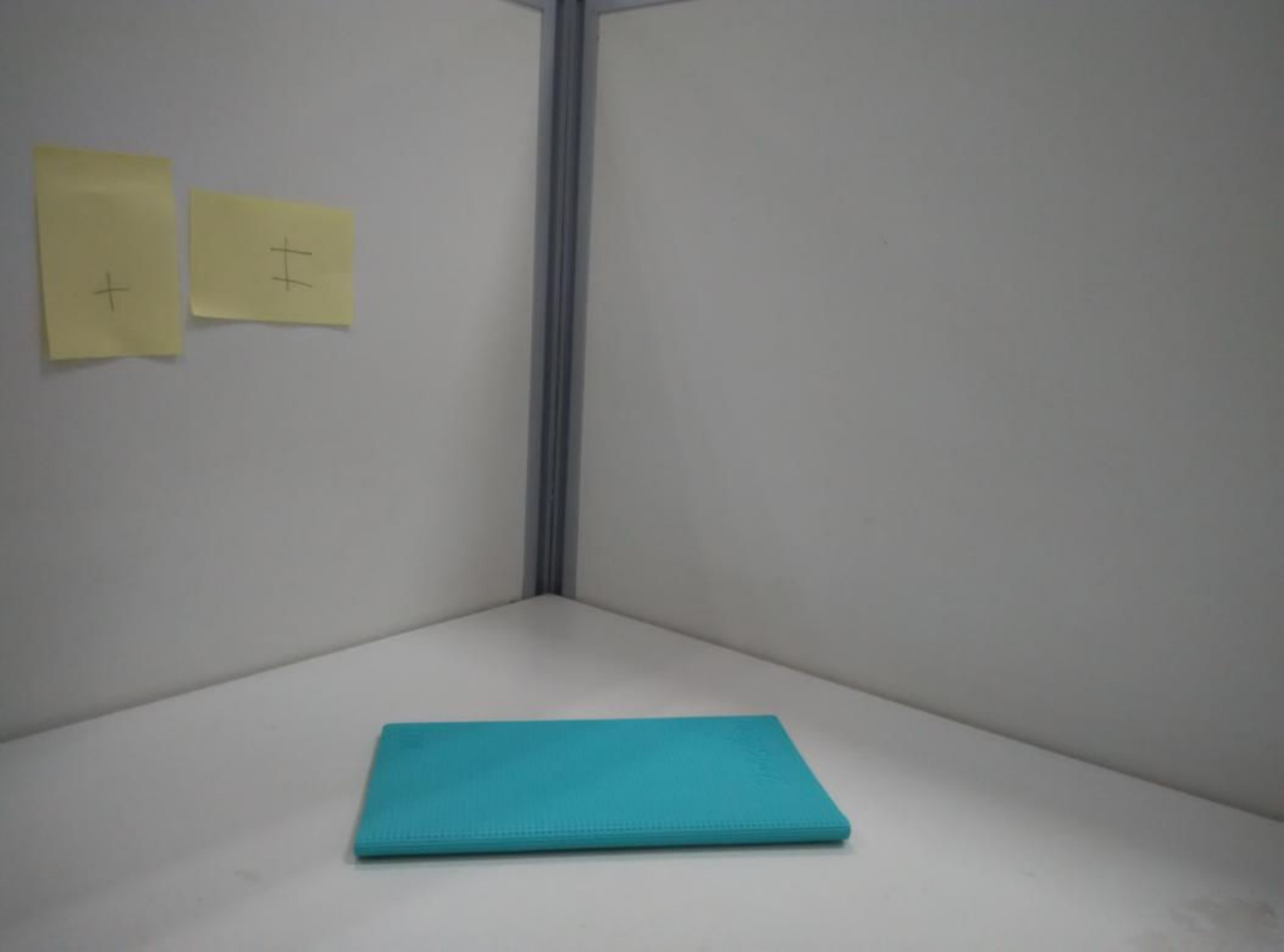}\
	\includegraphics[width=0.24\textwidth]{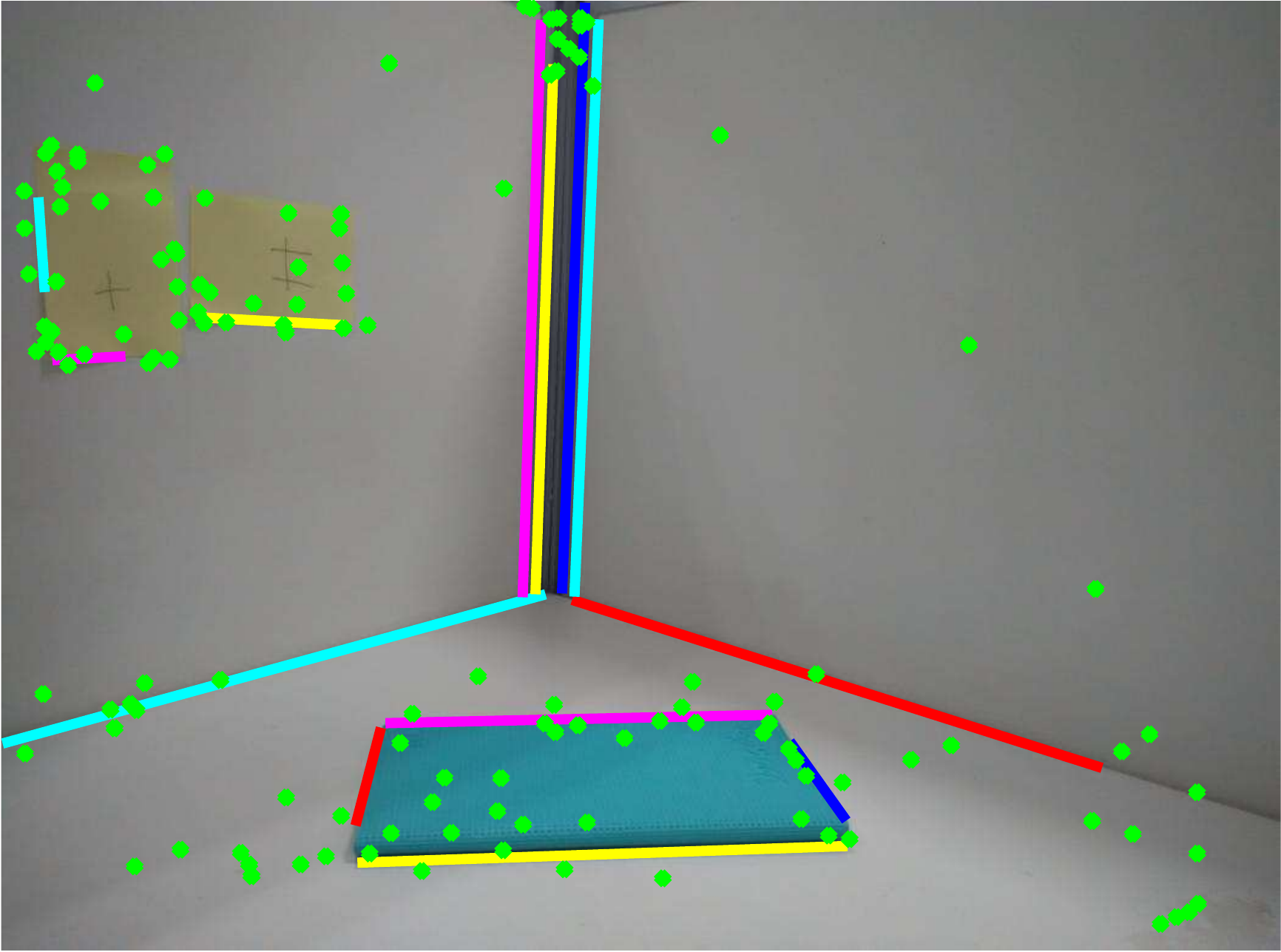}
	\caption{The \emph{Desk} image pair for the assessment of image alignment.}
	\label{fig_alignori}
\end{figure*}

Fig.~\ref{fig_alignori} shows the \emph{Desk} image pair and the detected feature. For most of the low-textured areas, the keypoints are difficult to extract, resulting in insufficient matching points for the estimation of warping model. However, line features can be used as an effective complement for alignment purposes.

\begin{figure*}[htp!]
	\centering
	\includegraphics[width=0.98\textwidth]{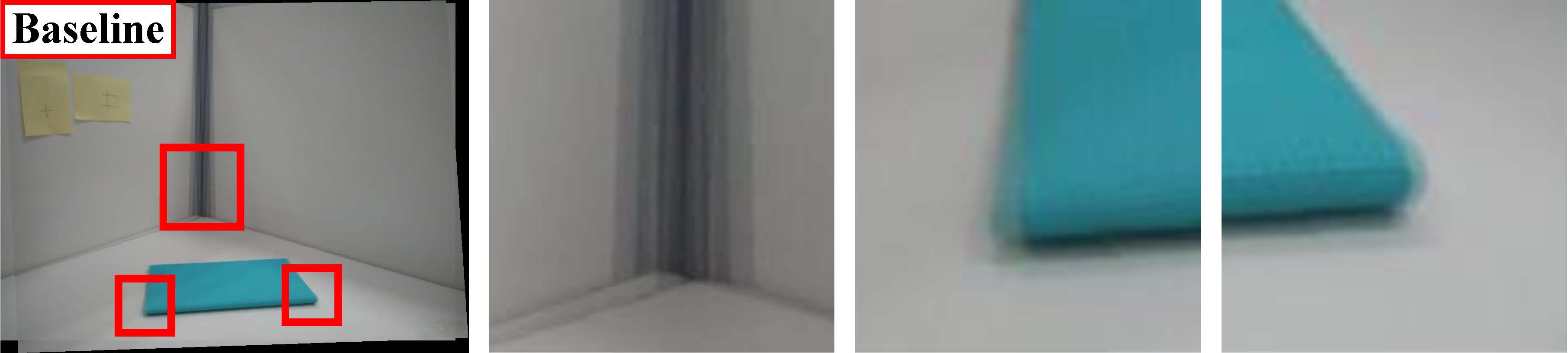}\\[1mm]
	\includegraphics[width=0.98\textwidth]{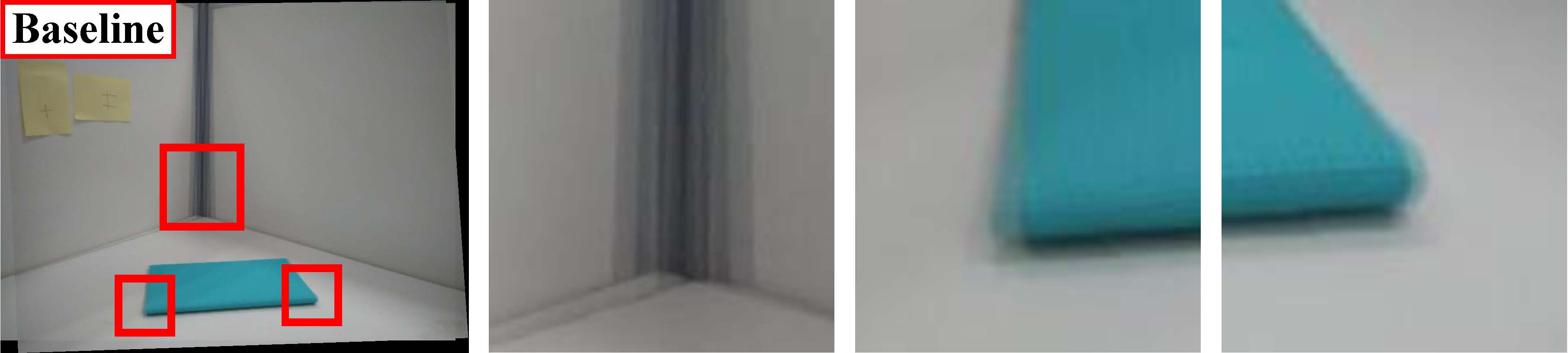}\\[1mm]
	\includegraphics[width=0.98\textwidth]{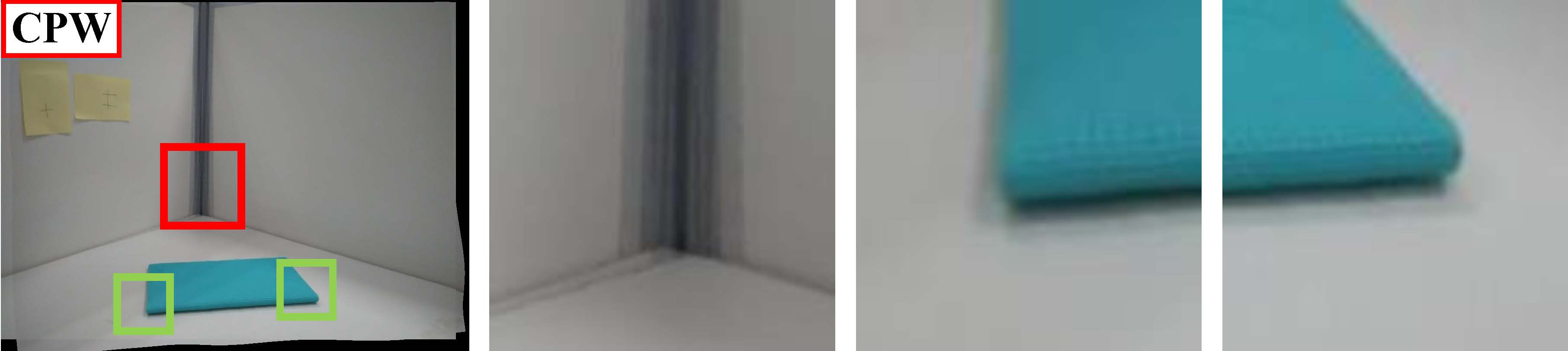}\\[1mm]
	\includegraphics[width=0.98\textwidth]{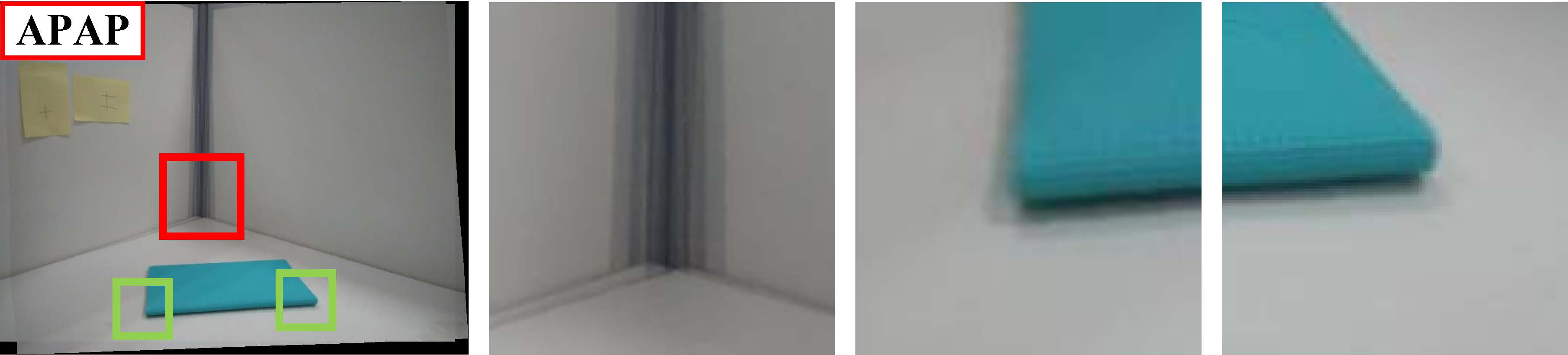}\\[1mm]
	\includegraphics[width=0.98\textwidth]{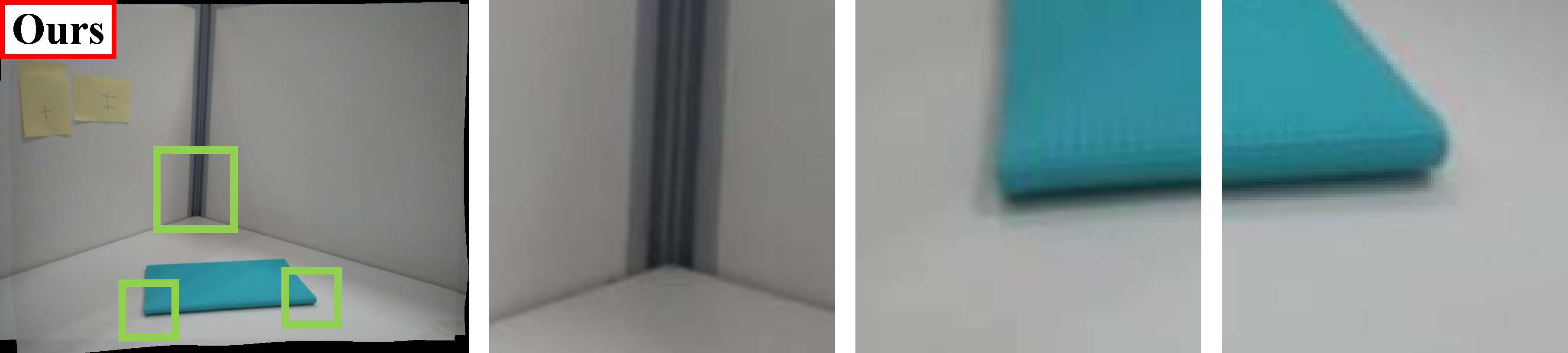}\\[1mm]
	\caption{Comparison of image alignment for stitching on the \emph{Desk} image pair. From top to bottom, the results are (a) Baseline~\cite{Brown2007}, (b) ICE~\cite{ICE2015}, (c) CPW~\cite{Liu2009}, (d) APAP~\cite{Zaragoza2014}, and (e) our method. Some details are highlighted to simplify the comparison. The red boxes show errors, and the green boxes show satisfactory stitching.}
	\label{fig_alignexperi}
\end{figure*}

The comparison results are shown in Fig.~\ref{fig_alignexperi}. Because the images violate the assumptions, the baseline warp is unable to align them properly; it produces obvious misalignments (see the red boxes in Fig.~\ref{fig_alignexperi} (a)). ICE, CPW, and APAP provide relatively better stitching results, but a non-negligible number of ghost artifacts remain. In Fig.~\ref{fig_alignexperi}(b), although ICE uses blending and pixel selection to conceal the misalignments, the post-processing is clearly not completely successful; for instance, there are obvious misalignments on the vertical edge of the desk. Due to an insufficient number of corresponding keypoints along the vertical edge of the desk, CPW and APAP cannot provide an accurate warping model for image alignment; consequently, ghosting occurs in these regions (see the red boxes in Fig.~\ref{fig_alignexperi} (c) and (d)). With the help of line correspondences and the two-stage robust alignment scheme, our method results in satisfactory stitching performance with accurate alignment and few ghost artifacts (Fig.~\ref{fig_alignexperi}(e)). Note that our method also reduces the need for post-processing.

Table~\ref{aligntab} depicts the \emph{Cor} and $Err_{mg}$ values of the compared methods on the \emph{Desk} image pair. As listed, CPW's stronger constraint on point correspondences results in a smaller alignment error on point ${Err}_{mg}^{(p)}$; however, the alignment errors on line ${Err}_{mg}^{(l)}$ and $Err_{mg}$ remain large. The proposed method reduces the geometric error and results in better accuracy than do the other tested methods.

\begin{table}[htp!]
	\centering
	\caption{Comparison of alignment on \emph{Desk}}
	\label{aligntab}
	\begin{tabular}{ccccc}
		\hline
		Methods   &{\textit{Cor}} &{${Err}_{mg}^{(p)}$} & {${Err}_{mg}^{(l)}$} & {$Err_{mg}$} \\ \hline
		Baseline  & 0.390         & 4.894             & 5.632               & 5.001   \\
		CPW       & 0.299         & \textbf{1.534}    & 3.703               & 1.849   \\
		APAP      & 0.360         & 2.652             & 4.407               & 2.907   \\
		Proposed  & \textbf{0.169}  & 1.562           & \textbf{0.594}      & \textbf{1.422} \\ \hline
	\end{tabular}
\end{table}

\subsection{Distortion reduction} 

\begin{figure*}[htp!]
	\centering
	\includegraphics[width=0.9\textwidth]{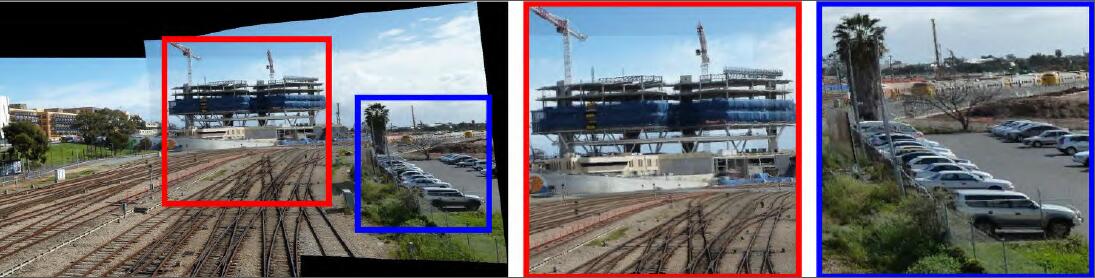} \\[1mm]
	\includegraphics[width=0.9\textwidth]{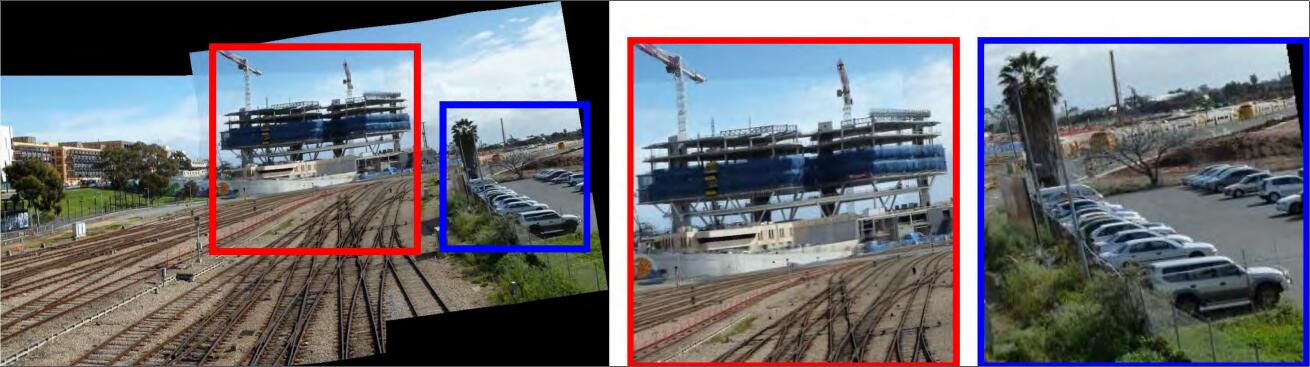}\\[1mm]
	\includegraphics[width=0.9\textwidth]{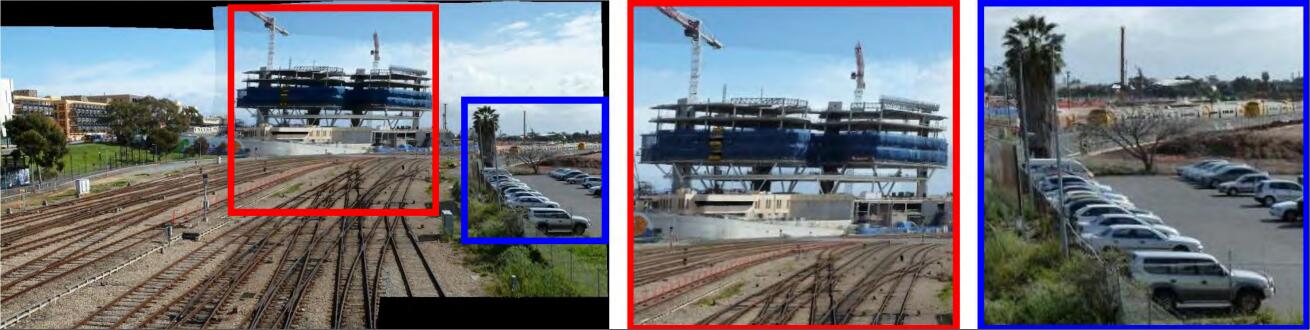} \\[1mm]
	\includegraphics[width=0.9\textwidth]{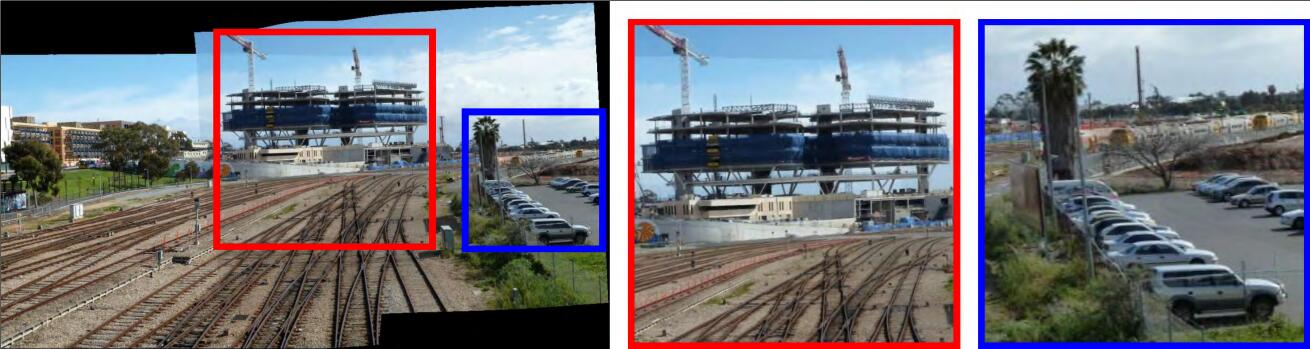}
	\caption{Comparison with SPHP method on distortion reduction. From top to bottom, the results are: (a) APAP~\cite{Zaragoza2014}, (b) SPHP0~\cite{Chang2014}, (c) SPHP1~\cite{Chang2014}, (d) our method. For better comparison, some details are highlighted.}
	\label{fig_raildis} 
\end{figure*}

To investigate the distortion reduction performance, SPHP~\cite{Chang2014} and AANAP~\cite{Lin2015} were compared with the proposed method on the \emph{Railtracks} and \emph{Temple Square} image pairs\footnote{The \emph{Railtracks} and \emph{Temple Square} images were acquired from the open dataset of~\cite{Zaragoza2014}.}.

Fig.~\ref{fig_raildis} shows the stitching results of the four methods, APAP~\cite{Zaragoza2014}, SPHP~\cite{Chang2014}, SPHP with an assumption of no rotation (SPHP1)~\cite{Chang2014}, and our method. Due to its simple extrapolation of projective transformation to non-overlapping regions, the APAP (Fig.~\ref{fig_raildis}(a)) result shows projective distortions in the non-overlapping regions. In the blue box in the closeup, the car is enlarged, and the palm tree is obviously slanted. By introducing the similarity transformation, SPHP can largely mitigate these projective distortions. In Fig.~\ref{fig_raildis}(b), SPHP preserves the shape of the car, but it has a problem with the unnatural rotation. In addition, the construction site (in the red box) is tilted to the left. In contrast, SPHP1 preserves the shape and reduces the perspective distortion, but the construction site is now tilted slightly to the right (\ref{fig_raildis}(c)). Using the global similarity constraint, the proposed method largely eliminates all these distortions, providing a pleasing stitching result, as is clearly shown in \ref{fig_raildis}(d).

\begin{figure*}[!t]
	\centering
	\includegraphics[width=0.9\textwidth]{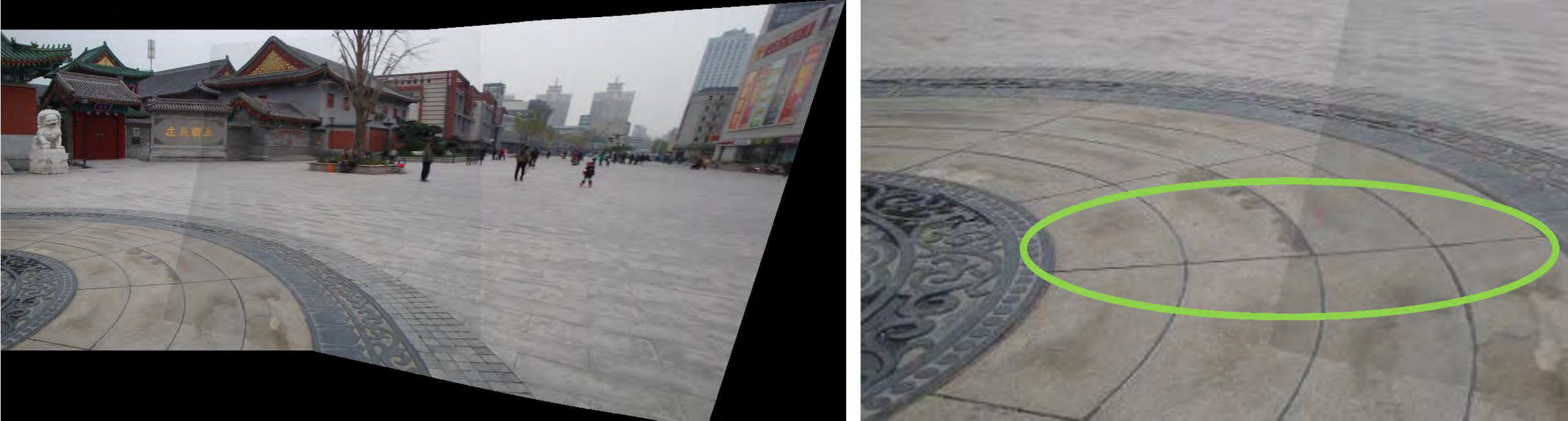}\\[1mm]
	\includegraphics[width=0.9\textwidth]{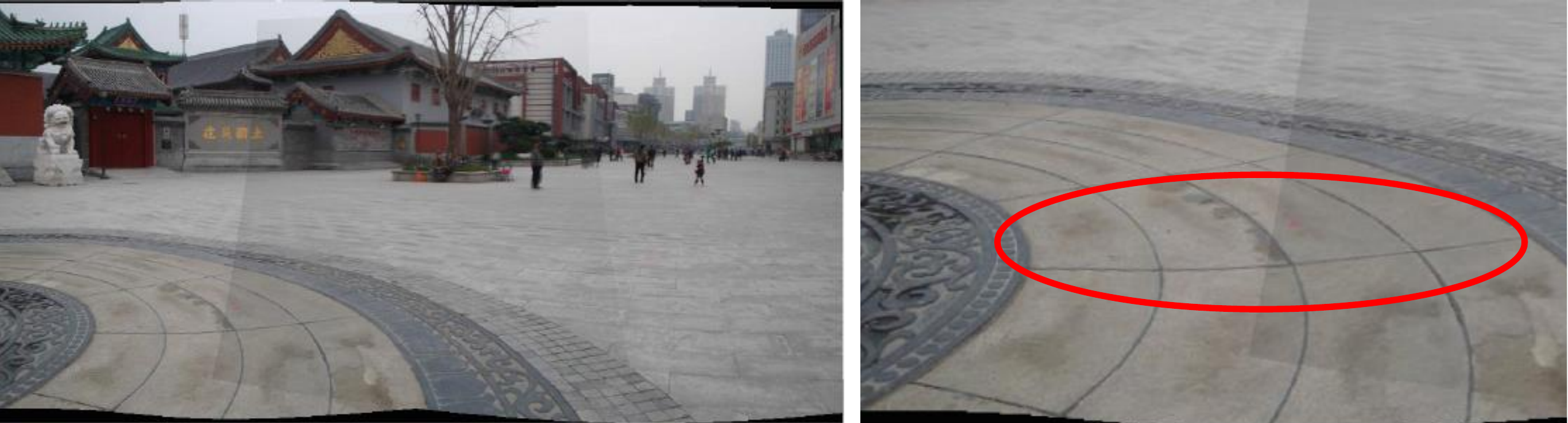}\\[1mm]
	\includegraphics[width=0.9\textwidth]{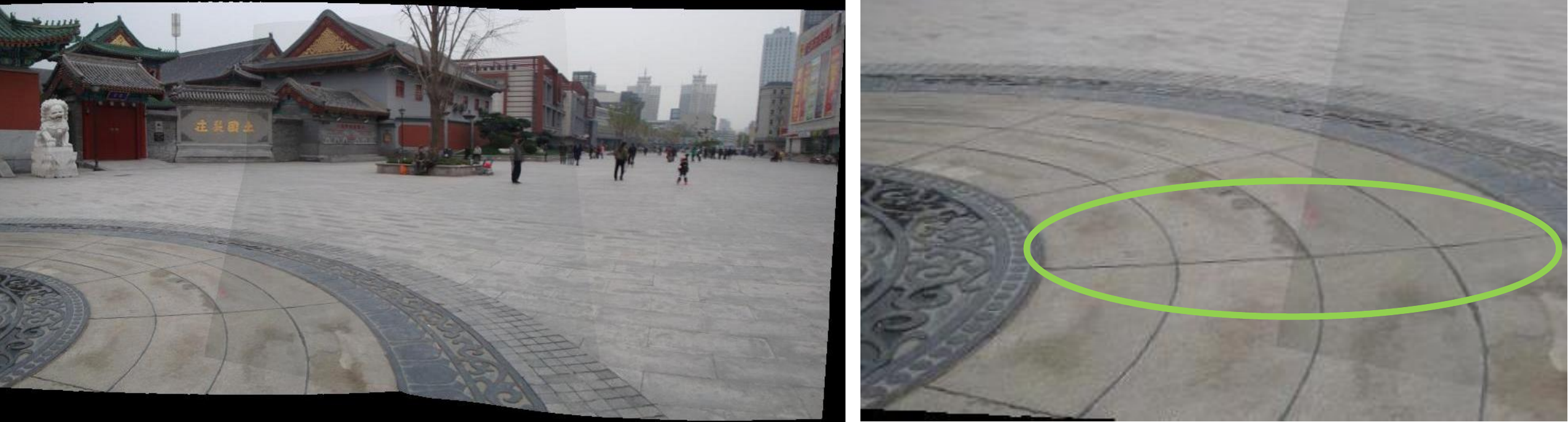}\\[1mm]
	\caption{Comparison with AANAP method on distortion reduction. From top to bottom, the results are: (a) APAP~\cite{Zaragoza2014}, (b) AANAP~\cite{Lin2015}, (c) our method. For better comparison, some details are highlighted. The red circle showes the distortion.}
	\label{fig_templedis}
\end{figure*}

Fig.~\ref{fig_templedis} shows a comparison of the proposed method with AANAP~\cite{Lin2015} on distortion reduction. Fig.~\ref{fig_templedis}(a) shows that APAP achieves good alignment, but it suffers from shape and perspective distortion problems, for example, in the stretched and tilted buildings at the right of the image. By linearizing the homography and using the similarity transformation, AANAP provides an attractive result in which the projective distortions have been largely mitigated (Fig.~\ref{fig_templedis}(b)). However, as shown in the red circle of the enlarged view, the lines on the ground are slightly deformed. Our method yields more appealing stitching results in this example (Fig.~\ref{fig_templedis}(c)).

\subsection{Comparisons with global-based methods}

\begin{figure*}[htp!]
	\centering
	\includegraphics[width=0.24\textwidth]{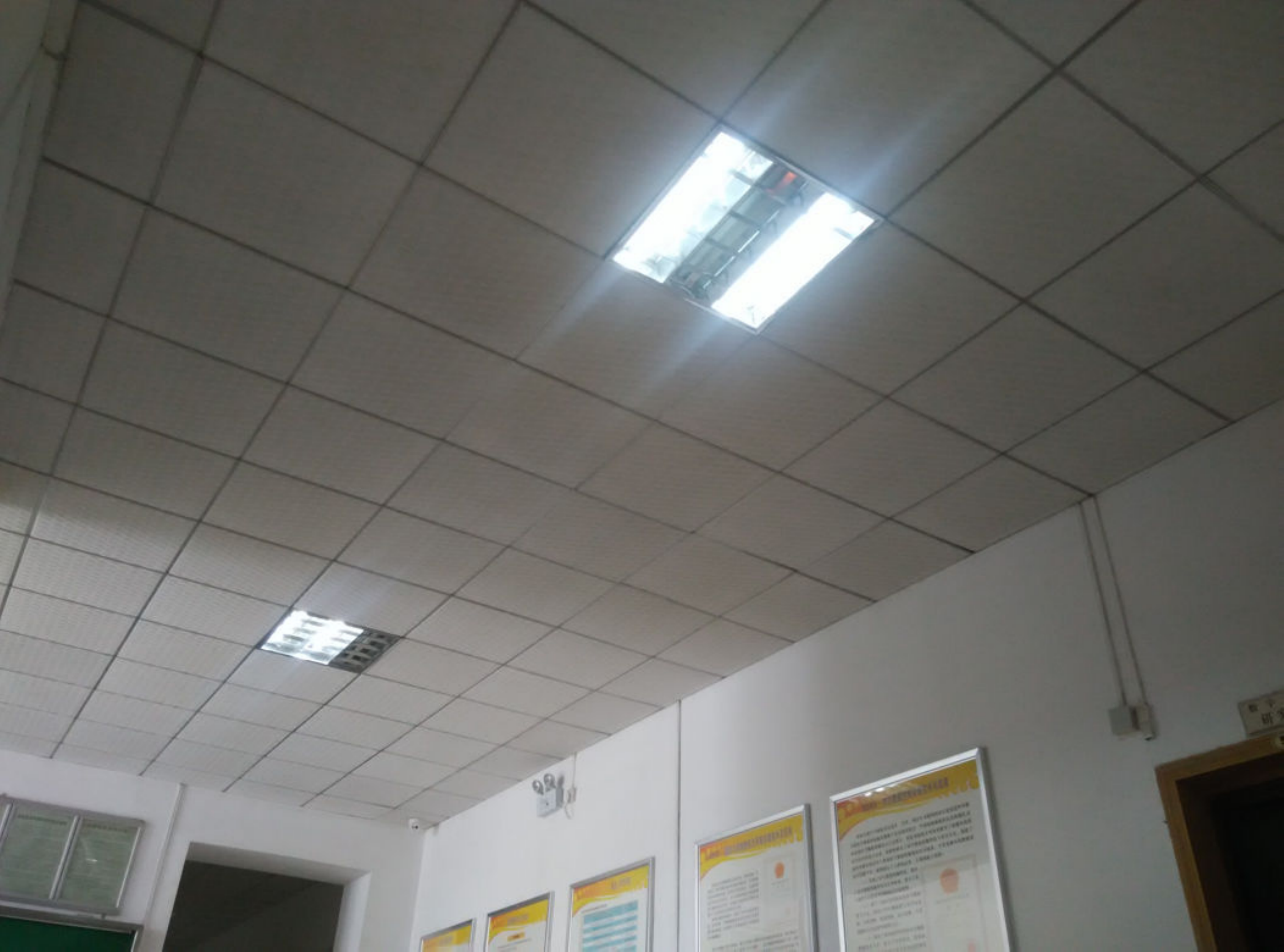}\
	\includegraphics[width=0.24\textwidth]{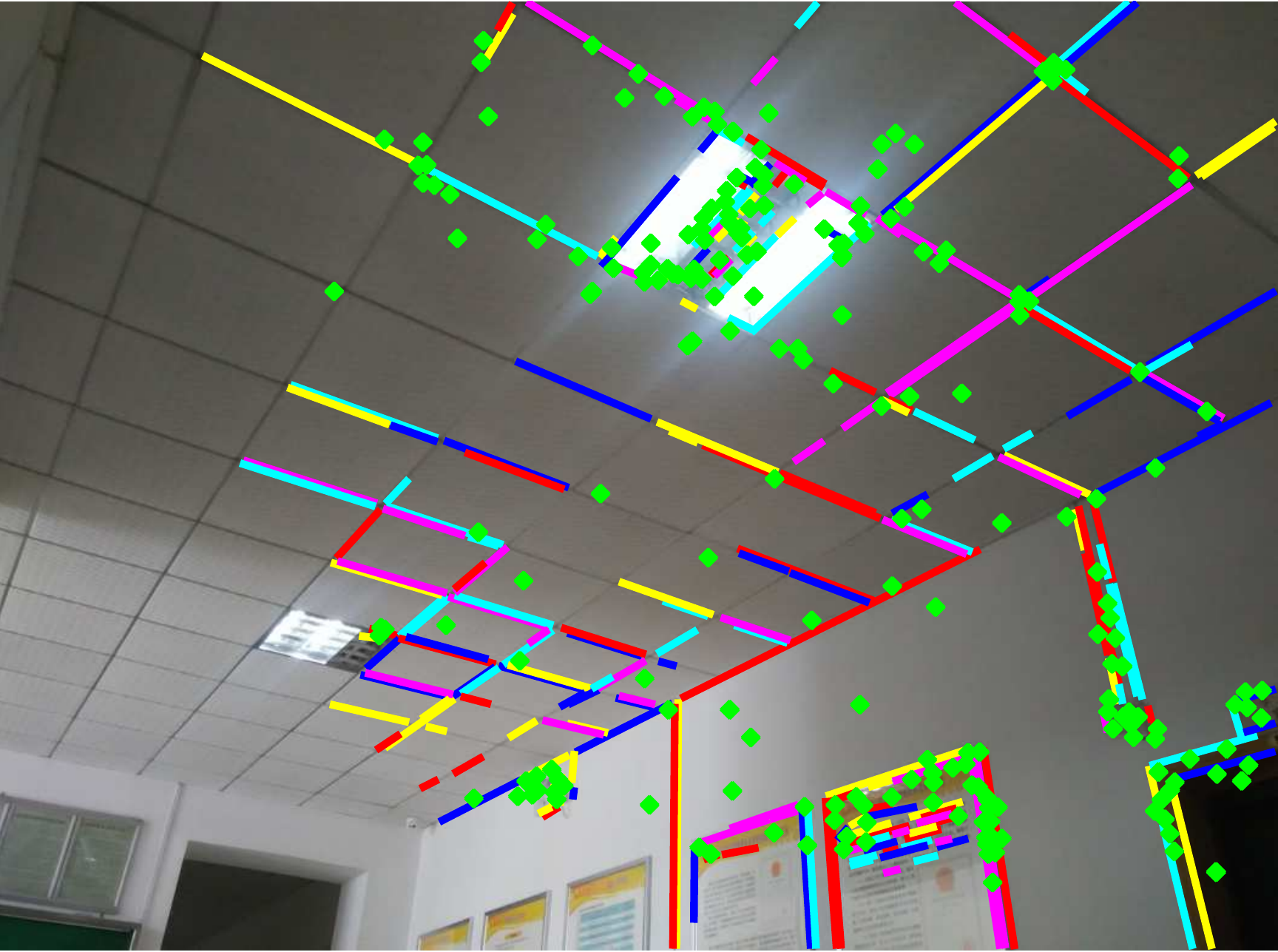}\
	\includegraphics[width=0.24\textwidth]{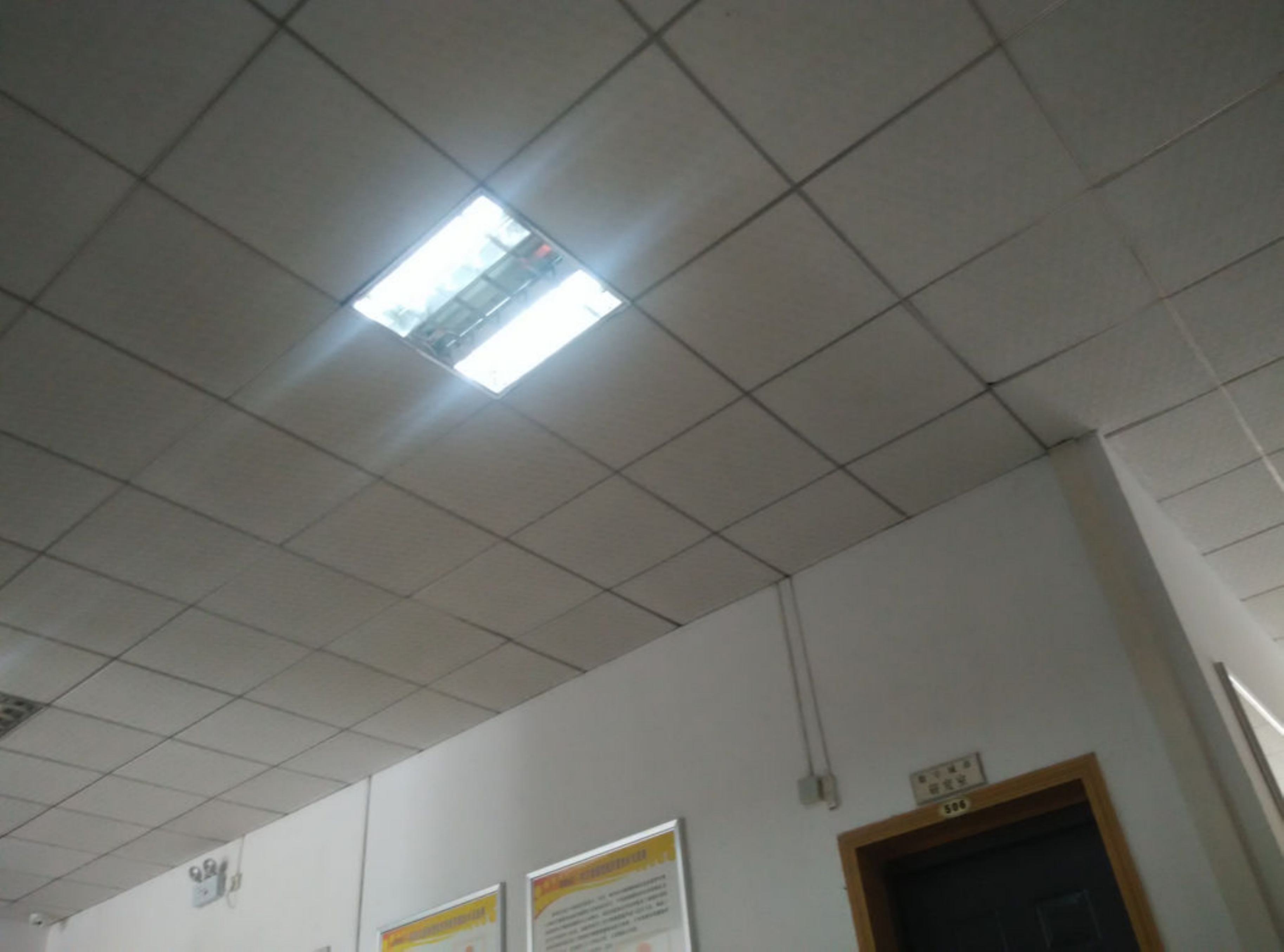}\
	\includegraphics[width=0.24\textwidth]{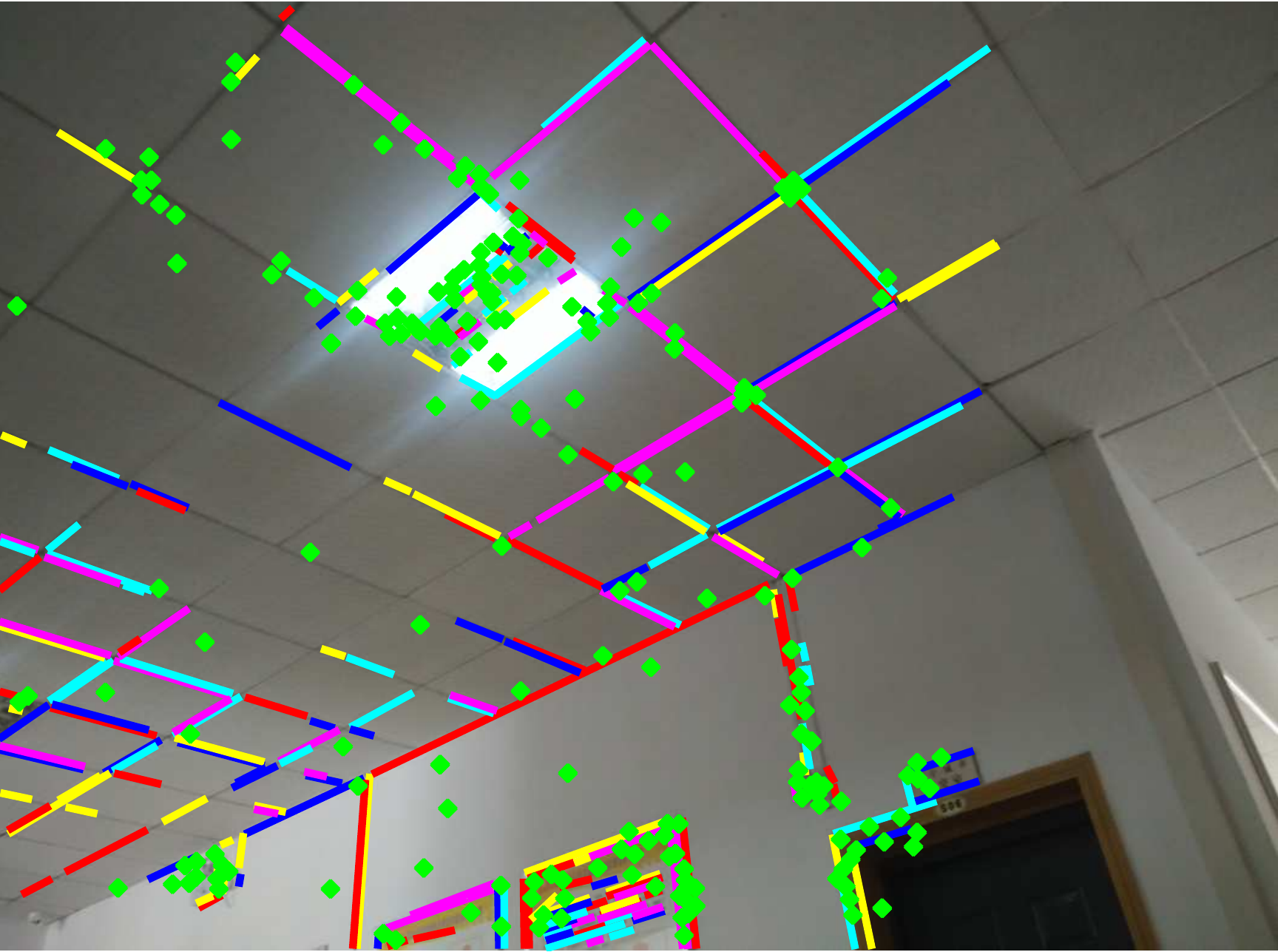}\\[1mm]
	\includegraphics[width=0.24\textwidth]{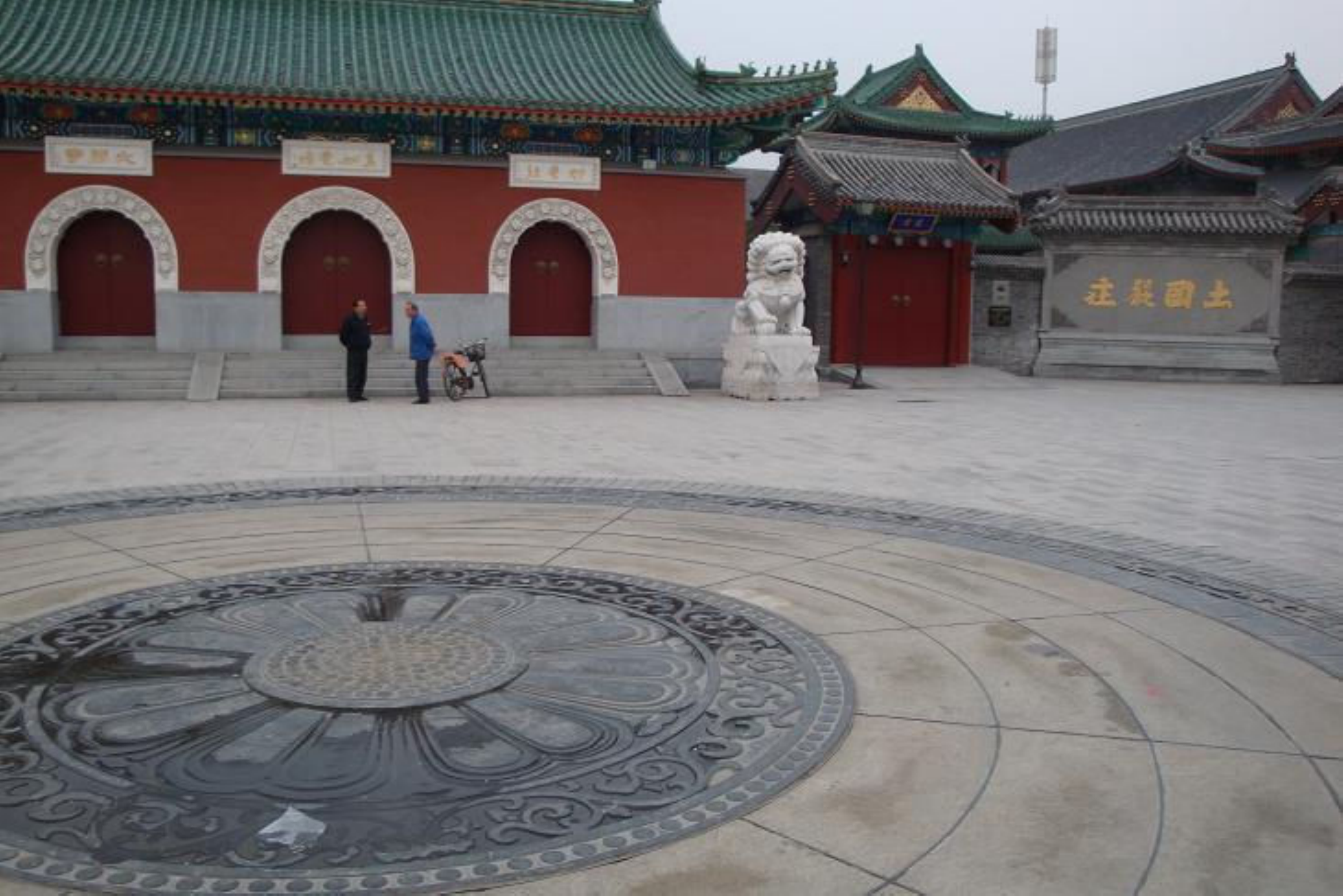}\
	\includegraphics[width=0.24\textwidth]{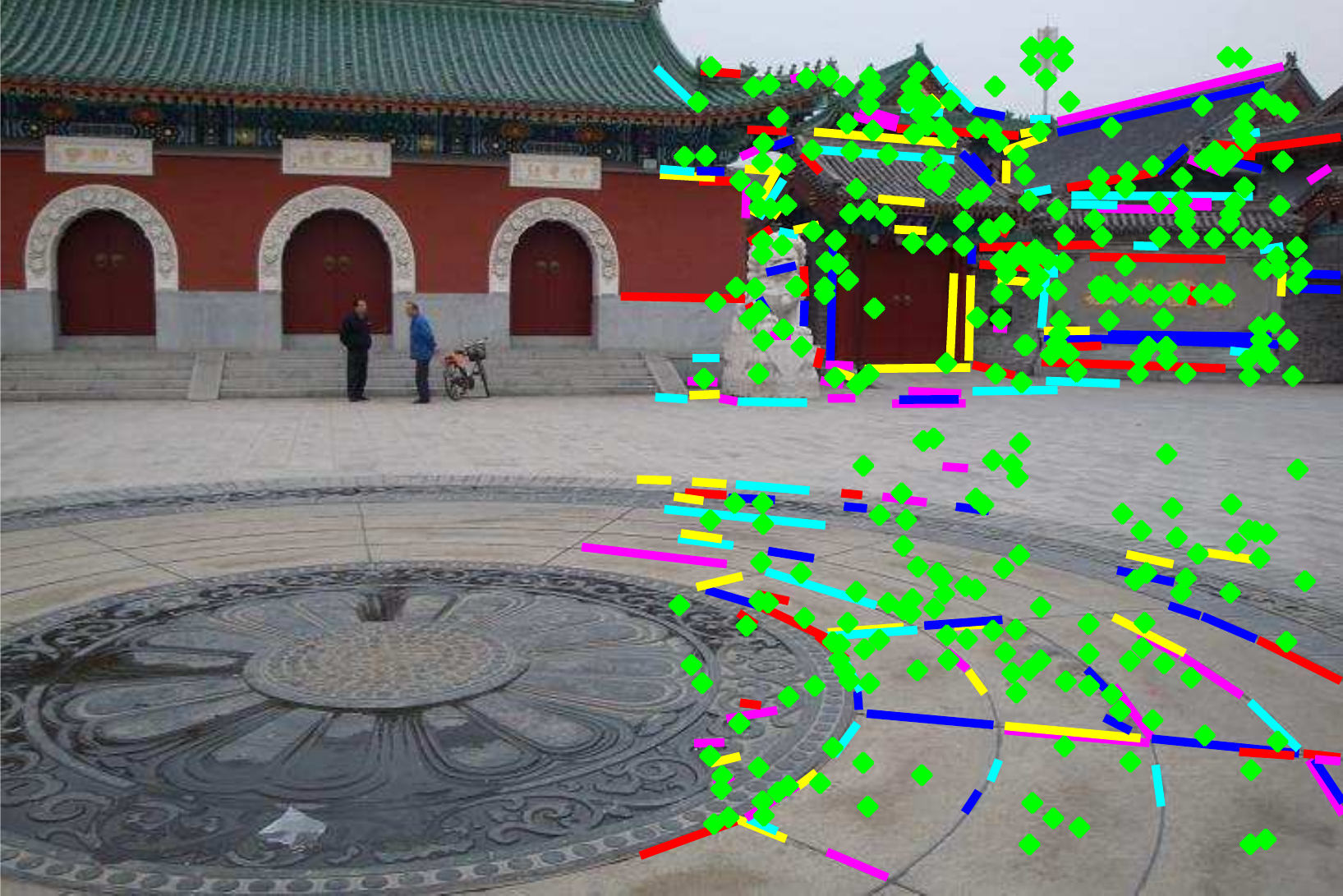}\
	\includegraphics[width=0.24\textwidth]{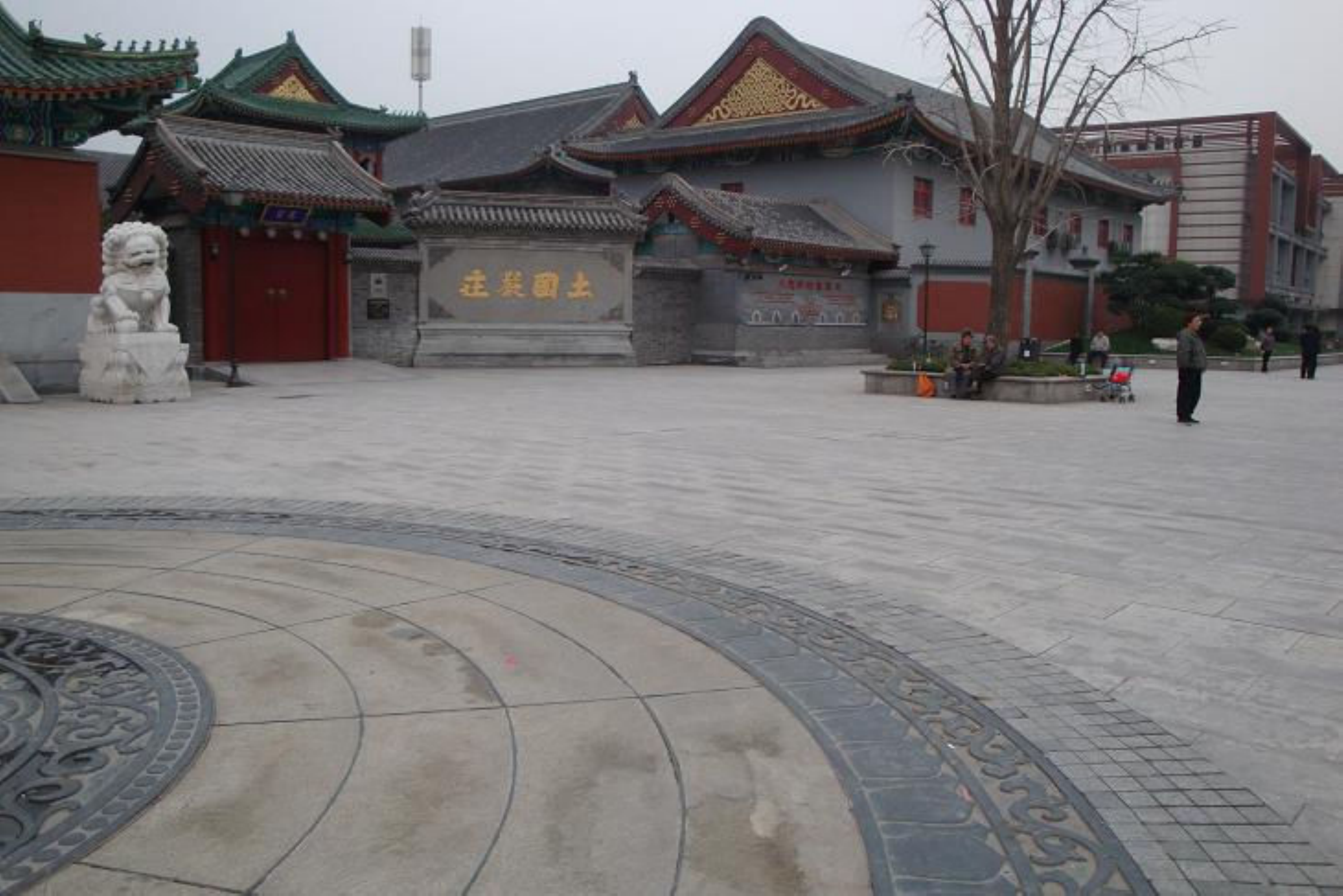}\
	\includegraphics[width=0.24\textwidth]{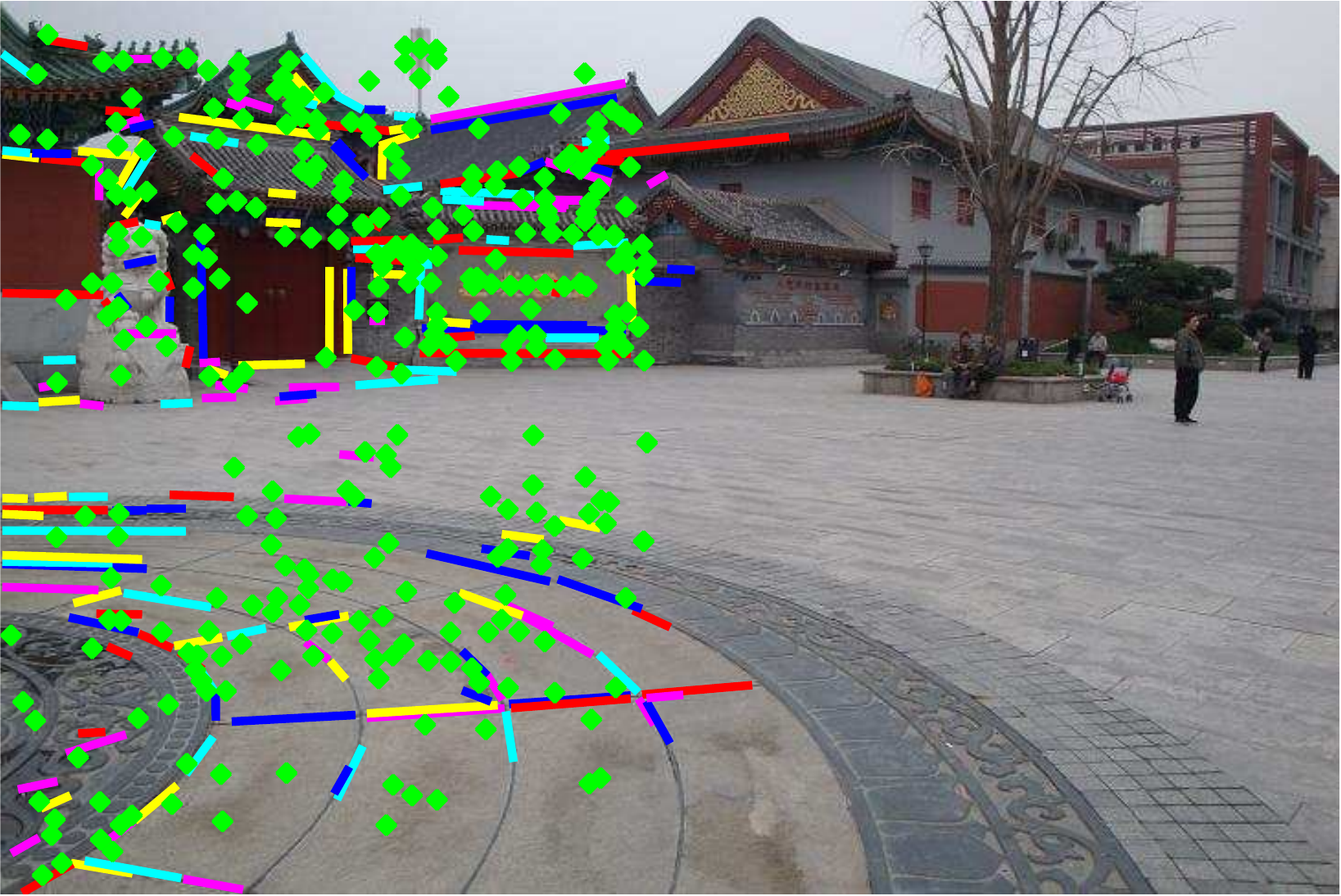}\\
	\caption{The original images for global-based methods. Top:~\emph{Ceiling}, bottom:~\emph{Temple}.}
	\label{fig_globalorig}
\end{figure*}

\begin{figure*}[htp!]
	\centering
	\includegraphics[width=0.98\textwidth]{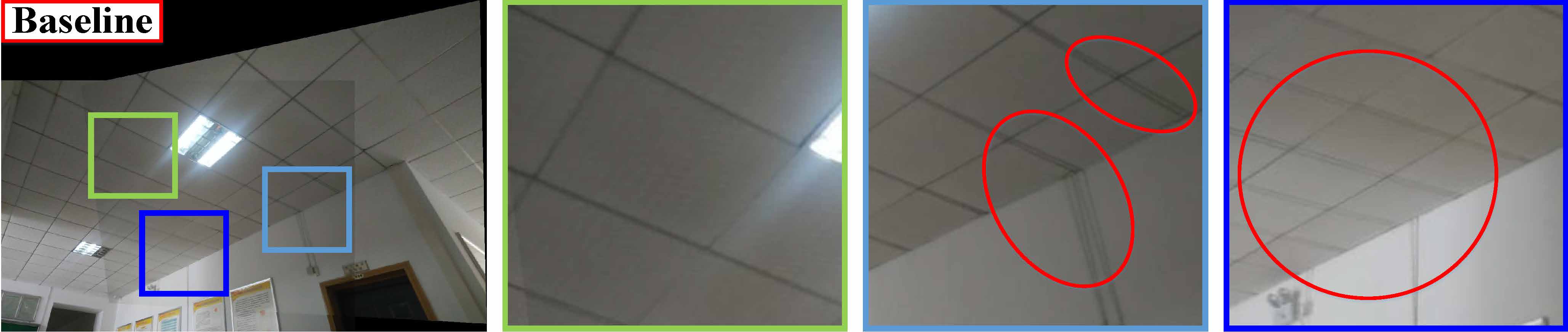}\\[1mm]
	\includegraphics[width=0.98\textwidth]{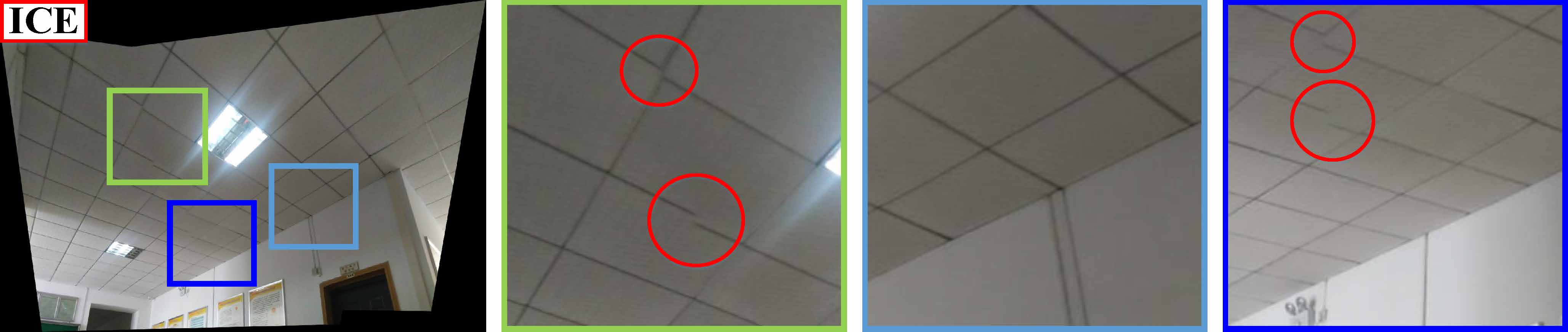}\\[1mm]
	\includegraphics[width=0.98\textwidth]{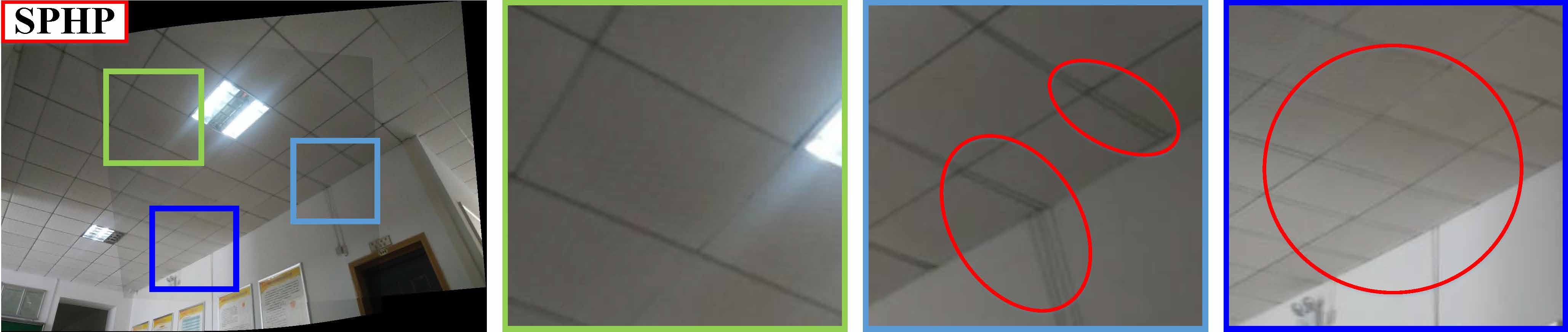}\\[1mm]
	\includegraphics[width=0.98\textwidth]{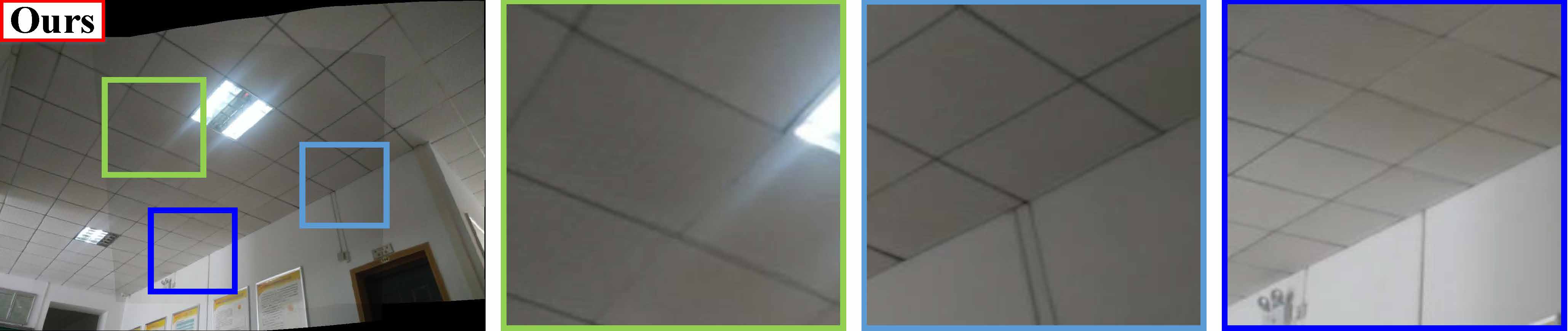}\\
	\caption{Comparison of image stitching on~\emph{Ceiling}. From top to bottom, the results are: (a) Baseline~\cite{Brown2007}, (b) ICE~\cite{ICE2015}, (c) SPHP~\cite{Chang2014}, (d) our method (global version). For better comparison, some details are highlighted. The red circles show alignment errors.}
	\label{fig_globalexperi1}
\end{figure*}

\begin{figure*}[htp!]
	\centering
	\includegraphics[width=0.98\textwidth]{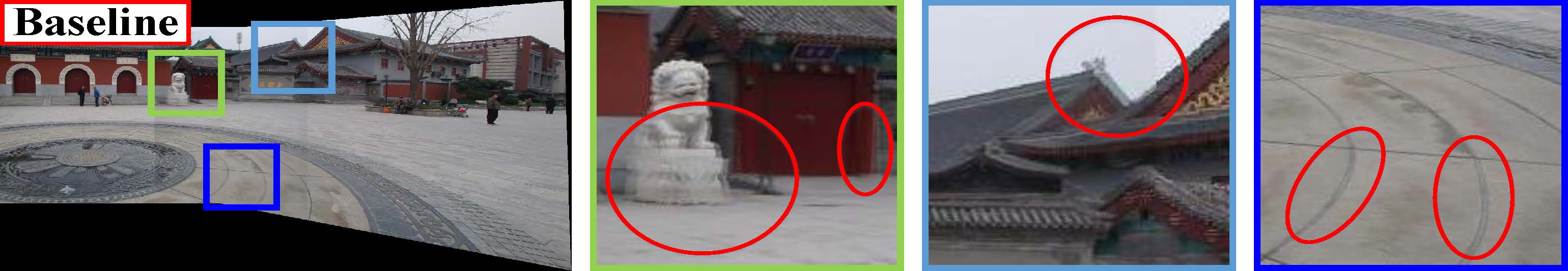}\\[1mm]
	\includegraphics[width=0.98\textwidth]{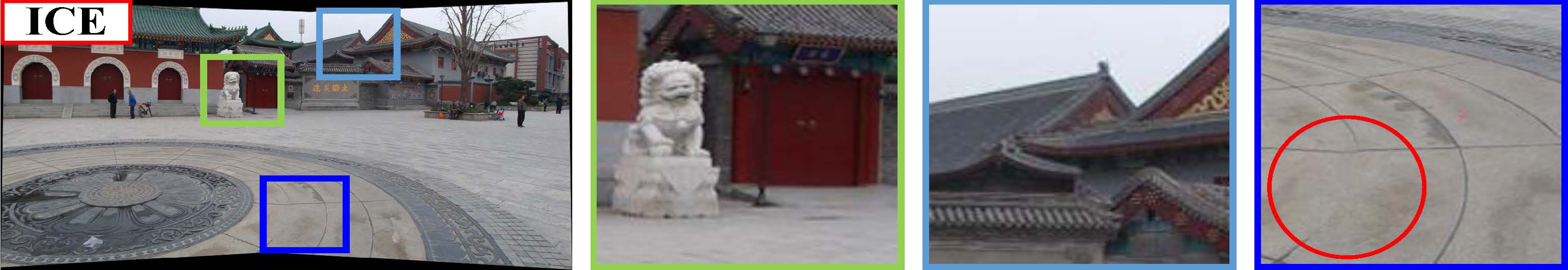}\\[1mm]
	\includegraphics[width=0.98\textwidth]{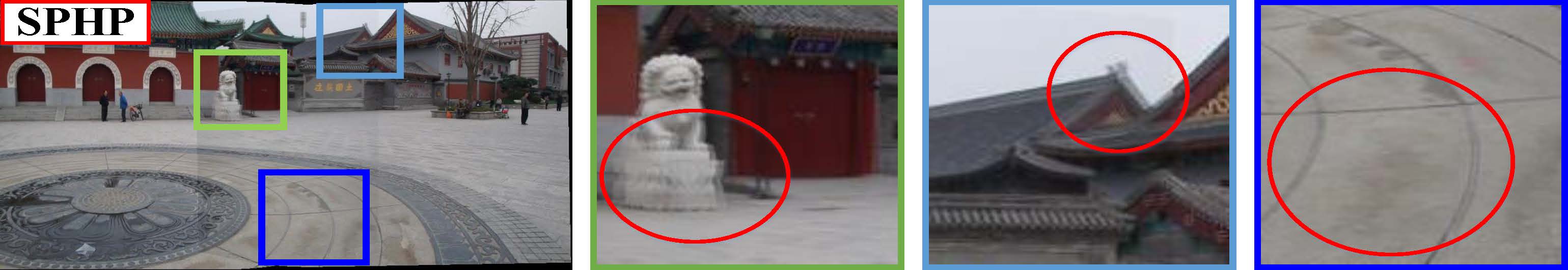}\\[1mm]
	\includegraphics[width=0.98\textwidth]{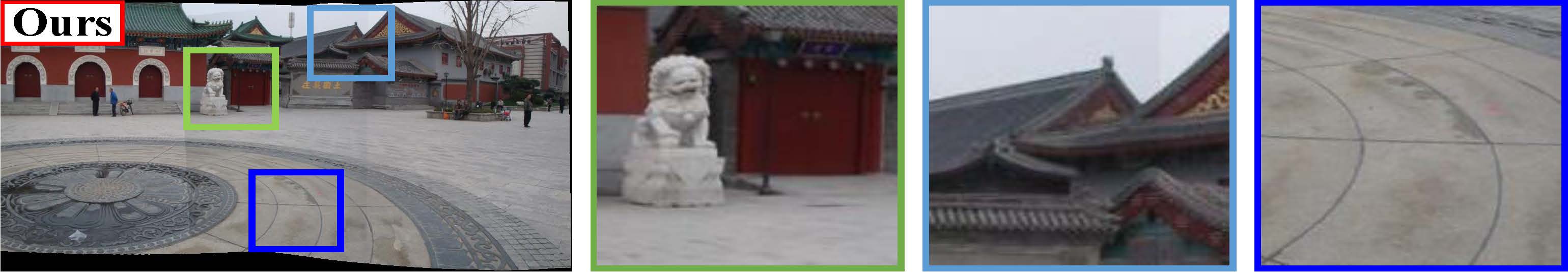}
	\caption{Comparison of image stitching on~\emph{Temple}. From top to bottom, the results are: (a) Baseline~\cite{Brown2007}, (b) ICE~\cite{ICE2015}, (c) SPHP~\cite{Chang2014}, and (d) our method (global version). For better comparison, some details are highlighted. The red circles show alignment errors.}
	\label{fig_globalexperi2}
\end{figure*}

\begin{table}[htp!]
	\centering
	\caption{Quantitative Evaluation on~\emph{Ceiling}}
	\label{globaltab1}
	\begin{tabular}{ccccc}
		\hline
		Methods  & \textit{Cor}   & ${Err}_{mg}^{(p)}$  & ${Err}_{mg}^{(l)}$  & ${Err}_{mg}$   \\ \hline
		Baseline & 0.755          & 3.200          & 2.059          & 2.452          \\
		SPHP     & 0.631          & 2.876          & 1.989          & 2.292          \\
		Proposed & \textbf{0.200} & \textbf{1.343} & \textbf{0.695} & \textbf{0.921} \\ \hline
	\end{tabular}
\end{table}

\begin{table}[htp!]
	\centering
	\caption{Quantitative Evaluation on~\emph{Temple}}
	\label{globaltab2}
	\begin{tabular}{ccccc}
		\hline
		Methods  & \textit{Cor}   & ${Err}_{mg}^{(p)}$ & ${Err}_{mg}^{(l)}$ & ${Err}_{mg}$  \\ \hline
		Baseline & 6.240          & 1.899          & 0.954          & 1.430          \\
		SPHP     & 4.334          & 1.756          & 0.919          & 1.341          \\
		Proposed & \textbf{1.515} & \textbf{0.592} & \textbf{0.529} & \textbf{0.561} \\ \hline
	\end{tabular}
\end{table}

In this section, the proposed method is compared with three global-based methods: global homography (Baseline)~\cite{Brown2007}, ICE~\cite{ICE2015}, and SPHP~\cite{Chang2014}. For our method (called the global version), global homography is adopted during the first alignment stage and jointly estimated by point and line correspondences to pre-warp the source images.

Fig.~\ref{fig_globalorig} shows the two pairs of original images for stitching: \emph{Ceiling} and \emph{Temple} \footnote{The \emph{Temple} images were acquired from the open dataset of~\cite{Zaragoza2014}.}. The low-textured content of \emph{Ceiling} results in the detection of only a limited number of unevenly distributed  keypoints, which may degrade the warping model's  estimations. However, line correspondences are abundant, which can improve the image alignment. \emph{Temple} provides rich point correspondences, but the scene contains multiple distinct planes, which is a challenge for the global-based methods.

Fig.~\ref{fig_globalexperi1} and \ref{fig_globalexperi2} show the results of the global-based methods on the \emph{Ceiling} and \emph{Temple} image pairs. As shown, due to the model deficiencies, the Baseline warp cannot provide satisfactory stitching results; there are numerous misalignments and projective distortions. The ICE and SPHP methods improve the stitching performance, especially in the aspect of the reduction of projective distortions. For instance, the door in Fig.~\ref{fig_globalexperi1} and the people in Fig. \ref{fig_globalexperi2} have few distortions, but the bricks of the ceiling in the non-overlapping area in the ICE result (Fig.~\ref{fig_globalexperi1}(b)) are slightly stretched. In addition, alignment errors in these two pairs of images (the red circles in Fig.~\ref{fig_globalexperi1} and \ref{fig_globalexperi2}) remain obvious. In contrast, the proposed method is more flexible and robust in handling the alignment not only because of the line-guided warping estimation but also because of the alignment constraints in the mesh-based framework. With the similarity constraint, our method provides good stitching results with minimal distortions.

Table~\ref{globaltab1} and Table~\ref{globaltab2} contains a quantitative comparison of \emph{Ceiling} and \emph{Temple}, showing that our method provides the results with the fewest errors. On \emph{Ceiling}, our method performs the best because the line features play an important role in scenes without reliable keypoint correspondences. On the \emph{Temple} image, which has rich and reliable keypoints, the role of the line feature may be reduced, but it still helps to improve the alignment accuracy.

\subsection{Comparisons with local-based methods}

\begin{figure*}[htp!]
	\centering  
	\includegraphics[width=0.23\linewidth, height=0.2\linewidth]{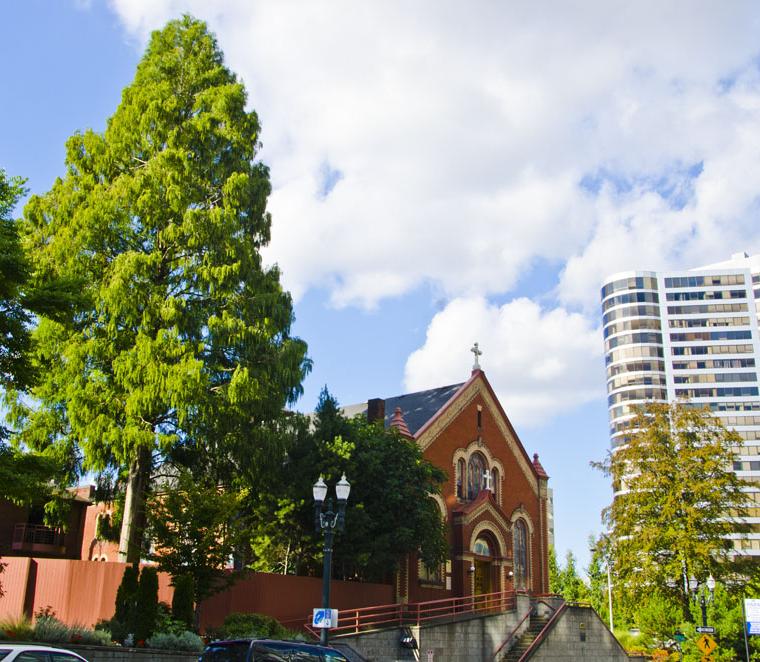}\
	\includegraphics[width=0.23\linewidth, height=0.2\linewidth]{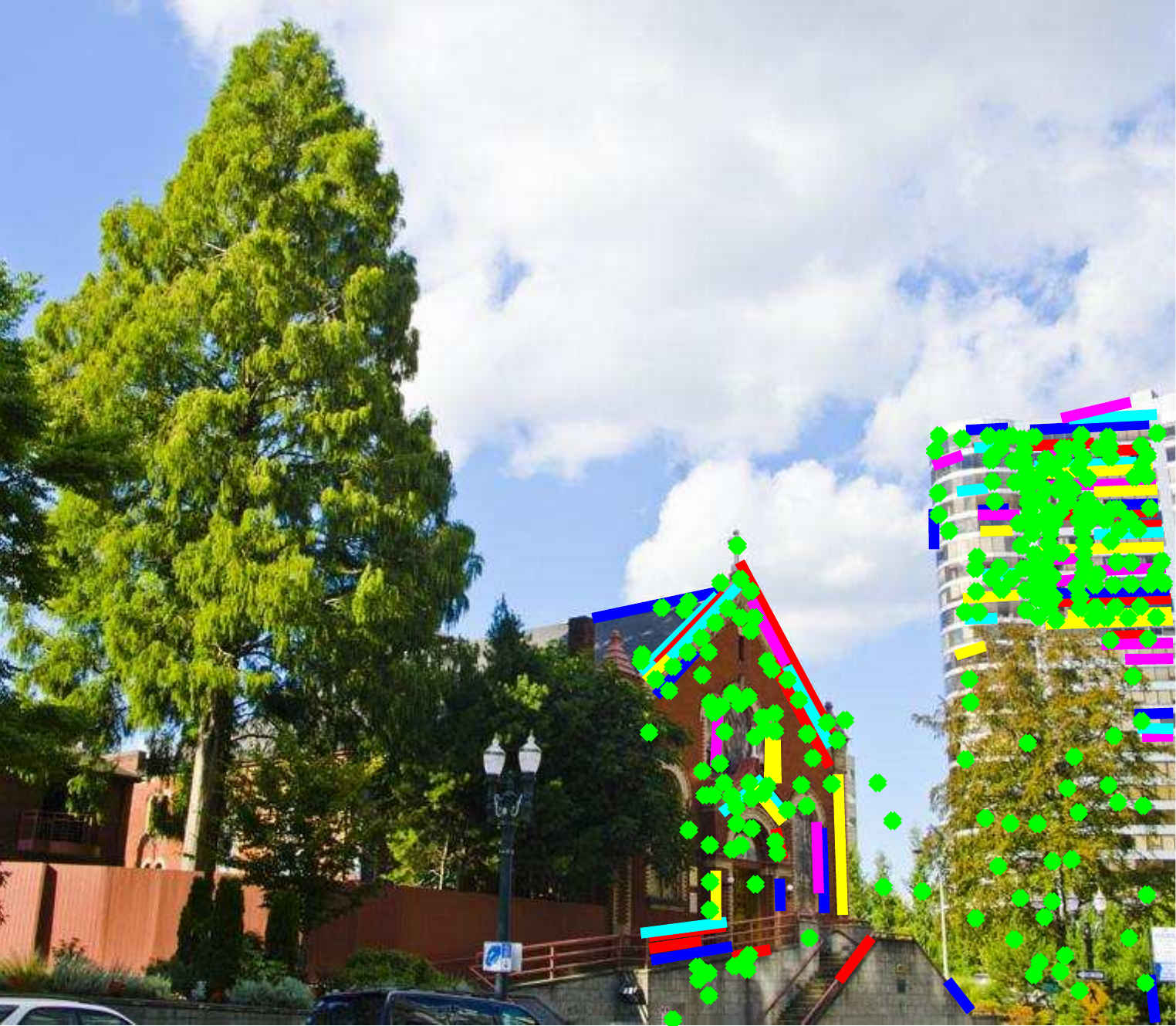}
	\includegraphics[width=0.23\linewidth, height=0.2\linewidth]{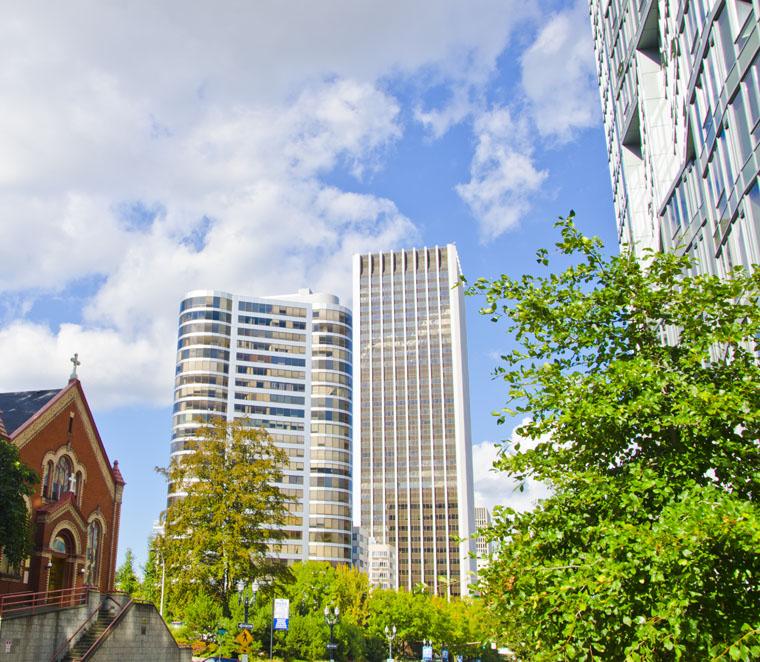}
	\includegraphics[width=0.23\linewidth, height=0.2\linewidth]{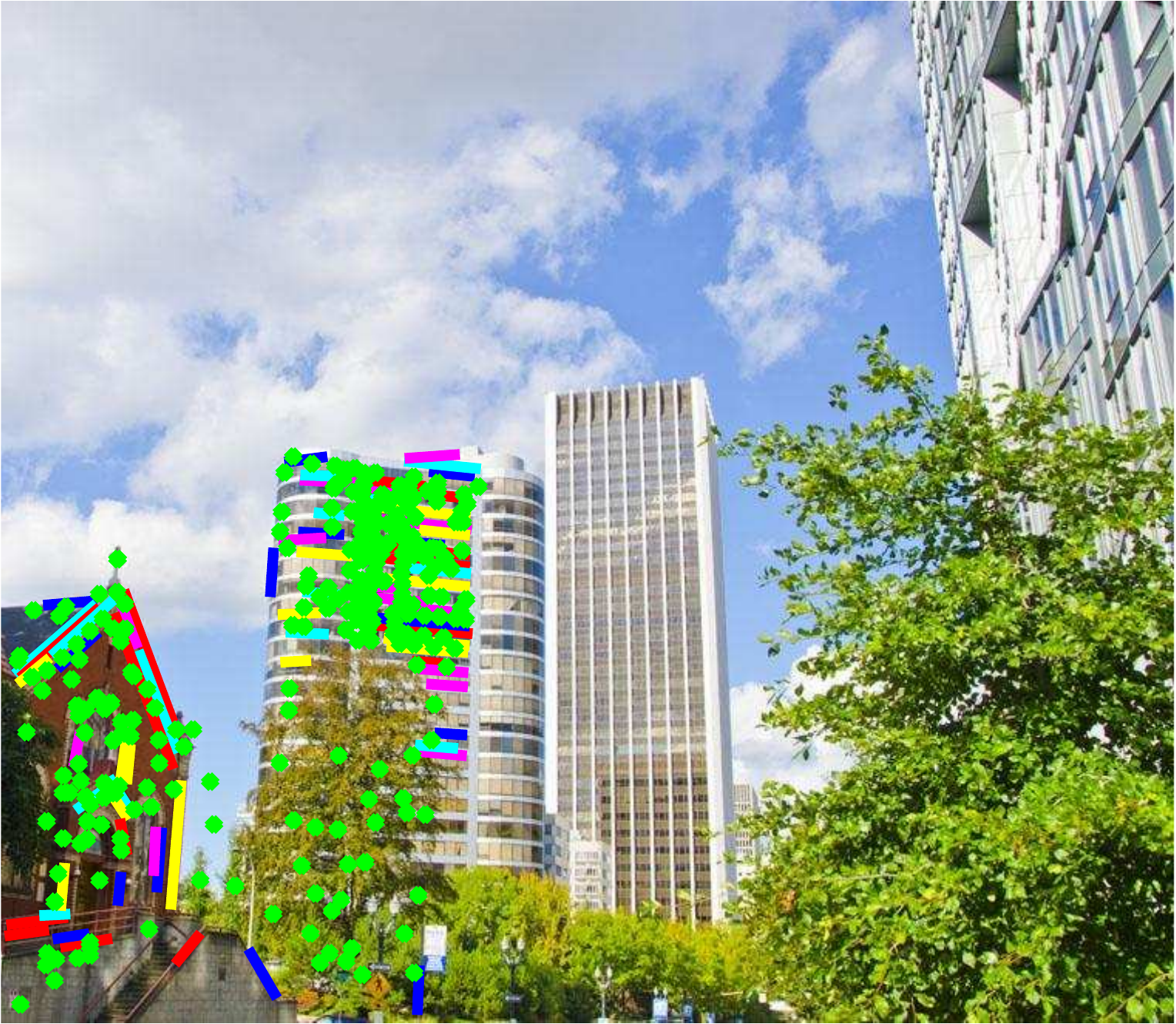}\\[1mm]
	\includegraphics[width=0.23\linewidth, height=0.2\linewidth]{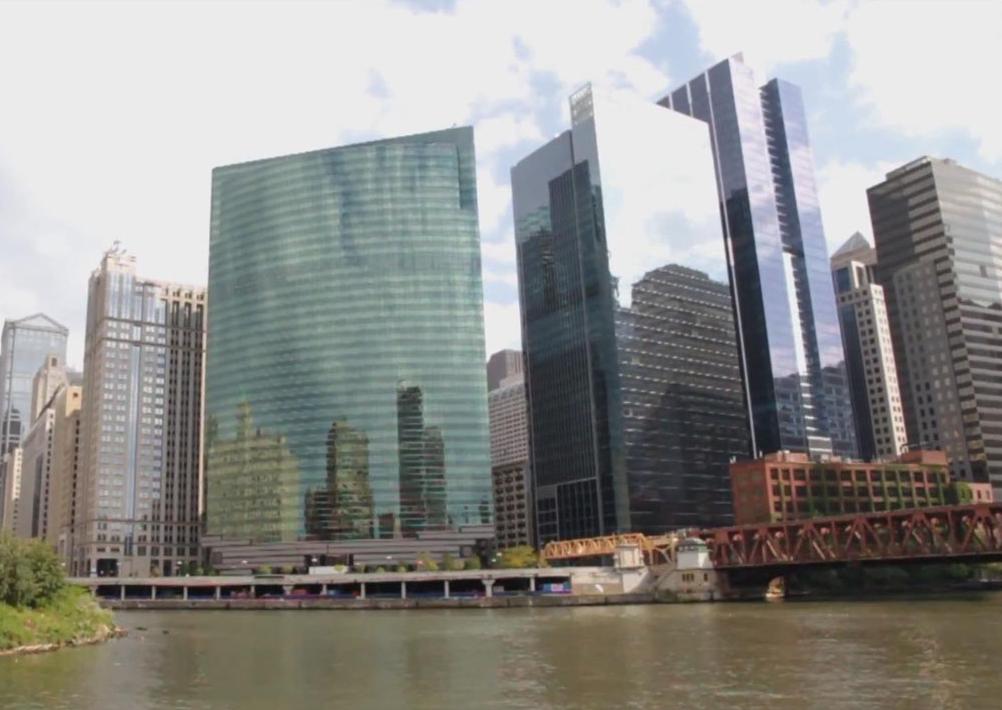}\
	\includegraphics[width=0.23\linewidth, height=0.2\linewidth]{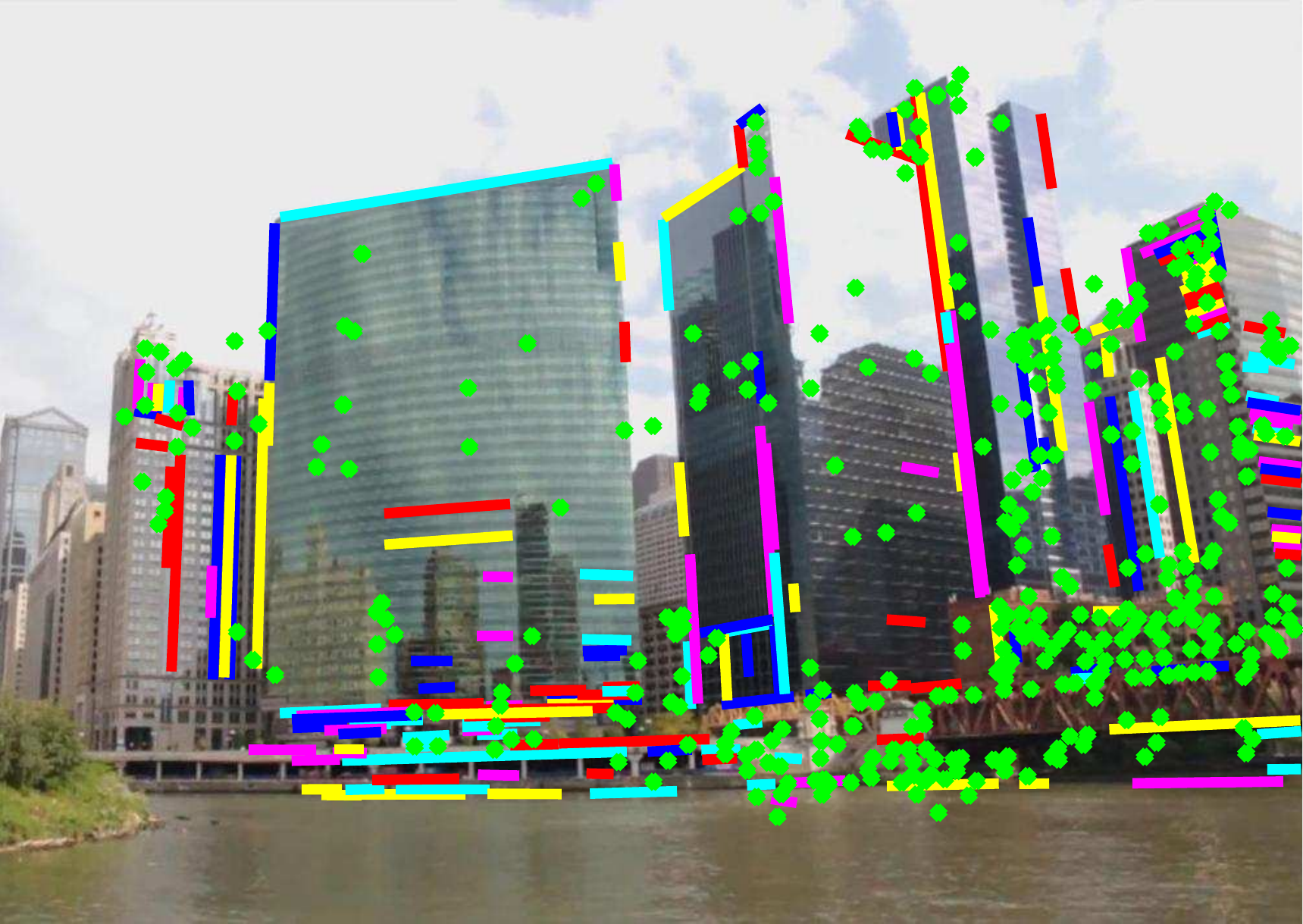}
	\includegraphics[width=0.23\linewidth, height=0.2\linewidth]{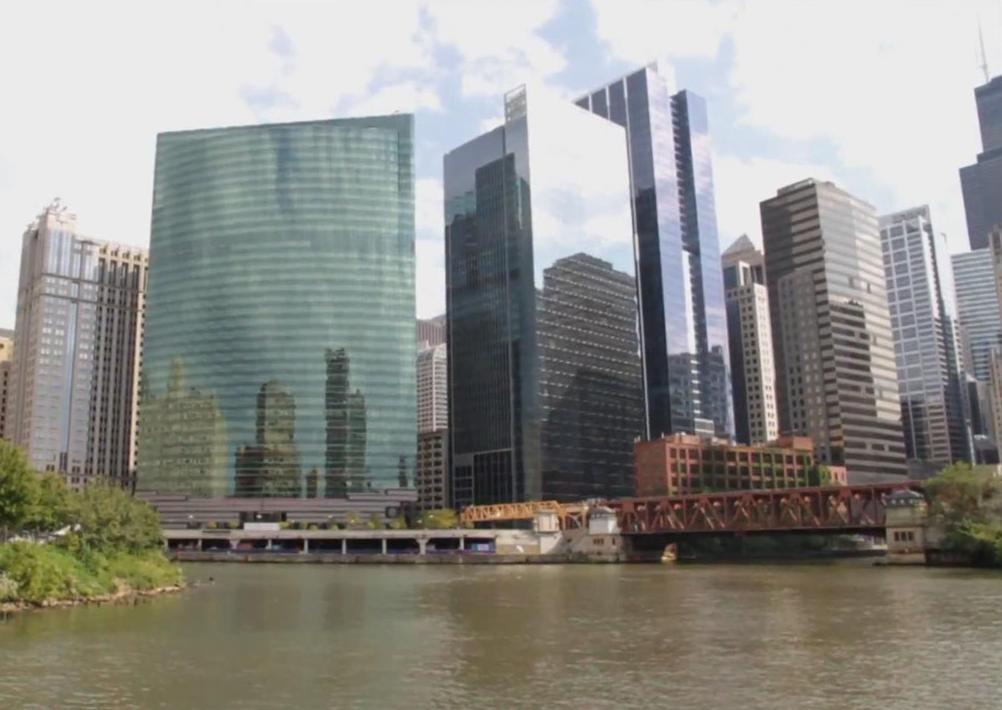}
	\includegraphics[width=0.23\linewidth, height=0.2\linewidth]{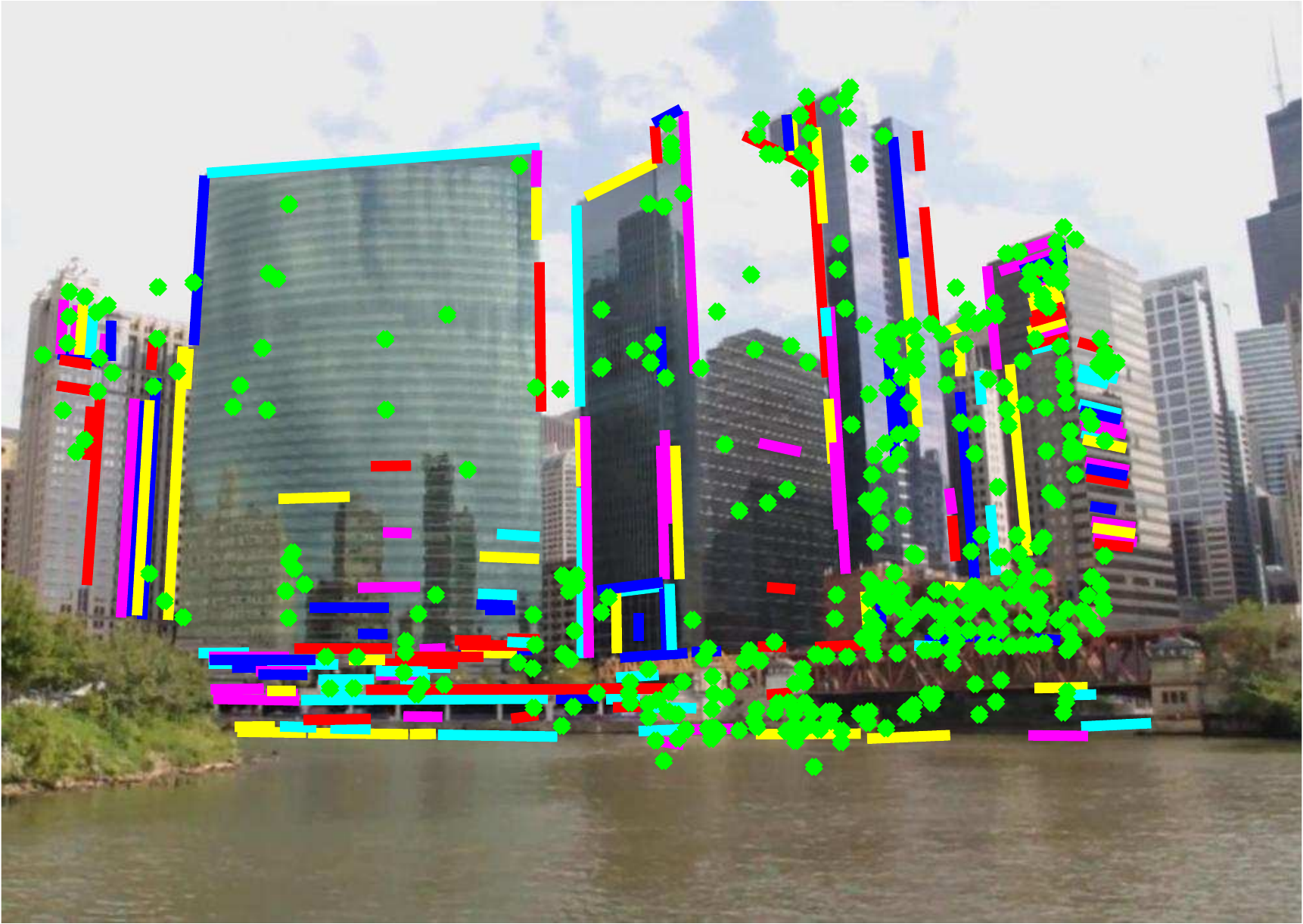}\\ [1mm]
	\includegraphics[width=0.23\linewidth, height=0.2\linewidth]{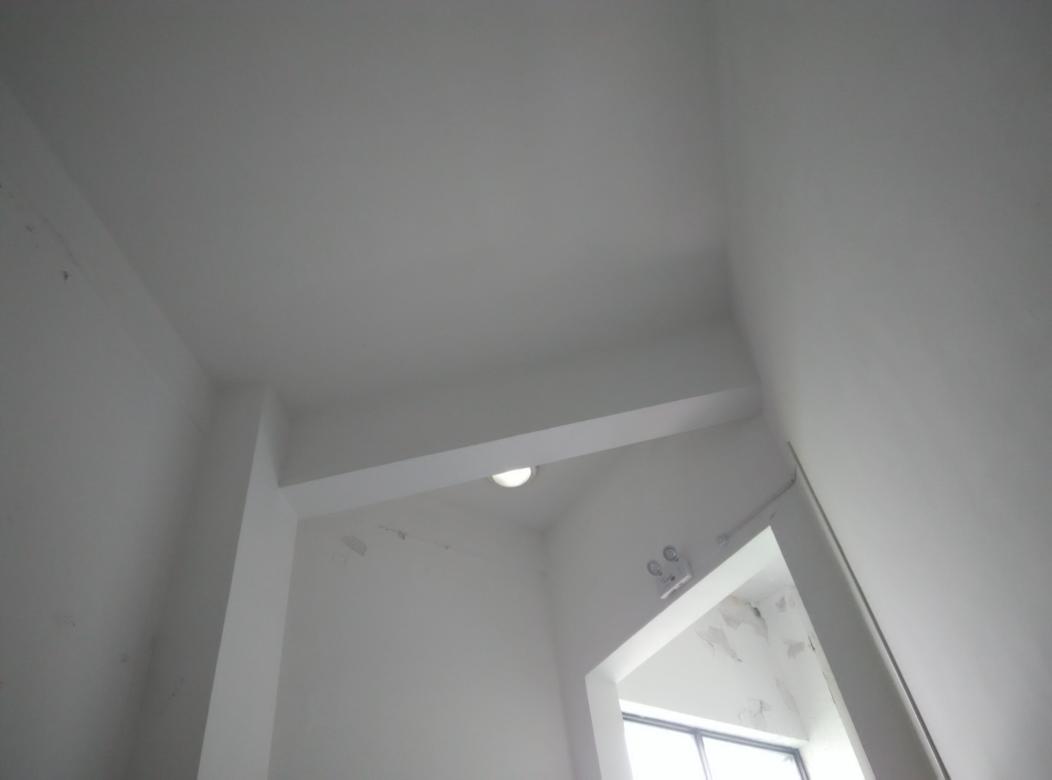}\
	\includegraphics[width=0.23\linewidth, height=0.2\linewidth]{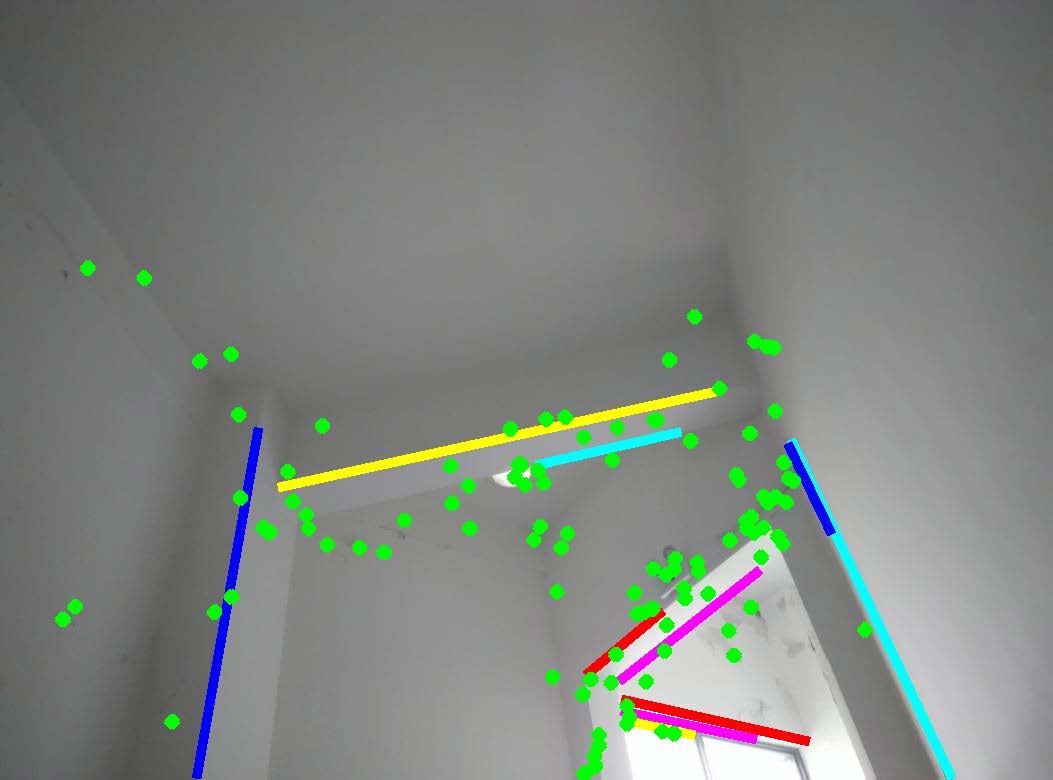}
	\includegraphics[width=0.23\linewidth, height=0.2\linewidth]{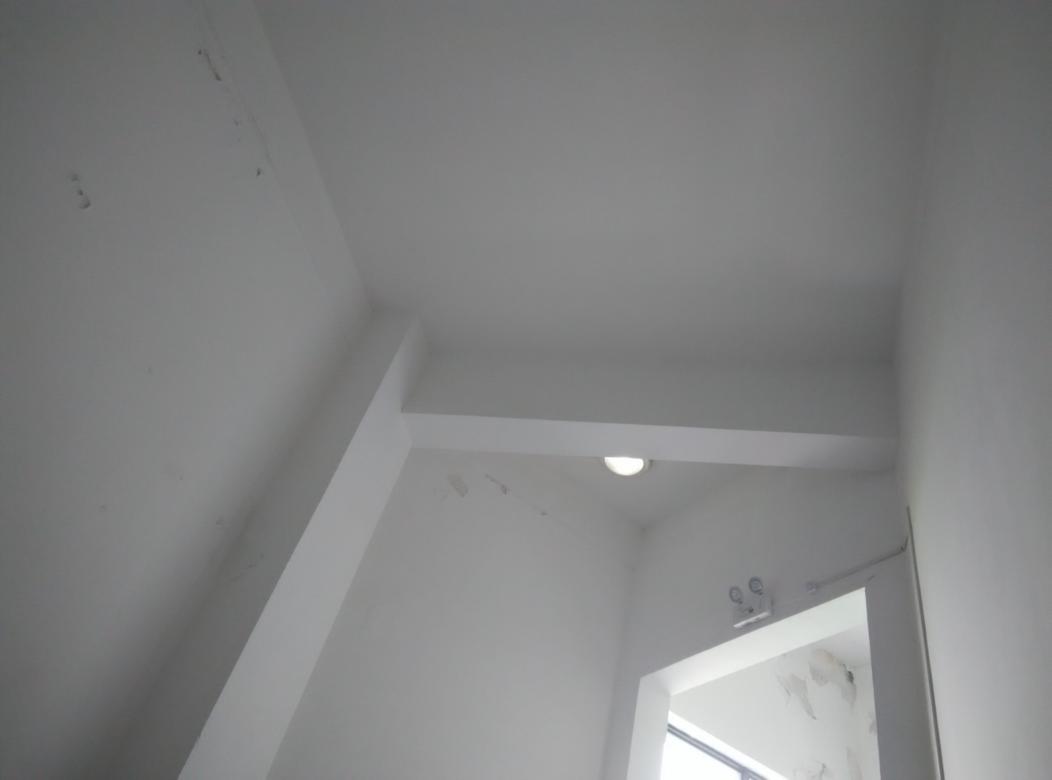}
	\includegraphics[width=0.23\linewidth, height=0.2\linewidth]{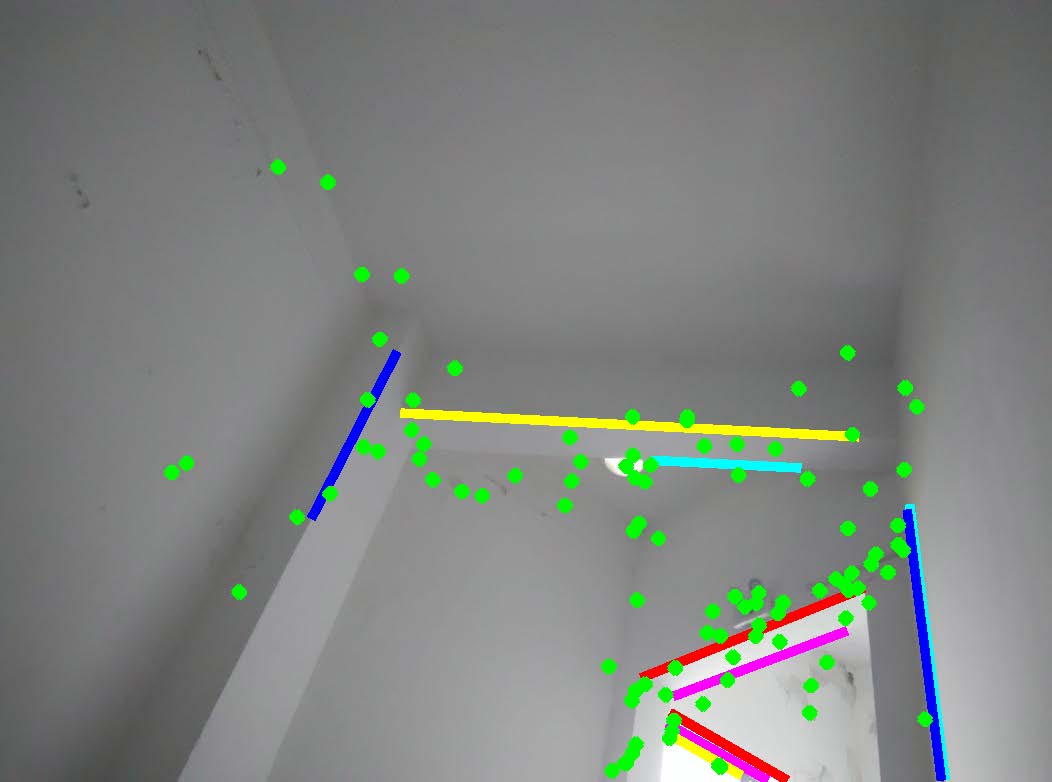}\\
	\caption{The original images for comparison of local-based methods. From top to bottom, they are:~\emph{Church},~\emph{Block}, and~\emph{Wall}.}
	\label{fig_localorig}
\end{figure*}

\begin{figure*}[htp!]
	\centering
	\includegraphics[width=0.24\textwidth]{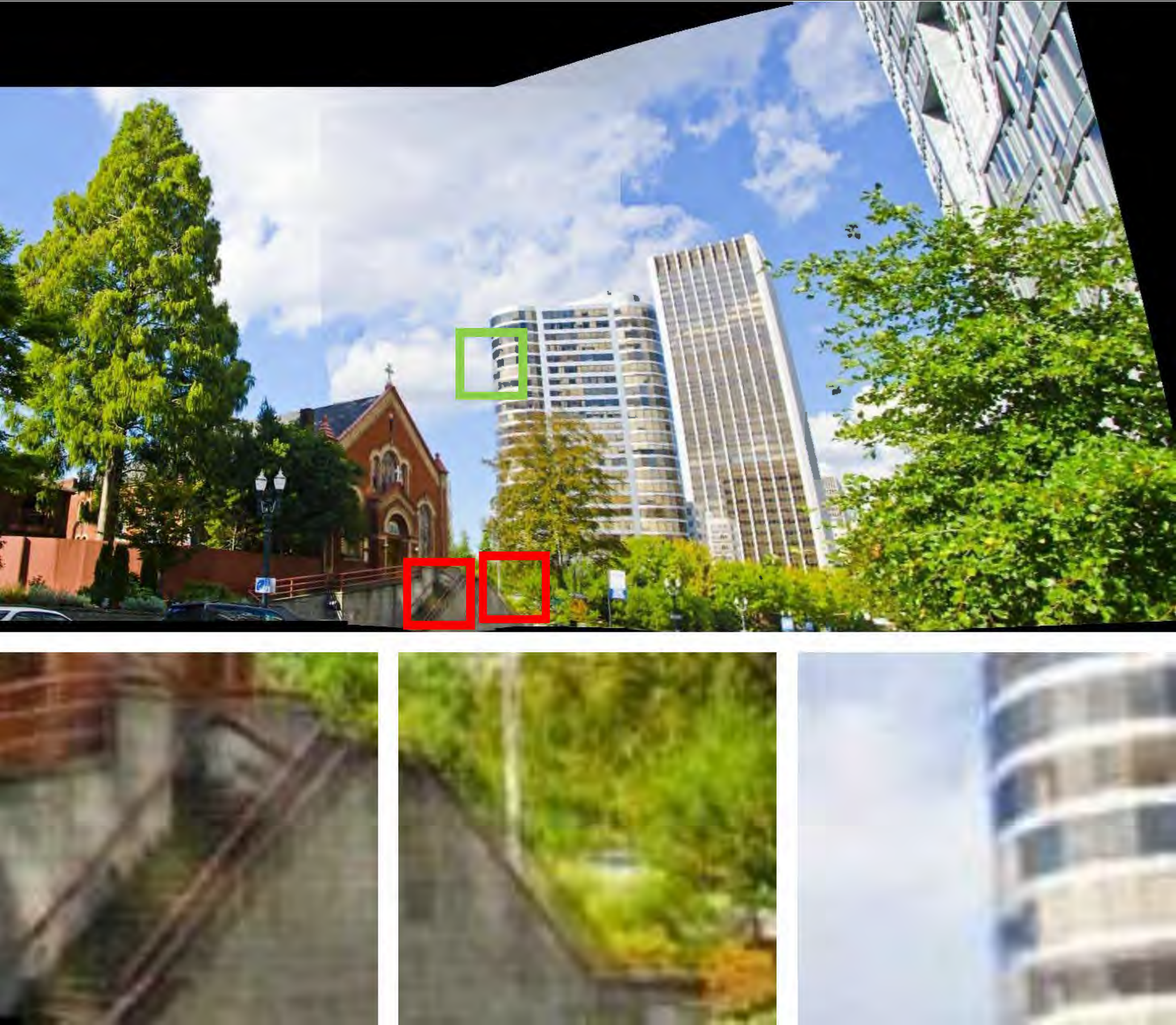}\
	\includegraphics[width=0.24\textwidth]{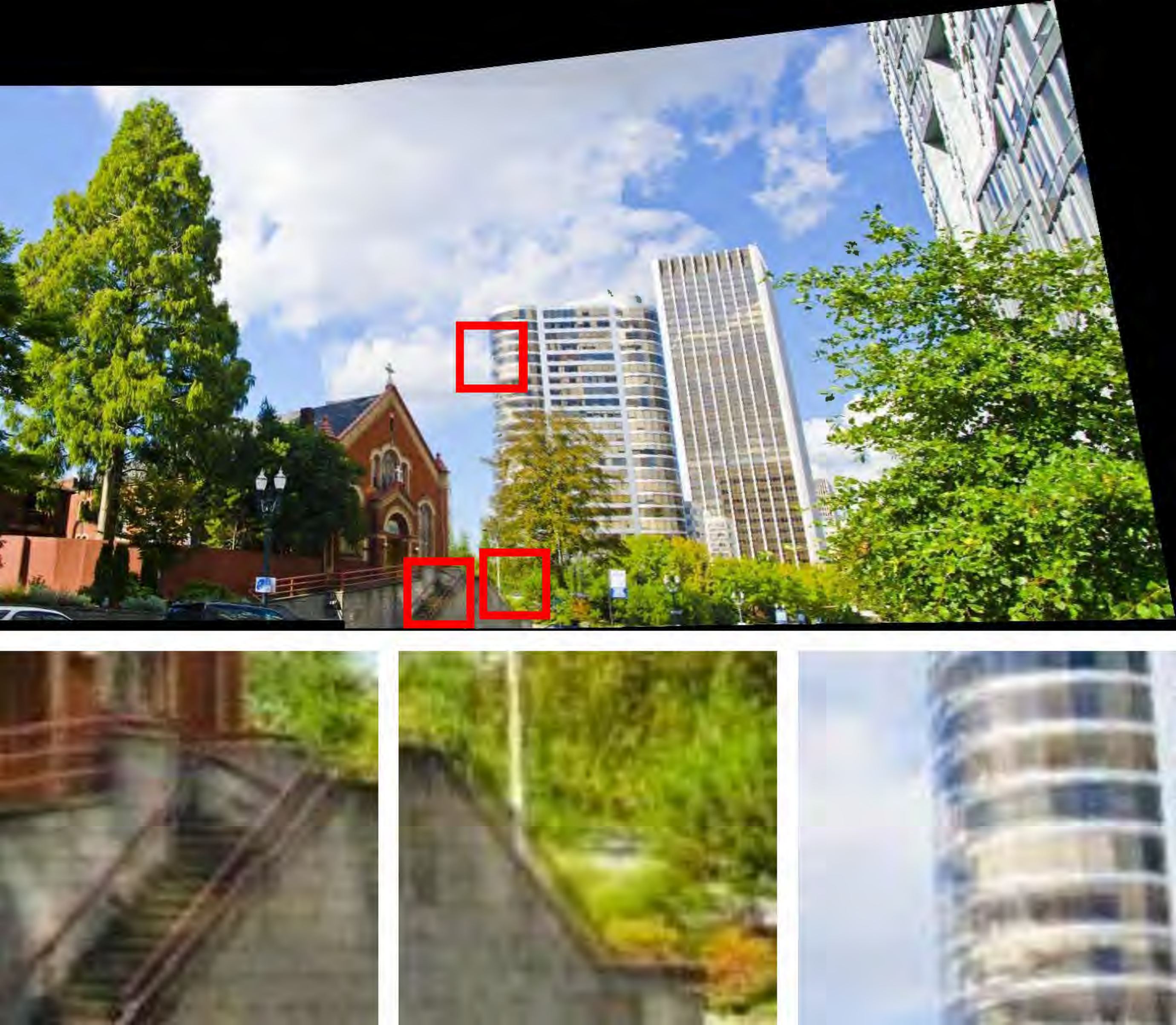}\
	\includegraphics[width=0.24\textwidth]{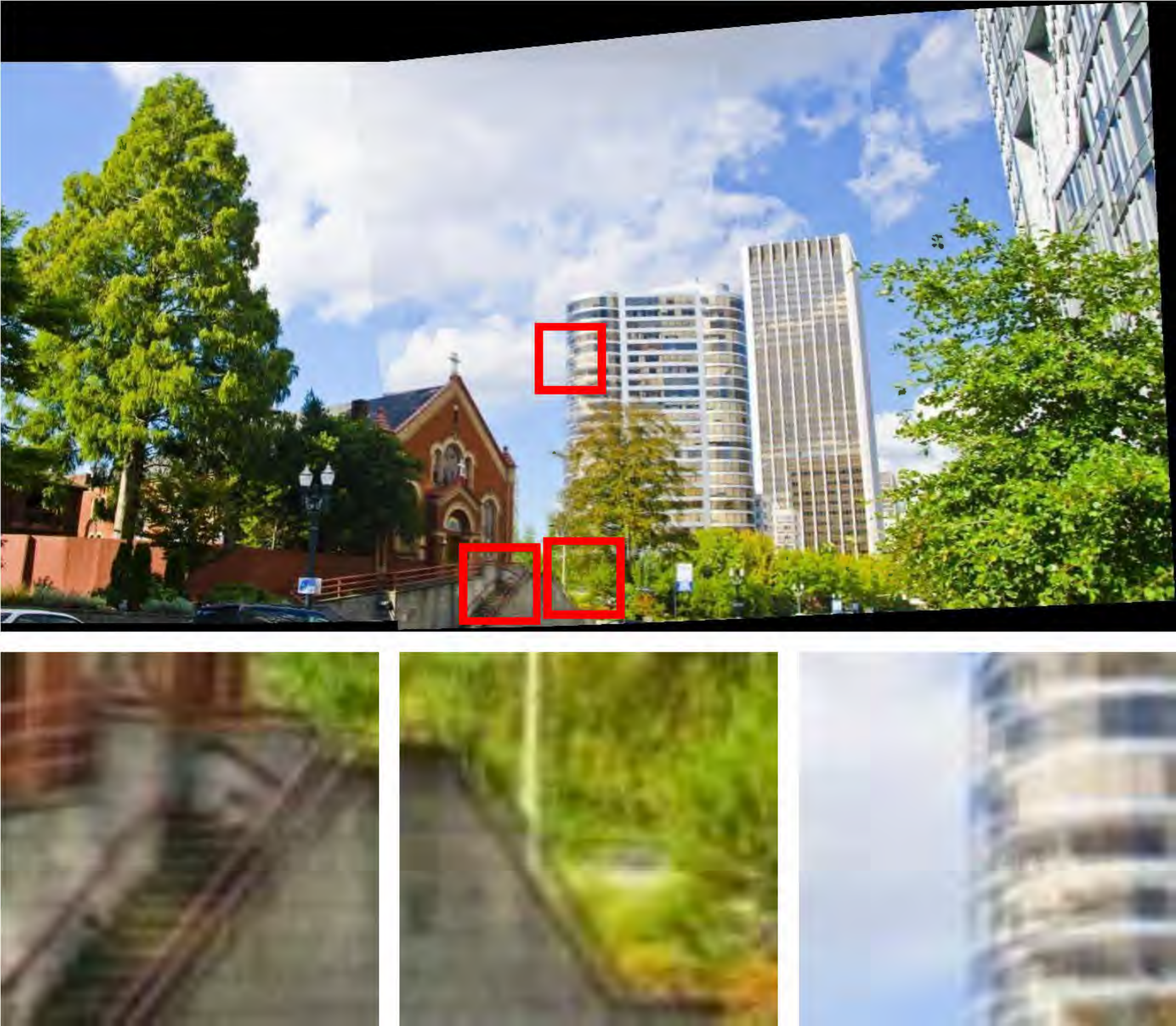}\
	\includegraphics[width=0.24\textwidth]{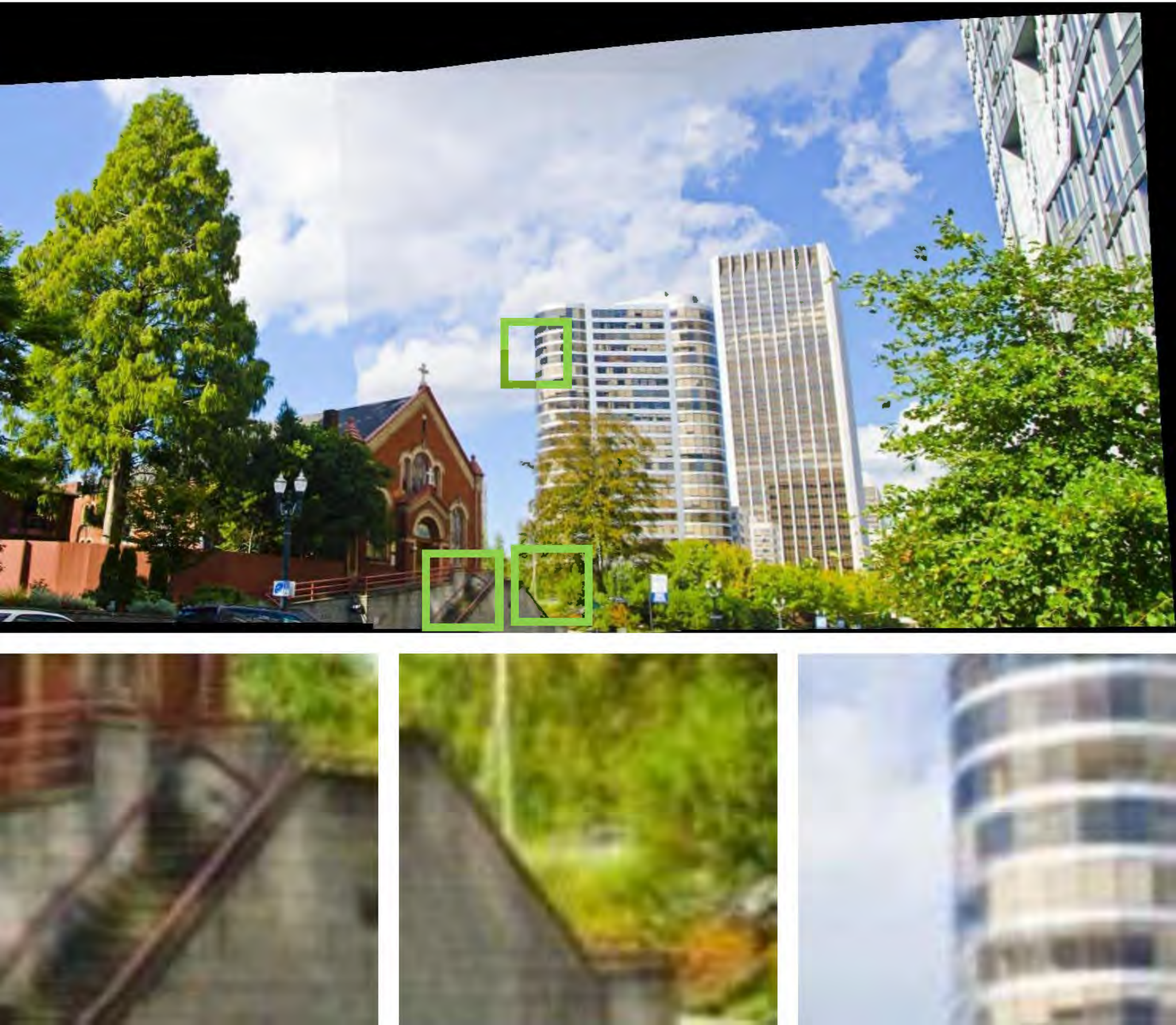}\\[1mm]
	\includegraphics[width=0.24\textwidth]{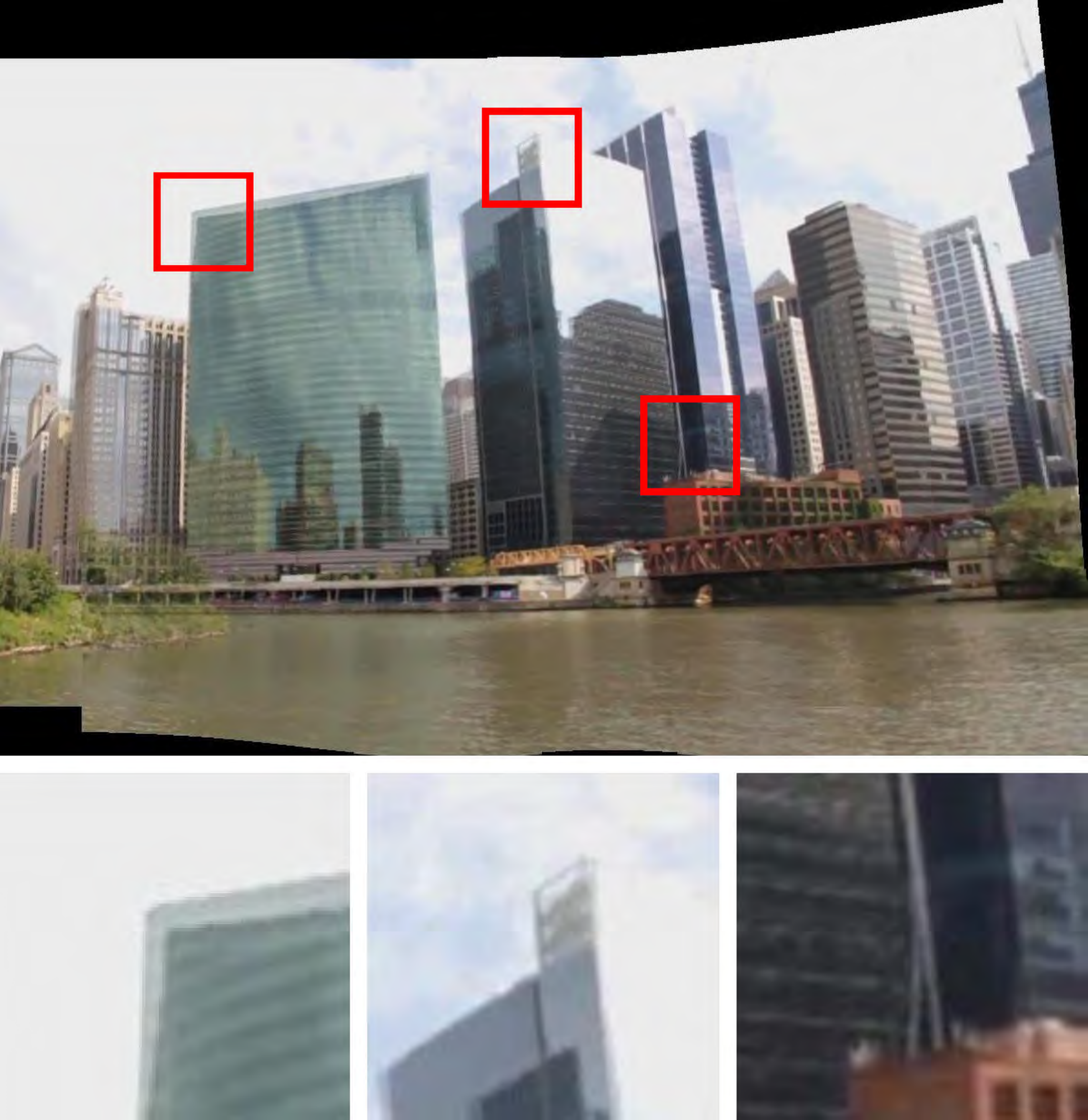}\
	\includegraphics[width=0.24\textwidth]{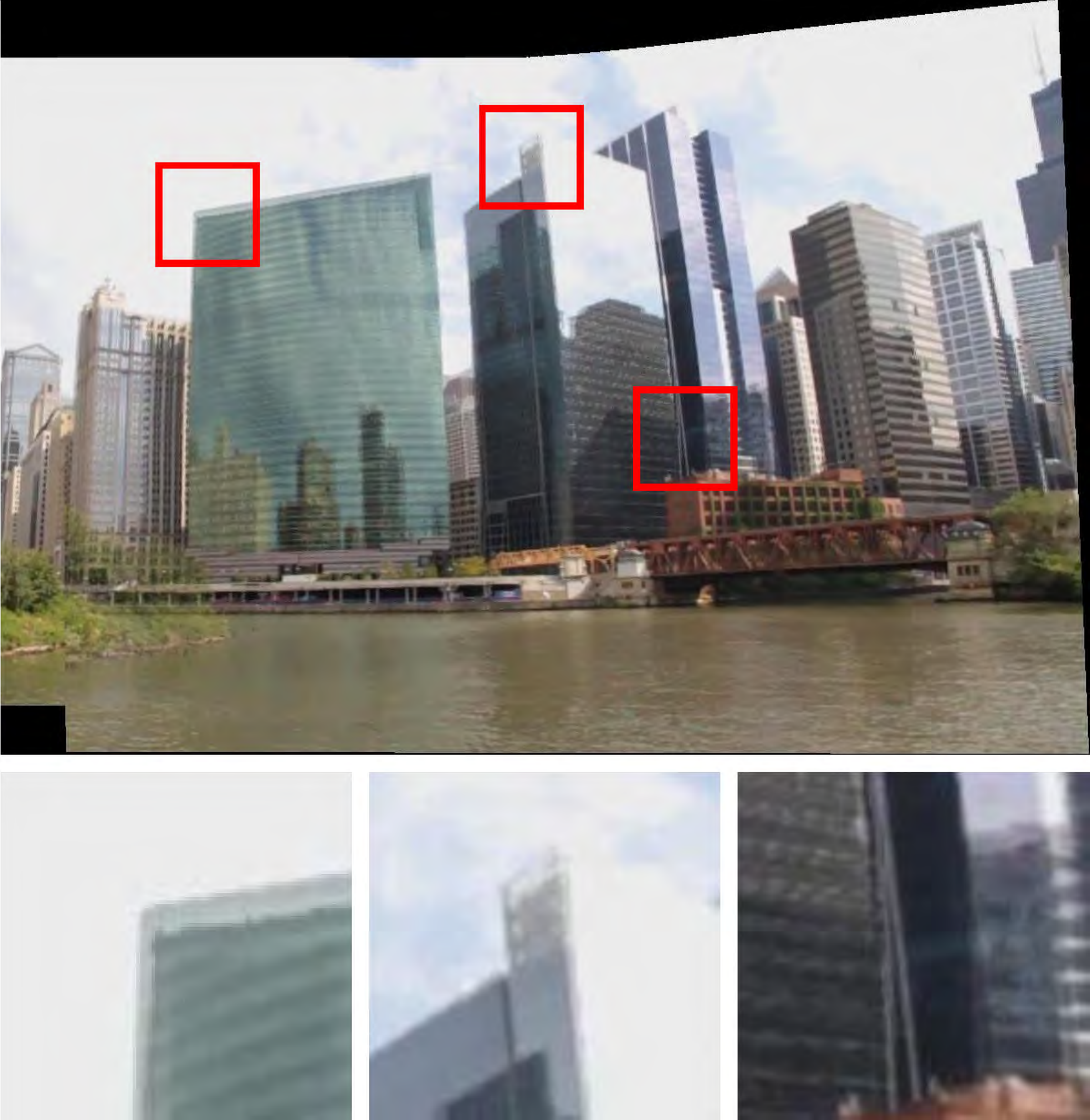}\
	\includegraphics[width=0.24\textwidth]{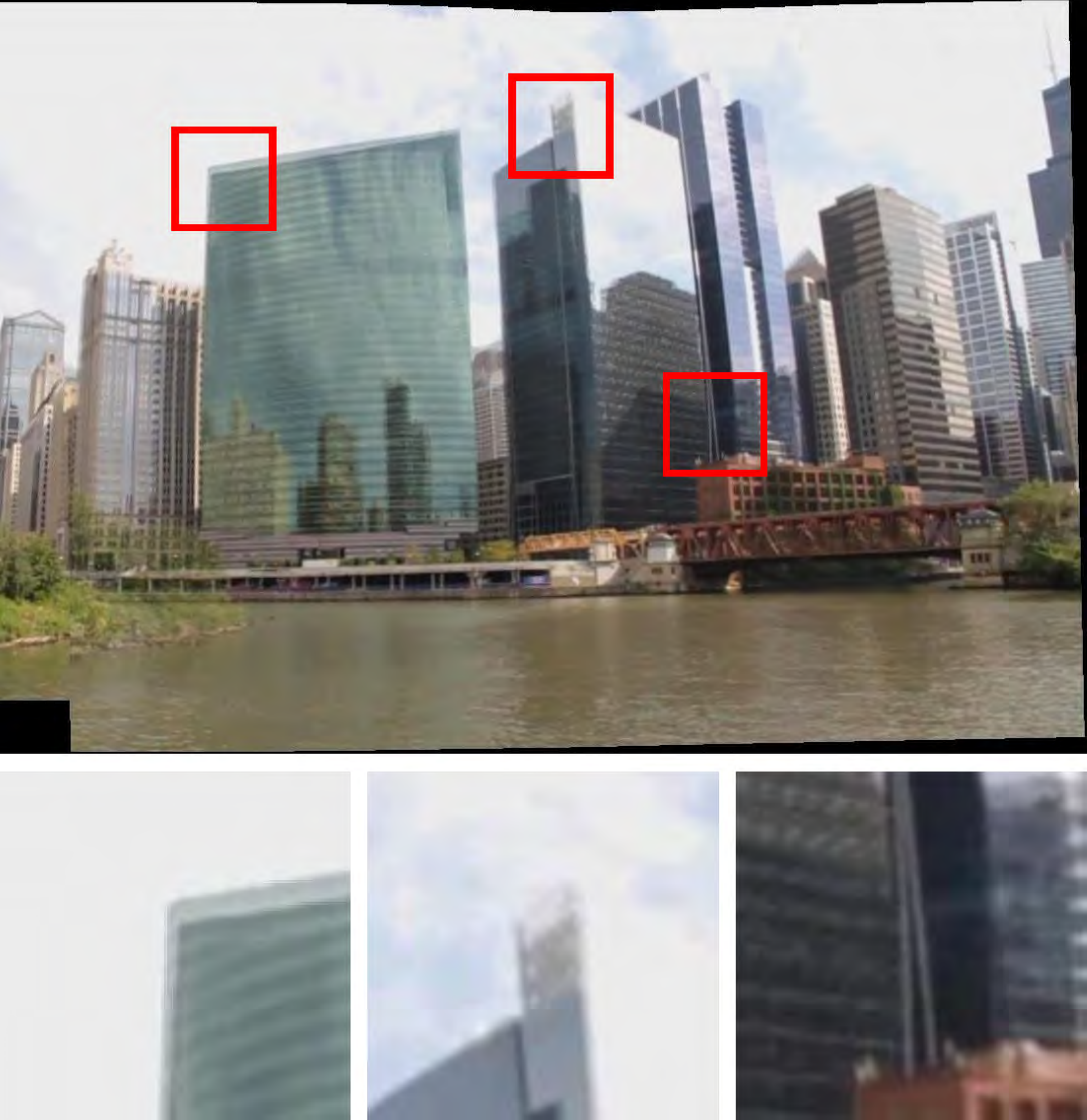}\
	\includegraphics[width=0.24\textwidth]{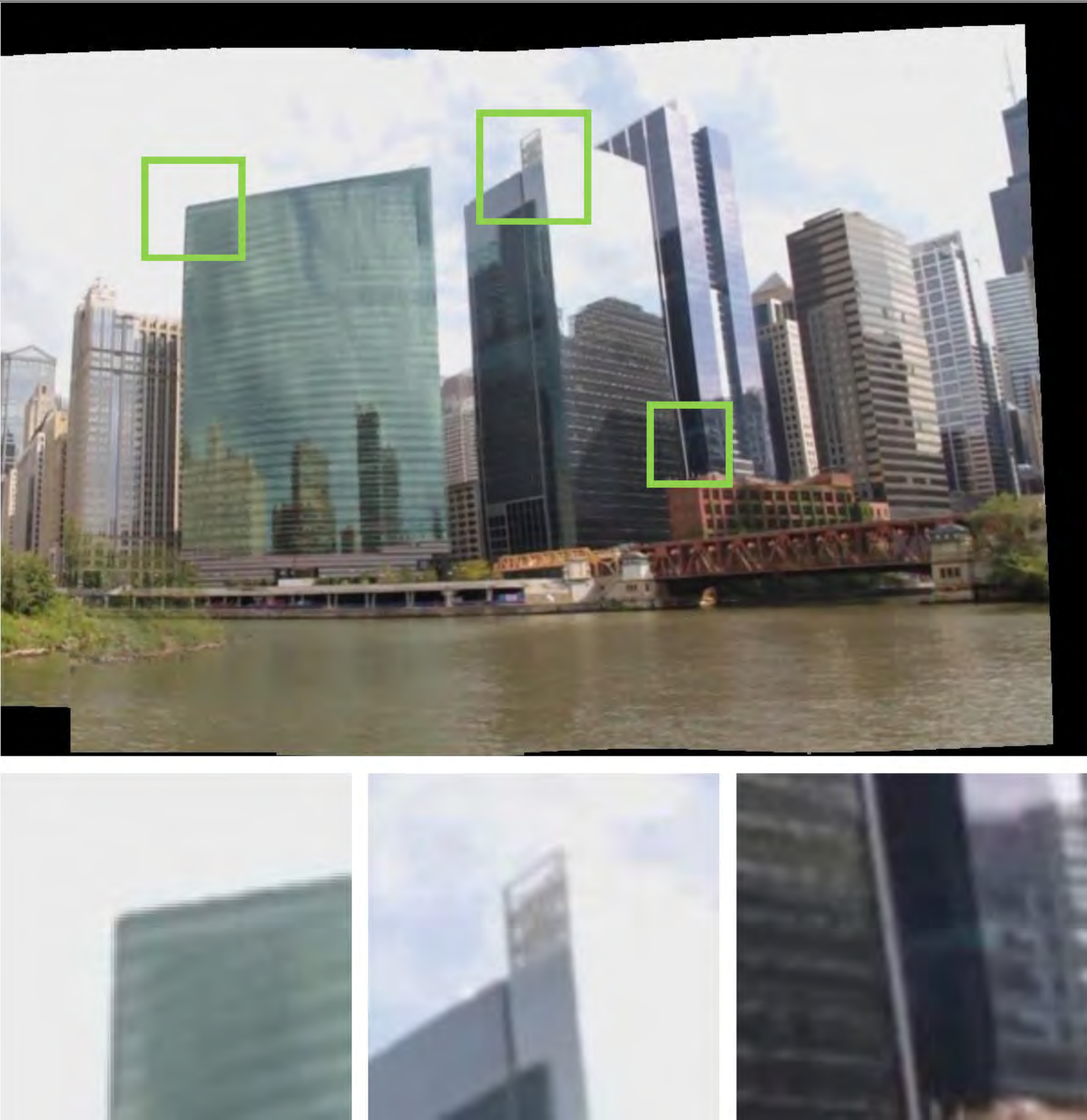}\\[1mm]
	\includegraphics[width=0.24\textwidth]{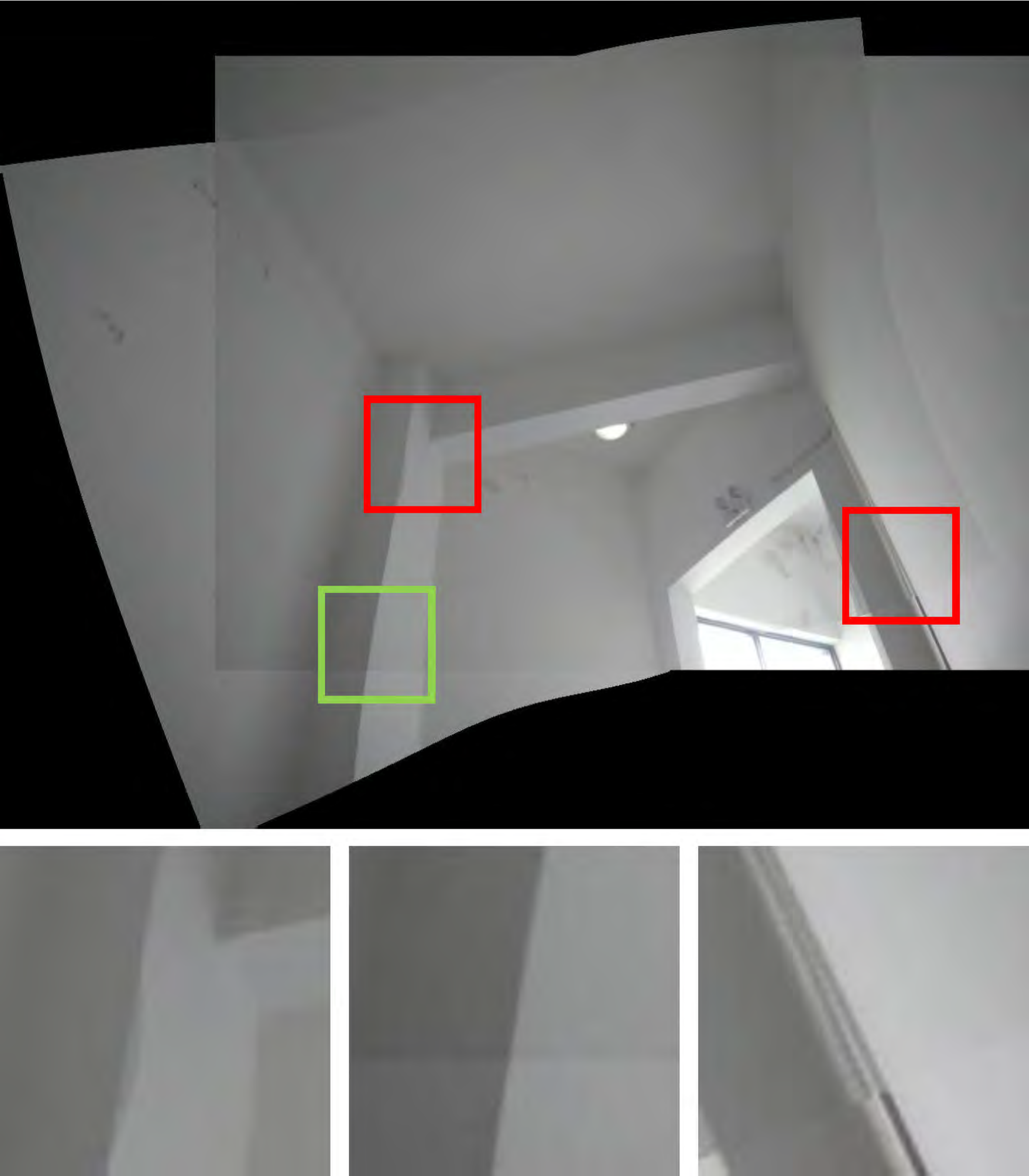}\
	\includegraphics[width=0.24\textwidth]{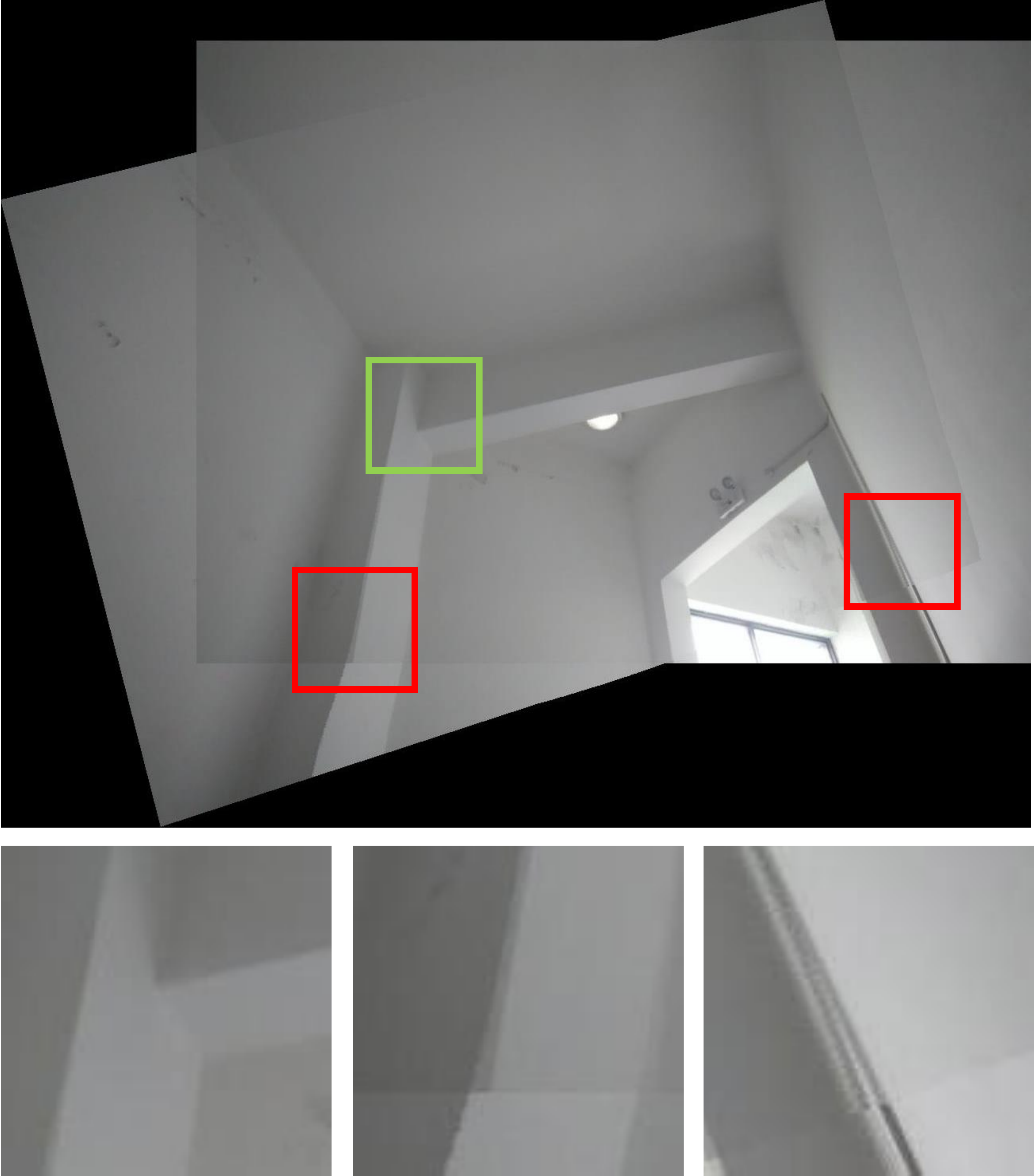}\
	\includegraphics[width=0.24\textwidth]{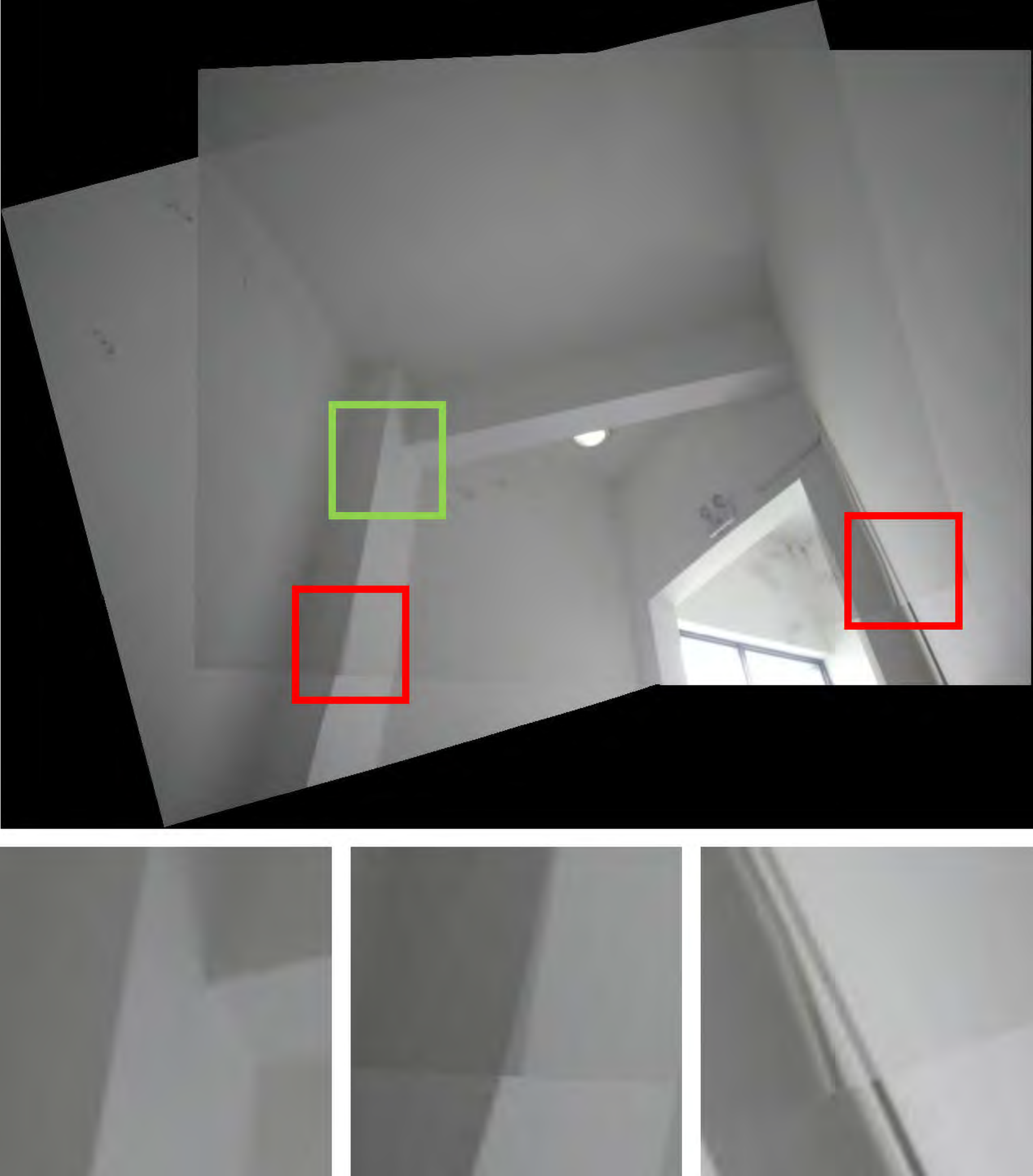}\
	\includegraphics[width=0.24\textwidth]{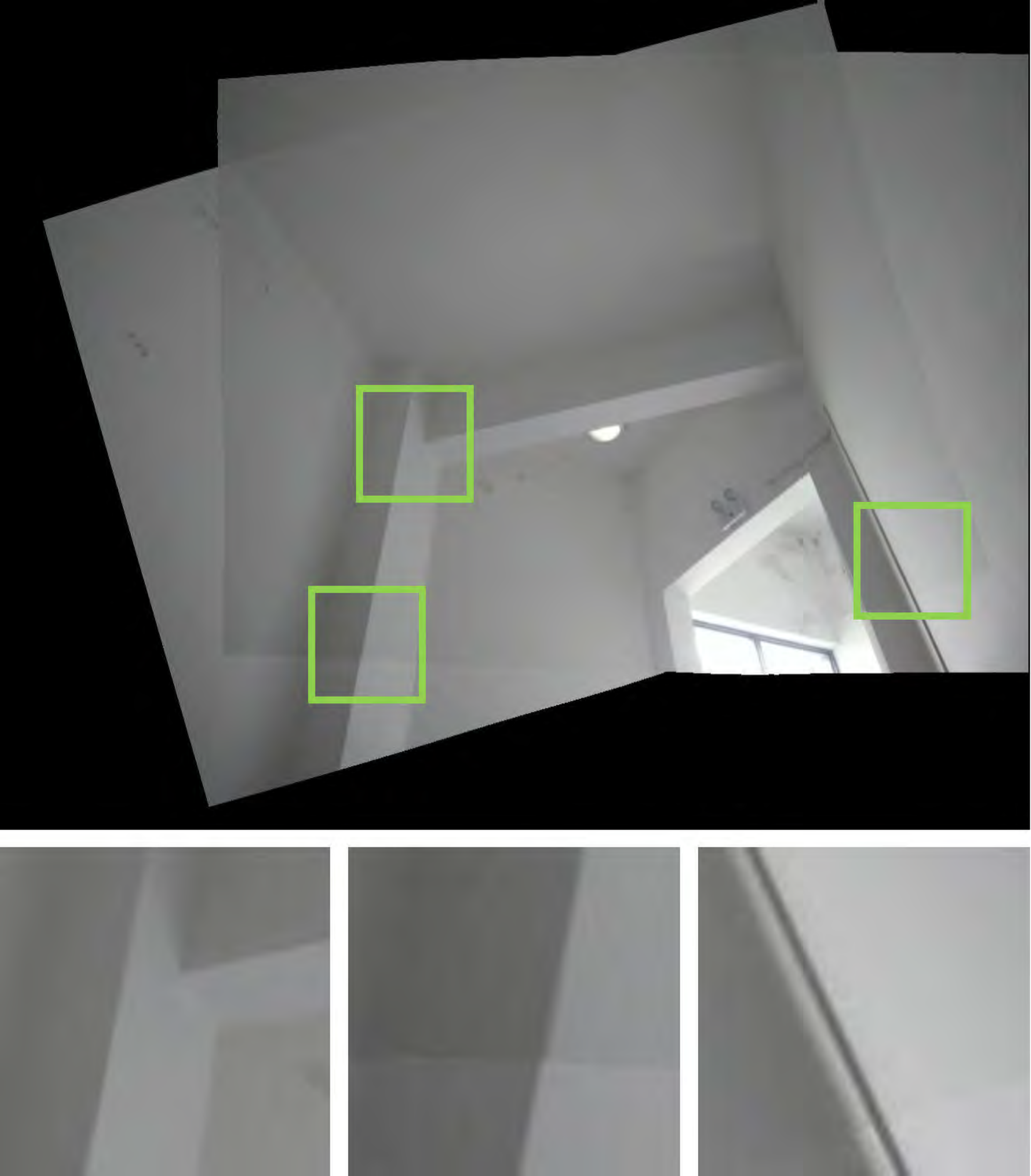} \\[1mm]
	\caption{Comparison of image stitching on~\emph{Church},~\emph{Block}, and~\emph{Wall}. From left to right, the images show the results of (a) CPW~\cite{Liu2009}, (b) APAP~\cite{Zaragoza2014}, (c) SPHP+APAP~\cite{Chang2014}, (d) our method (local version). The details are highlighted to simplify the comparison. The red boxes show alignment errors or distortions, and the green boxes show satisfactory stitching. Enlarged views are displayed below each image stitching result.}
	\label{fig_localexperi}
\end{figure*}

The global version works well in preserving the content and perspective, but it is somewhat less robust when aligning images taken with large views. For high DoFs and flexible local homographies, our method that uses local homography in the pre-warping stage (called the local version) can handle the parallax issue. Therefore, in this section, we compared it with three other local-based methods: CPW~\cite{Liu2009}, APAP~\cite{Zaragoza2014}, and SPHP+APAP~\cite{Chang2014}. Fig.~\ref{fig_localorig} shows the original \emph{Church}, \emph{Block}, and \emph{Wall} images for the comparison experiments\footnote{The \emph{Church} and \emph{Block} images were acquired from the open dataset of~\cite{Zhang2014}.}. Some images have little texture, which limits the extracted features. Moreover, the images$'$ corresponding views vary greatly.

The stitching results on these three pairs of images are provided in Fig.~\ref{fig_localexperi}. In terms of alignment accuracy, CPW and APAP allow higher DoFs than does global homography, but they also produce misalignments in regions that lack point correspondences (the areas partially highlighted in red boxes). In addition, CPW and APAP may cause local structure deformation in structural regions that lack keypoints. The red closeups clearly show that straight lines are bent (\emph{e.g.}, the stair railing in \emph{Church}, the building edge in \emph{Block}, and the wall edge in \emph{Wall}). Using the similarity transformation, SPHP+APAP reduces the projective distortions and preserves the shape and perspective, mitigating the building distortion in the non-overlapping regions in both \emph{Church} and \emph{Block}. In comparison, our method not only provides accurate alignment, which benefits from the two-stage alignment scheme, but also preserves image structures and perspectives due to the line and similarity constraints.

Table~\ref{localtab} shows the quantitative results of the compared methods. Our method consistently achieves better accuracy than CPW, APAP and SPHP+APAP except for ${Err}_{mg}^{(p)}$ in \emph{Church} result. CPW adopts feature alignment as a strong constraint; therefore, it provides a good quantitative result in ${Err}_{mg}^{(p)}$. However, its results are unsatisfactory on other criteria on the \emph{Church} image. Overall, our method achieves the best quantitative results.

\begin{table}[htp!]
	\centering
	\caption{Quantitative evaluation of local-based methods}
	\label{localtab}
	\resizebox{0.98\textwidth}{!}{		
		\begin{tabular}{c|cccc|cccc|cccc}
			\hline
			\multirow{2}{*}{Methods}  &\multicolumn{4}{c|}{\textit{Church}}  &\multicolumn{4}{c|}{\textit{Block}}  & \multicolumn{4}{c}{\textit{Wall}}  \\
			&\textit{Cor} &${Err}_{mg}^{(p)}$ &${Err}_{mg}^{(l)}$  &${Err}_{mg}$
			&\textit{Cor} &${Err}_{mg}^{(p)}$ &${Err}_{mg}^{(l)}$  &${Err}_{mg}$
			&\textit{Cor} &${Err}_{mg}^{(p)}$ &${Err}_{mg}^{(l)}$  &${Err}_{mg}$         \\ \hline
			CPW        &4.950          &\textbf{0.599} &0.876  &0.686            &2.561 & 1.600 &1.582  &1.592      &0.308   &2.348  & 2.100  &2.312        \\
			APAP       &6.485          &1.319          &1.261  &1.300            &3.013 &2.719  &1.710  &2.263      &0.252   &3.490  &2.178   & 3.302       \\
			SPHP+APAP  &4.281          &1.310          &1.280  &1.301            &2.849 &2.668  &1.651  &2.208      &0.198   &3.498  &2.249   &3.318      \\
			Proposed   &\textbf{3.090} &0.630          &\textbf{0.515} &\textbf{0.594}     &\textbf{1.880} &\textbf{1.550} &\textbf{0.627} &\textbf{1.133}
			&\textbf{0.081} &\textbf{2.248} &\textbf{0.478} &\textbf{1.993}     \\ \hline
		\end{tabular}
	}
\end{table}

\subsection{Stitching of multiple images} 
\begin{figure*}[!t]
	\centering  
	\includegraphics[width=0.8\textwidth]{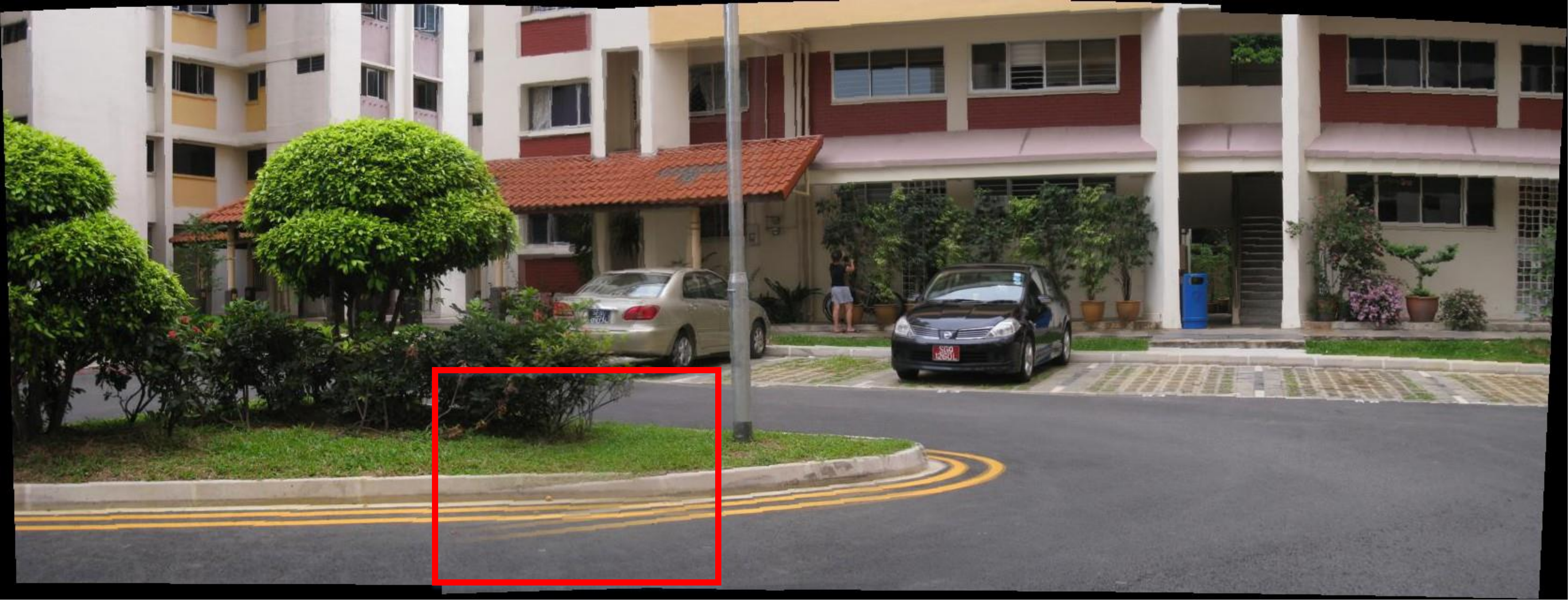}\\[1mm]
	\includegraphics[width=0.8\textwidth]{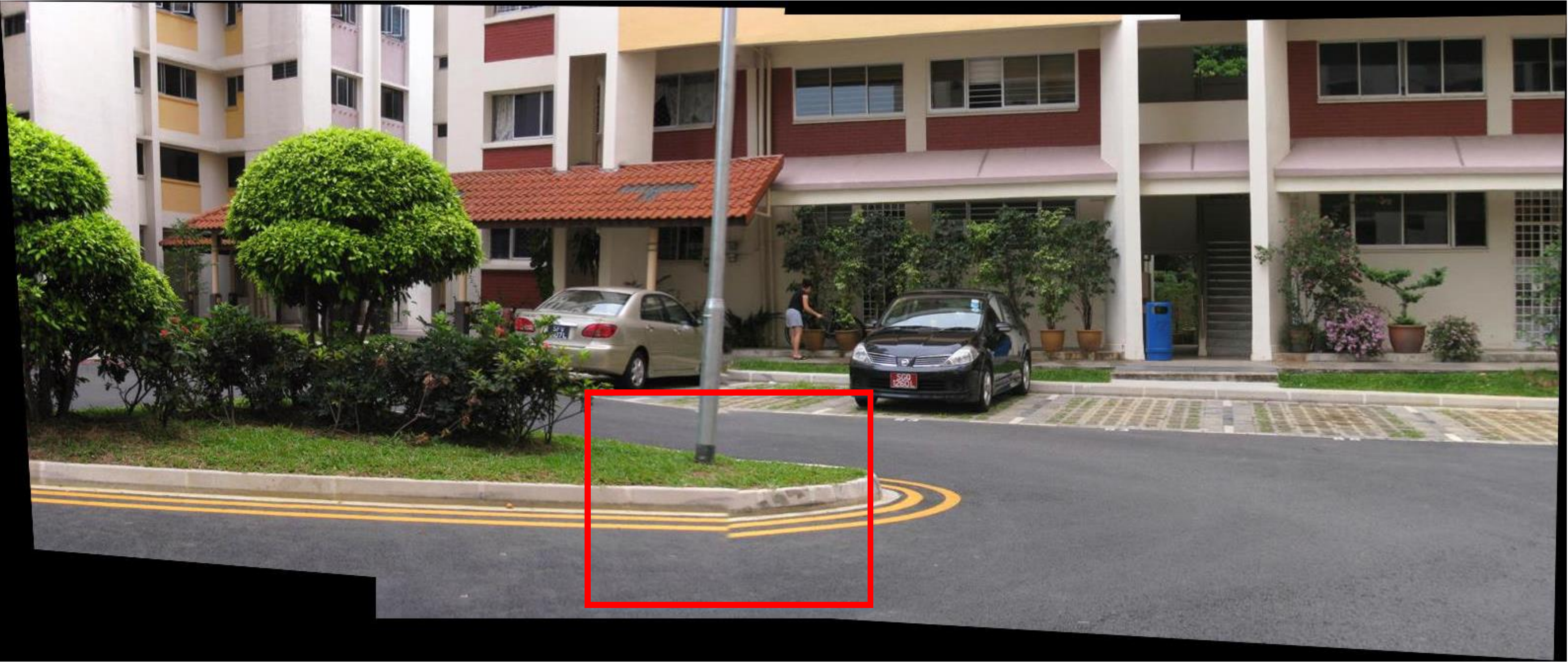}\\[1mm]
	\includegraphics[width=0.8\textwidth]{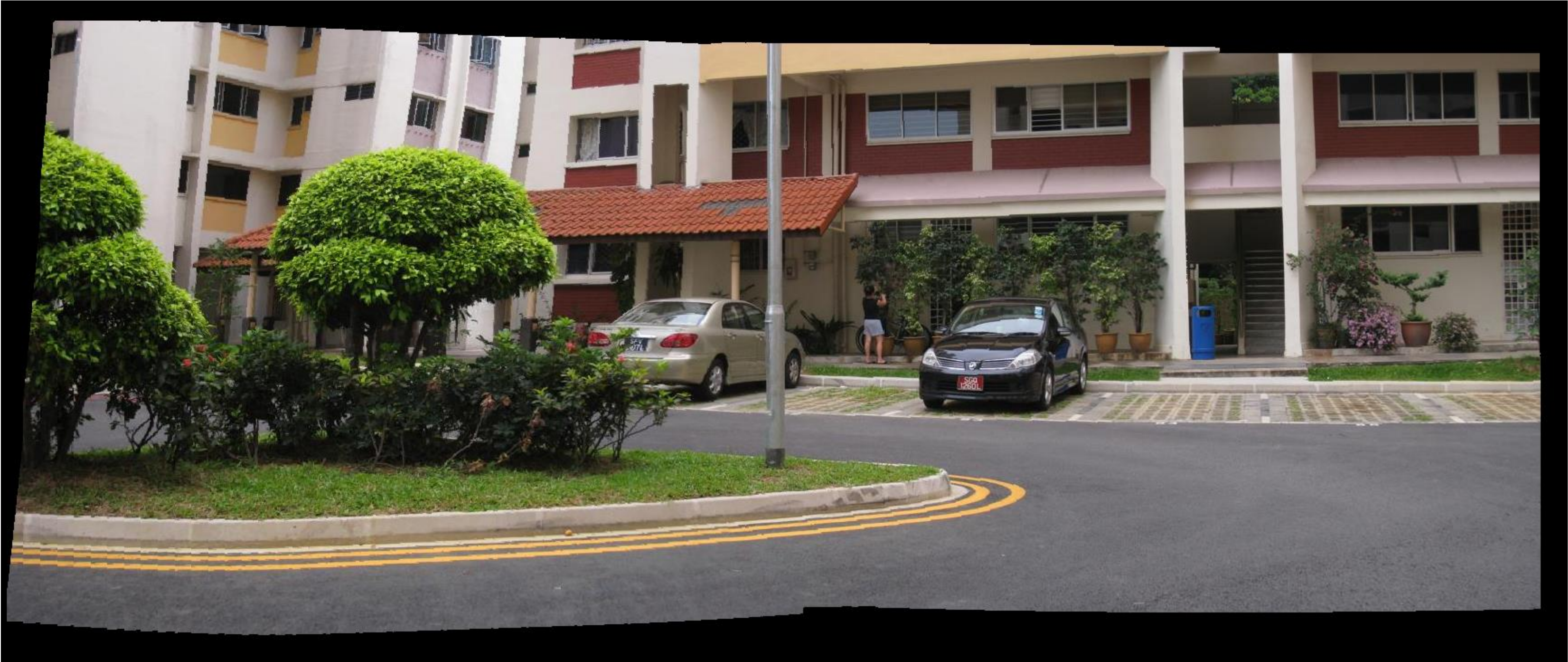}\\[1mm]
	\caption{Stitching of multiple images on~\emph{Apartments} (3 images). From top to bottom, the results are (a) Autostitch~\cite{Brown2007}, (b) ICE~\cite{ICE2015}, and (c) our method. Some errors are highlighted by red boxes to simplify the comparison.}
	\label{fig_pano}
\end{figure*}

\begin{figure*}[!t]
	\centering  
	\includegraphics[width=0.9\textwidth]{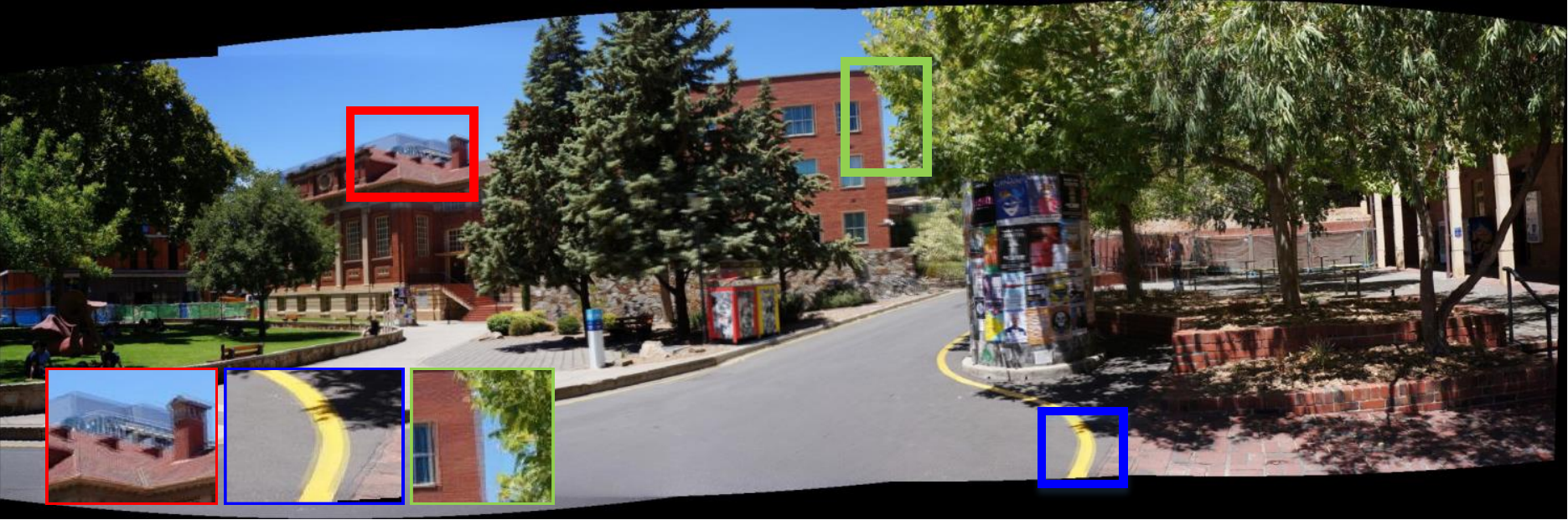}\\[1mm]
	\includegraphics[width=0.9\textwidth]{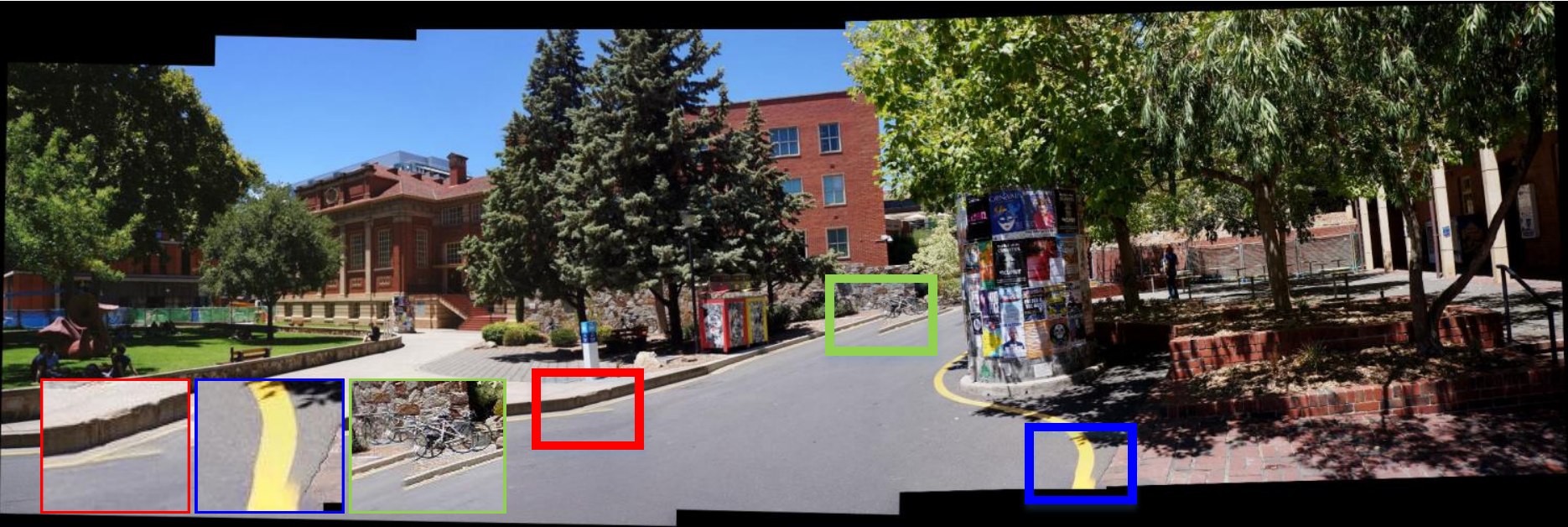}\\[1mm]
	\includegraphics[width=0.9\textwidth]{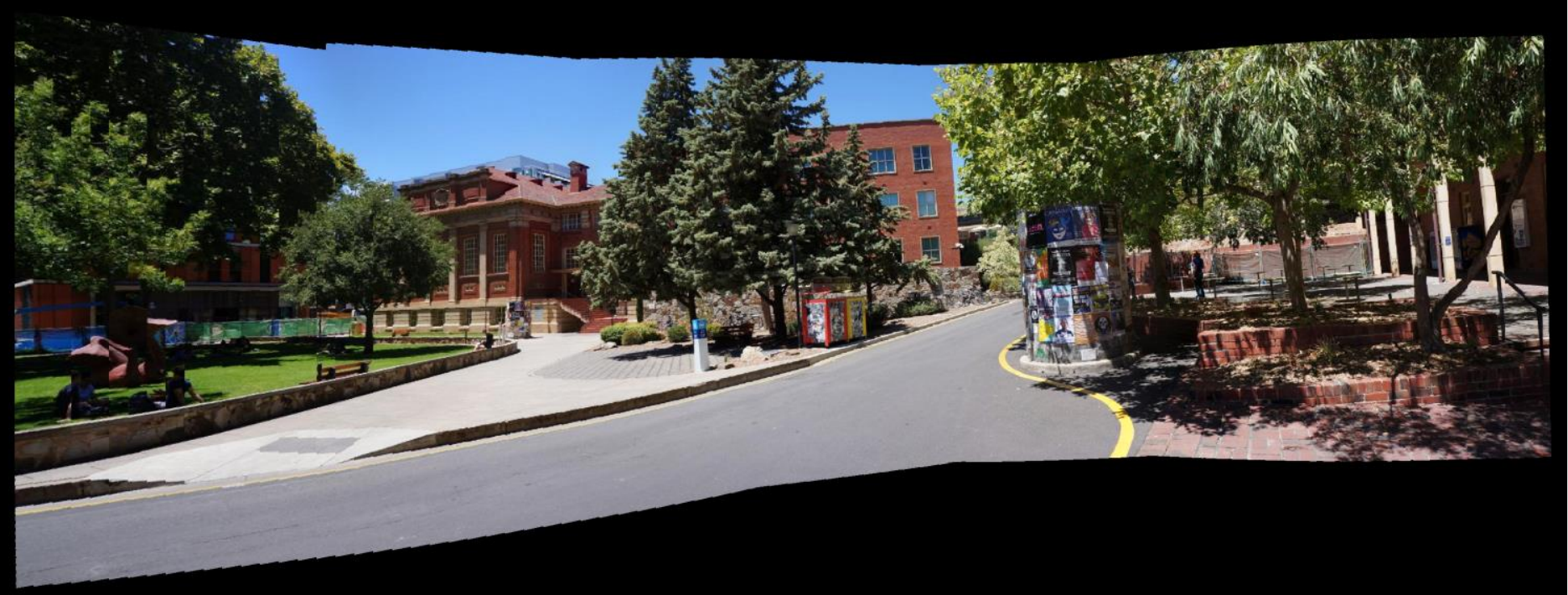}\\[1mm]
	\caption{Stitching of multiple images on~\emph{Garden} (5 images). From top to bottom, the results are (a) Autostitch~\cite{Brown2007}, (b) ICE~\cite{ICE2015}, and (c) our method. Some errors are highlighted in the closeups to simplify the comparison.}
	\label{fig_pano2}
\end{figure*}

Figs.~\ref{fig_pano} and \ref{fig_pano2} show the stitching results of multiple images on the \emph{Apartments} and \emph{Garden} data, respectively\footnote{These images were acquired from the open dataset of~\cite{Zaragoza2014}}. Some distinct errors are highlighted in boxes. As can be seen, Autostitch and ICE result in some obvious misalignments because they use only global homography for alignment, which is unsuitable for images whose views differ by factors other than pure rotation. In contrast, our method largely improves the stitching performance because of the flexible line-guided local homographies and mesh optimization. Thus, the proposed method produces satisfactory stitching results that contain few misalignments and distortions.

\section{Conclusion}
\label{sec:conclusion}

This paper proposed \emph{a line-guided local warping for image stitching by imposing similarity  constraint}. Our method integrates multiple constraints, including line features and global similarity constraints, into a two-stage image stitching framework that achieves accurate alignment and mitigates distortions. The line features are employed as an effective supplement to point features for alignment. Then, the line feature constraints (line matching and line collinearity) are integrated into the mesh-based warping framework, which further improves the alignment while preserving the image structures. Additionally, the global similarity transformation is combined with the projective warping to maintain the image content and perspective. As shown by the results of performed experiments, the proposed method achieves a good image stitching result that yields the fewest alignment errors and distortions compared to other methods. The proposed method depends on line detection and matching; thus, incomplete or broken line segments may influence its structure-preserving performance. In future work, we would like to explore other complex structure constraints, such as contours~\cite{WangFBLL2014, BaiYL2008}, to improve the image stitching performance, and explore the possibility of applying our warping model to other applications, such as video stabilization~\cite{LinJLL2017}.

%

\bibliographystyle{IEEEtran}
\bibliography{mybibfile}
\end{document}